\documentclass[letterpaper]{article} % DO NOT CHANGE THIS
\usepackage{aaai25}  % DO NOT CHANGE THIS
\usepackage{times}  % DO NOT CHANGE THIS
\usepackage{helvet}  % DO NOT CHANGE THIS
\usepackage{courier}  % DO NOT CHANGE THIS
\usepackage[hyphens]{url}  % DO NOT CHANGE THIS
\usepackage{graphicx} % DO NOT CHANGE THIS
\usepackage[table,xcdraw]{xcolor}
\usepackage{tikz}
\usepackage{booktabs}
\usepackage{amsmath}
\usepackage{natbib}  % DO NOT CHANGE THIS AND DO NOT ADD ANY OPTIONS TO IT
\usepackage{caption} % DO NOT CHANGE THIS AND DO NOT ADD ANY OPTIONS TO IT
\usepackage{algorithm}
\usepackage{algorithmic}

\usepackage{newfloat}
\usepackage{listings}
\DeclareCaptionStyle{ruled}{labelfont=normalfont,labelsep=colon,strut=off} % DO NOT CHANGE THIS
\floatstyle{ruled}
\newfloat{listing}{tb}{lst}{}
\floatname{listing}{Listing}
\usepackage{setspace}

\usepackage[many]{tcolorbox}

\newtcolorbox{defin}{colback=Teal!5!White,enhanced,title=Contributions,
	attach boxed title to top left={xshift=-4mm},boxrule=0pt,after skip=1cm,before skip=1cm,right skip=0cm,breakable,fonttitle=\bfseries,toprule=0pt,bottomrule=0pt,rightrule=0pt,leftrule=3pt,arc=0mm,skin=enhancedlast jigsaw,sharp corners,colframe=Teal!55!black,colbacktitle=Teal!55!black,boxed title style={
		frame code={ 
			\fill[Teal!25!black](frame.south west)--(frame.north west)--(frame.north east)--([xshift=3mm]frame.east)--(frame.south east)--cycle;
			\draw[line width=1mm,Teal!25!black]([xshift=2mm]frame.north east)--([xshift=5mm]frame.east)--([xshift=2mm]frame.south east);
			\draw[line width=1mm,Teal!25!black]([xshift=5mm]frame.north east)--([xshift=8mm]frame.east)--([xshift=5mm]frame.south east);
			\fill[Teal!25!black](frame.south west)--+(4mm,-2mm)--+(4mm,2mm)--cycle;
		}
	}
}
\usetikzlibrary{shapes.geometric, arrows}
\usetikzlibrary{decorations.markings}

\definecolor{first}{RGB}{210,255,140}
\definecolor{second}{RGB}{136, 162, 190}
\definecolor{third}{RGB}{129, 222, 228}
\definecolor{fourth}{RGB}{132, 84, 246}
\definecolor{fifth}{RGB}{250, 223, 112}
\definecolor{sixth}{RGB}{203, 193, 172}
\definecolor{seventh}{RGB}{88, 112, 246}
\definecolor{eighth}{RGB}{245, 192, 106}
\definecolor{nine}{RGB}{171, 162, 111}
\definecolor{ten}{RGB}{217, 217, 217}

\definecolor{paired-light-blue}{RGB}{198, 219, 239}
\definecolor{paired-dark-blue}{RGB}{49, 130, 188}
\definecolor{paired-light-orange}{RGB}{251, 208, 162}
\definecolor{paired-dark-orange}{RGB}{230, 85, 12}
\definecolor{paired-light-green}{RGB}{199, 233, 193}
\definecolor{paired-dark-green}{RGB}{49, 163, 83}
\definecolor{paired-light-purple}{RGB}{218, 218, 235}
\definecolor{paired-dark-purple}{RGB}{117, 107, 176}
\definecolor{paired-light-gray}{RGB}{217, 217, 217}
\definecolor{paired-dark-gray}{RGB}{99, 99, 99}
\definecolor{paired-light-pink}{RGB}{222, 158, 214}
\definecolor{paired-dark-pink}{RGB}{123, 65, 115}
\definecolor{paired-light-red}{RGB}{231, 150, 156}
\definecolor{paired-dark-red}{RGB}{131, 60, 56}
\definecolor{paired-light-yellow}{RGB}{231, 204, 149}
\definecolor{paired-dark-yellow}{RGB}{141, 109, 49}
\definecolor{Teal}{RGB}{0, 50, 50}
\definecolor{White}{RGB}{250, 250, 250}

\definecolor{bg1}{HTML}{FF9966}
\definecolor{bg2}{HTML}{CCE5FF}
\definecolor{bg3}{HTML}{FFCC99}
\definecolor{bg4}{HTML}{FFC107}
\definecolor{bg5}{HTML}{FFCCCC}
\definecolor{bg6}{HTML}{D5E8D4}
\definecolor{bg7}{HTML}{eeeeee}
\definecolor{bg8}{HTML}{cdeb8b}
\definecolor{bg9}{HTML}{dae8fc}
\definecolor{bg10}{HTML}{a2e6eb}

\definecolor{bg31}{HTML}{FFCDD2} % light pink

\definecolor{bg32}{HTML}{F8BBD0}

\definecolor{bg33}{HTML}{E1BEE7} % lavender

\definecolor{bg34}{HTML}{D7CCC8} % light tan

\definecolor{bg35}{HTML}{B2DFDB} % light teal

\definecolor{bg36}{HTML}{A5D6A7} % light green

\definecolor{bg37}{HTML}{FFF9C4} %light yellow

\definecolor{bg38}{HTML}{FFECB3} % peach

\definecolor{bg111}{HTML}{CB6843}

\definecolor{bg112}{HTML}{D77C5C}

\definecolor{bg113}{HTML}{E28E6E}
\definecolor{bg114}{HTML}{E89F7D}
\definecolor{bg115}{HTML}{EDAE8A}
\definecolor{bg116}{HTML}{F0BA95}
\definecolor{bg117}{HTML}{F3C29F}
\definecolor{bg118}{HTML}{F6CCAA}
\definecolor{bg119}{HTML}{F8D5B3}
\definecolor{bg120}{HTML}{FADCBD}
\definecolor{bg121}{HTML}{FCE6C7}

\definecolor{bg39}{HTML}{FFE0B2} % apricot

\definecolor{bg40}{HTML}{3CB371} % blush pink

\definecolor{bg43}{HTML}{ffe5d9}

\definecolor{bg15}{HTML}{7FFFD4}

\definecolor{bg17}{HTML}{F0FFFF}

\definecolor{bg18}{HTML}{F5FFFA}

\definecolor{bg19}{HTML}{F8F8FF}

\definecolor{bg20}{HTML}{FFFFFF}

\definecolor{bg21}{HTML}{E1F5FE}

\definecolor{bg22}{HTML}{B3E5FC}

\definecolor{bg23}{HTML}{81D4FA}

\definecolor{bg24}{HTML}{4FC3F7}

\definecolor{bg25}{HTML}{29B6F6}

\definecolor{bg26}{HTML}{03A9F4}

\definecolor{bg27}{HTML}{039BE5}

\definecolor{bg28}{HTML}{0288D1}

\definecolor{bg29}{HTML}{0277BD}

\definecolor{bg30}{HTML}{01579B}

\definecolor{bg16}{HTML}{FFCC99}

\definecolor{pg51}{HTML}{E8F5E9} % pale green
\definecolor{pg52}{HTML}{C8E6C9} % honeydew green
\definecolor{pg53}{HTML}{B9F6CA} % light mint green
\definecolor{pg54}{HTML}{A9DFBF} % pale sage green
\definecolor{pg55}{HTML}{BCF5A6} % lemon green

\definecolor{pg56}{HTML}{BEF1CE} % seashell green
\definecolor{pg57}{HTML}{CEF6EC} % icy green
\definecolor{pg58}{HTML}{B7F0B1} % feijoa green
\definecolor{pg59}{HTML}{B1F2B5} % pastel light green
\definecolor{pg60}{HTML}{9DF3C4} % greenish cyan

\definecolor{pg61}{HTML}{DEF7E0} % pale green
\definecolor{pg62}{HTML}{E8F8DC} % greenish beige

\definecolor{pg63}{HTML}{EBF7E7} % seafoam green
\definecolor{pg64}{HTML}{F0FDF4} % pale turquoise

\definecolor{pg65}{HTML}{F1FEE7} % mint cream
\definecolor{pg66}{HTML}{F7FFF6} % foam green
\definecolor{pg67}{HTML}{FCFFE7} % pale spring bud
\definecolor{pg68}{HTML}{F4FFD2} % light lime green
\definecolor{pg69}{HTML}{EEFFE2} % tea green
\definecolor{pg70}{HTML}{E3FDF5} % tropical green

\definecolor{connect-color}{RGB}{0,0,0}
\definecolor{middle-color}{RGB}{255,255,255}
\definecolor{leaf-color}{RGB}{173,216,230}
\definecolor{line-color}{RGB}{25,25,112}

\usepackage{environ}

\newlength{\myl}
\expandafter\let\expandafter\origequation\csname equation*\endcsname
\expandafter\let\expandafter\endorigequation\csname endequation*\endcsname
\long\def\[#1\]{\begin{equation*}#1\end{equation*}}
\RenewEnviron{equation*}{
  \settowidth{\myl}{$\displaystyle\BODY$} % calculate width and save as \myl
  \origequation
    \ifdim\myl>\linewidth
      \resizebox{\linewidth}{!}{$\displaystyle\BODY$}% \myl > \linewidth
    \else
      \BODY % \myl <= \linewidth
    \fi
  \endorigequation
}

\usepackage{microtype}

\usepackage{inconsolata}

\usepackage[edges]{forest}
\definecolor{hidden-draw}{RGB}{20,68,106}
\definecolor{hidden-pink}{RGB}{255,245,247}
\definecolor{red}{RGB}{255,0,0}
\definecolor{hidden-draw}{RGB}{0,0,0}
\definecolor{hidden-pink}{RGB}{255,182,193}
\usepackage{tikz}

\usepackage{longtable}
\usepackage{tablefootnote}
\usepackage{url}            % simple URL typesetting
\usepackage{booktabs}       % professional-quality tables
\usepackage{amsfonts}       % blackboard math symbols
\usepackage{nicefrac}       % compact symbols for 1/2, etc.
\usepackage{microtype}      % microtypography
\usepackage{amsmath,amssymb}
\usepackage{graphicx}
\usepackage{enumitem}
\usepackage{multirow}
\usepackage{subcaption}
\usepackage{listings}
\usepackage{xspace}
\usepackage{makecell}
\usepackage{soul}               % to strikethrough
\setstcolor{red}            %to strikethrough in red
\usepackage{tikz}

\usepackage{longtable}
\usepackage{tablefootnote}
\usepackage{inconsolata}
\usepackage{graphicx}
\usepackage{caption}
\usepackage{subcaption}
\usepackage{tabularx}
\usepackage{soul}
\usepackage{float}
\usepackage{enumitem}
\usepackage{pifont}
\usepackage{arydshln}
\definecolor{vred}{HTML}{CE3A29}

\usepackage{tikz}

\tikzset{rndblock/.style={rounded corners,rectangle,draw,outer sep=0pt}}

\usepackage[draft,textsize=footnotesize,textwidth=15mm]{todonotes}

\title{The Brittleness of AI-Generated Image Watermarking Techniques: Examining Their Robustness Against Visual Paraphrasing Attacks}
\author {
    Niyar R Barman\textsuperscript{\rm 1}\equalcontrib,
    Krish Sharma\textsuperscript{\rm 1}\equalcontrib,
    Ashhar Aziz\textsuperscript{\rm 2},
    Shashwat Bajpai\textsuperscript{\rm 3},
    Shwetangshu Biswas\textsuperscript{\rm 1},
    Vasu Sharma\textsuperscript{\rm 4},
    Vinija Jain\textsuperscript{\rm 5},
    Aman Chadha\textsuperscript{\rm 5,6}\footnote{Work does no relate to position at Amazon.},
    Amit Sheth\textsuperscript{\rm 7},
    Amitava Das\textsuperscript{\rm 7}
}
\affiliations {
    \textsuperscript{\rm 1}NIT Silchar, India \quad
    \textsuperscript{\rm 2}IIIT Delhi, India \quad 
    \textsuperscript{\rm 3}BITS Pilani Hyderabad, India \quad
    \textsuperscript{\rm 4}Meta AI, USA \quad
    \textsuperscript{\rm 5}Stanford University, USA \quad
    \textsuperscript{\rm 6}Amazon GenAI, USA \quad
    \textsuperscript{\rm 7}AI Institute, University of South Carolina, USA
}

\usepackage{bibentry}
\begin{document}

\maketitle

\begin{abstract}
The rapid advancement of text-to-image generation systems, exemplified by models like Stable Diffusion, Midjourney, Imagen, and DALL-E, has heightened concerns about their potential misuse. In response, companies like Meta and Google have intensified their efforts to implement watermarking techniques on AI-generated images to curb the circulation of potentially misleading visuals. However, in this paper, we argue that current image watermarking methods are fragile and susceptible to being circumvented through visual paraphrase attacks. The proposed visual paraphraser operates in two steps. First, it generates a caption for the given image using KOSMOS-2, one of the latest state-of-the-art image captioning systems. Second, it passes both the original image and the generated caption to an image-to-image diffusion system. During the denoising step of the diffusion pipeline, the system generates a visually similar image that is guided by the text caption. The resulting image is a visual paraphrase and is free of any watermarks. Our empirical findings demonstrate that visual paraphrase attacks can effectively remove watermarks from images. This paper provides a critical assessment, empirically revealing the vulnerability of existing watermarking techniques to visual paraphrase attacks. While we do not propose solutions to this issue, this paper serves as a call to action for the scientific community to prioritize the development of more robust watermarking techniques. Our first-of-its-kind visual paraphrase dataset\footnote{https://tinyurl.com/58vf2aj5} and accompanying code\footnote{https://tinyurl.com/djt9j9jz} are 
publicly available. 
\end{abstract}



%\vspace{-10mm}
\begin{figure*}
\begin{center}
% \vspace{-13mm}
\includegraphics[width=0.8\textwidth]{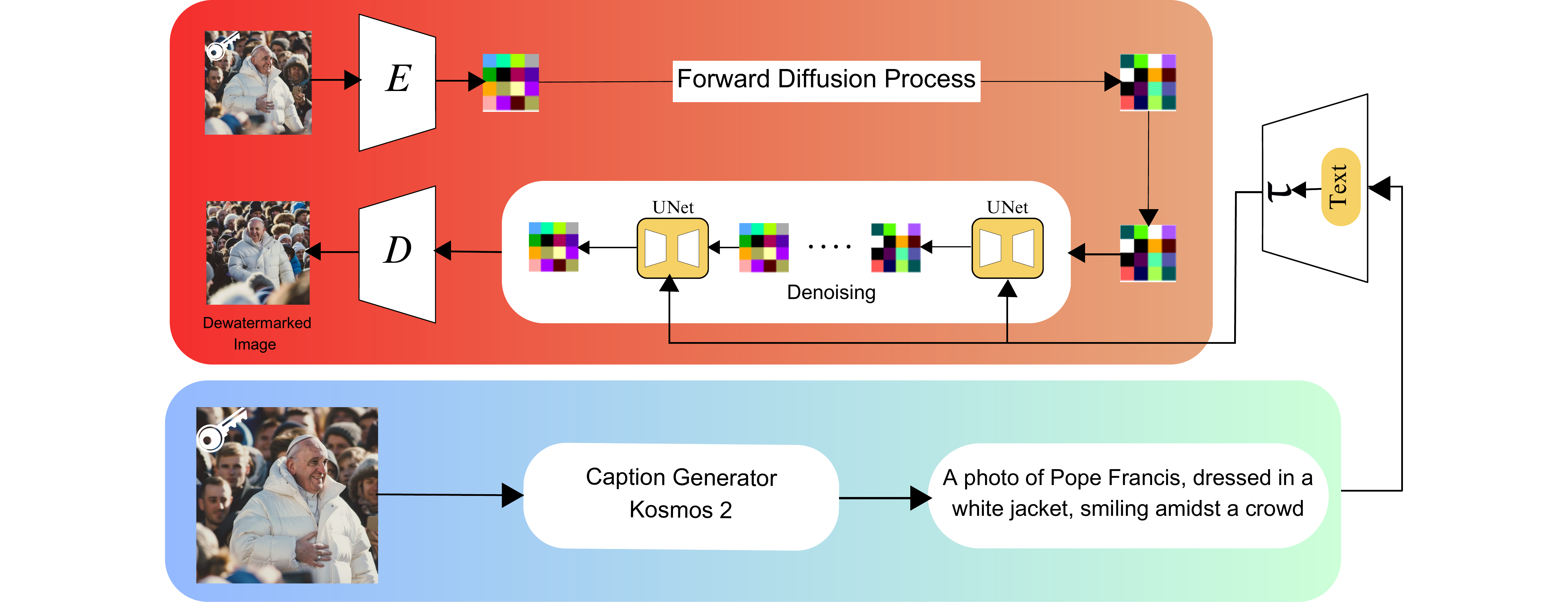}
% \vspace{-10mm}
%\caption{\cite{bommasani2023eu-ai-act}.}

\caption{The proposed visual paraphraser operates in two steps. First, it generates a caption for the given image using KOSMOS-2 \cite{peng2023kosmos2}. Second, it passes both the original image and the generated caption to an image-to-image diffusion system. During the denoising step of the diffusion pipeline, the system generates a visually similar image that is guided by the text caption. The resulting image is a visual paraphrase and is free of any watermarks.}
\vspace{-2mm}
\label{fig:vp-main-block}
\end{center}
\end{figure*}

\vspace{-3mm}
\section{Watermarking AI-Generated Images: The Necessity}
\label{sec:introduction}

\begin{figure}[ht!]
    \centering

    \begin{subfigure}{0.35\linewidth}
        \includegraphics[width=\linewidth]{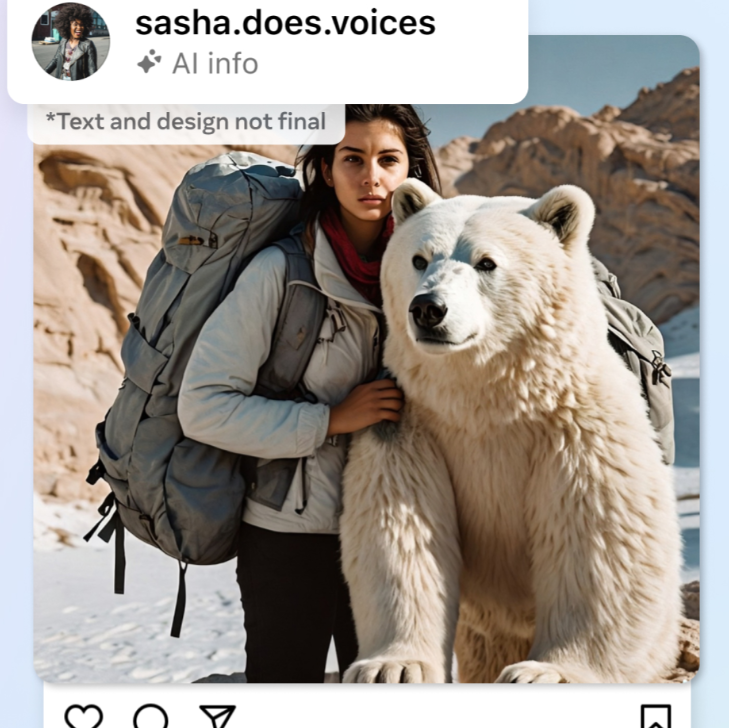}
        \caption{Original image with visible watermark patch}
    \end{subfigure} \hspace{1cm}
    \begin{subfigure}{0.35\linewidth}
        \includegraphics[width=\linewidth]{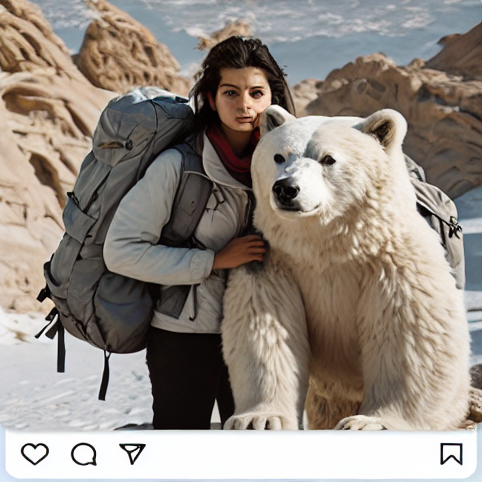}
        \caption{1\textsuperscript{st} mask-filled image}
    \end{subfigure}
    \begin{subfigure}{0.35\linewidth}
        \includegraphics[width=\linewidth]{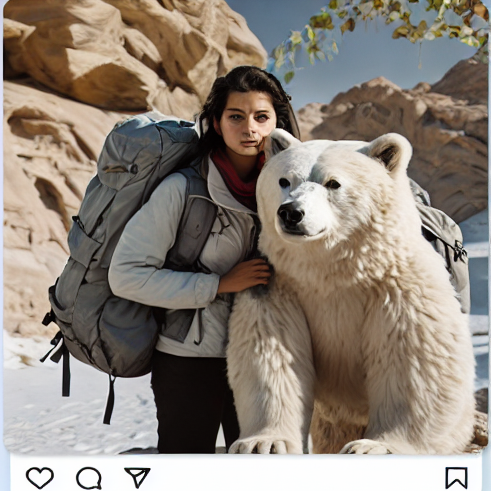}
        \caption{2\textsuperscript{nd} mask-filled image}
    \end{subfigure} \hspace{1cm}
    \begin{subfigure}{0.35\linewidth}
        \includegraphics[width=\linewidth]{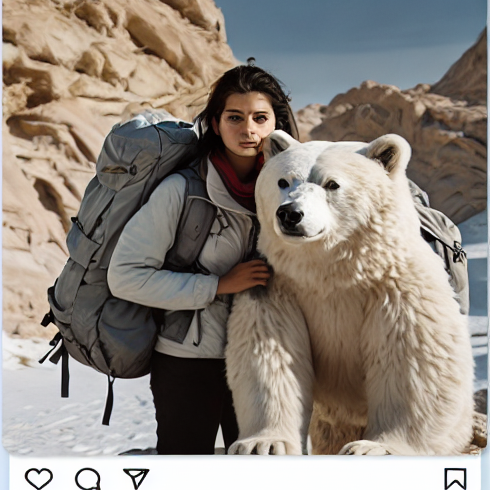}
        \caption{3\textsuperscript{rd} mask-filled image}
    \end{subfigure}

    \vspace{-2mm}
    \caption{Meta recently announced their strategies \cite{fbLabelingAIGenerated} to combat AI-generated misinformation, including a proposal to place visible markers on images. However, we argue that these visible markers are easily detectable and can be removed or altered using image inpainting techniques \cite{zeng2020highresolutionimageinpaintingiterative}, which involve reconstructing missing regions in an image. In Image (a), the original image from Meta's blog is shown, while Images (b), (c), and (d) demonstrate how image inpainting can generate different versions of the image with the markers effectively removed or replaced. Therefore, visible markers cannot be considered a reliable countermeasure in the era of generative AI.}
    \label{fig:visual marker removal}
\end{figure}

% Watermarking techniques originated within the computer vision community; however, recent advancements in LLMs have spurred interest in the development of text watermarking methods. Last year, OpenAI alluded to the development of watermarking techniques \cite{tc-gpt} for ChatGPT, although specific details were not disclosed. 

With the rapid proliferation of AI-generated visual content from models such as Stable Diffusion \cite{rombach2022highresolution, podell2023sdxl}, DALL-E \cite{ramesh2021zeroshot, ramesh2022hierarchicaltextconditionalimagegeneration}, Midjourney \cite{holzmidjourney}, Imagen \cite{saharia2022photorealistic}, among others, and their dangerous potential for misuse by malicious actors, the field of image watermarking has become a critical area of research. Given that, as of 2020, approximately 3.2 billion images and 720,000 hours of video are uploaded to social media platforms daily \cite{CONVERSATION2023}, the volume of visual content is staggering. When considering how AI-generated visuals can significantly contribute to misinformation strategies by serving as deceptive evidence for fabricated anomalies, the demand for robust watermarking techniques for AI-generated content becomes more pressing than ever. Governments worldwide have initiated discussions and implemented measures to develop policies concerning AI systems. The European Union \cite{euaiproposal} has taken a decisive step by enacting legislation, while the United States \cite{whitehouseaibill} and other countries have introduced preliminary proposals for a regulatory framework for AI. A primary concern among policymakers is that \textit{"Generative AI could act as a force multiplier for political disinformation. The combined effect of generative text, images, videos, and audio may surpass the influence of any single modality"} \cite{janjevagenai2023}. Moreover, AI policymakers have raised significant concerns regarding the use of automatic labeling or invisible watermarks as technical solutions to the challenges posed by generative AI-enabled disinformation. Nevertheless, persistent concerns remain about the susceptibility of these measures to deliberate tampering and the potential for malicious actors to circumvent them entirely.

In response to the increasing concern over AI-generated misinformation, companies such as Meta, Google, and OpenAI have begun exploring methods to watermark their generated image content. Meta recently announced its strategy \cite{ fernandez2023stablesignaturerootingwatermarks} to address AI-generated misinformation, emphasizing three primary approaches: (i) the inclusion of visible markers on images, (ii) the application of invisible watermarks, and (iii) the embedding of metadata within image files. This paper contends that these strategies are inadequate in the context of advanced generative AI systems. For example, with the rapid progression of image inpainting systems \cite{jeevan2023wavepaintresourceefficienttokenmixerselfsupervised, zheng2022cmganimageinpaintingcascaded, li2022matmaskawaretransformerlarge, wang2022zeroshotimagerestorationusing}, detecting and removing visible markers has become increasingly straightforward, as illustrated in Figure \ref{fig:visual marker removal}. Similarly, metadata, which comprises additional tags, can be easily stripped from files using a simple wrapper, as demonstrated in the detailed example provided in Appendix 7.1.

Watermarking techniques originated within the computer vision community; however, recent advancements in LLMs have spurred interest in the development of text watermarking methods. Last year, OpenAI alluded to the development of watermarking techniques \cite{tc-gpt} for ChatGPT, although specific details were not disclosed. \citet{kirchenbauer2023watermark} presented the first functional watermarking models for LLMs, albeit they were met with criticism. Furthermore, \citet{sadasivan2023aigenerated} illustrated that paraphrasing could effectively remove text watermarks. This prompted us to investigate the impact of visual paraphrase attacks on image watermarking. Though the term ``\emph{visual paraphrase attack}" is not yet widely recognized, we aim to formally introduce it to the community through this paper.

This paper exclusively critiques SoTA image watermarking techniques and empirically illustrates their brittleness towards visual paraphrase attacks. Figure \ref{fig:vp-main-block} illustrates the pipeline for generating visual paraphrases, wherein we encode and decode watermarked images to generate visually paraphrased dewatermarked outputs. Further details of the model are explained in Section \ref{sec:approach}. 
%To achieve this, the system employs a forward diffusion process, where the textual embeddings needed for image-to-image diffusion are generated using the Kosmos 2 \cite{peng2023kosmos2} caption generator. These embeddings are then processed through a denoising block, which constructs the visually paraphrased image. 
Through extensive experimentation, we aim to offer a comprehensive understanding of how visual paraphrasing can effectively remove watermarks from AI-generated images, emphasizing the urgent need for more robust and resilient watermarking strategies. 
%We seek to inspire the scientific community to prioritize the development of advanced techniques capable of withstanding the sophisticated manipulation capabilities introduced by modern generative models. 
Our contributions can be summarized as follows:

\begin{figure*}[ht!]
  \centering
  \resizebox{0.8\textwidth}{!}{%
    \begin{forest}
      forked edges,
      for tree={
        grow=east,
        reversed=true,
        anchor=base west,
        parent anchor=east,
        child anchor=west,
        base=center,
        font=\Huge,
        rectangle,
        draw=hidden-draw,
        rounded corners,
        align=center,
        text centered,
        minimum width=5em,
        edge+={darkgray, line width=2pt},
        s sep=8pt,
        inner xsep=2pt,
        inner ysep=3pt,
        line width=0.8pt,
        ver/.style={rotate=90, child anchor=north, parent anchor=south, anchor=center},
        leaf/.style={fill=#1, draw=black},
      },
      where level=1{minimum width=25em,font=\Huge,xshift=0.5em}{},
      where level=2{text width=14em,font=\Huge}{},
      where level=3{minimum width=10em,font=\Huge}{},
      where level=4{text width=26em,font=\Huge}{},
      where level=5{text width=20em,font=\Huge}{},
      [Watermarking Techniques, text width=25em, ver, for tree={fill=red!20!white}
        [Learning-based, text width=40em, for tree={fill=blue!20!white} 
          [The Stable Signature \cite{fernandez2023stable}, text width=40em, for tree={fill=magenta!40!white}, leaf]
          [Tree-Ring Watermark \cite{wen2023treering}, text width=40em, for tree={fill=magenta!40!white}, leaf]
          [ZoDiac \cite{zhang2024zodiac}, text width=40em, for tree={fill=magenta!40!white}, leaf]
          [SynthID \cite{deepmind_synthid}, text width=40em, for tree={fill=magenta!40!white}, leaf]
          [HiDDen \cite{zhu2018hidden}, text width=40em, for tree={fill=magenta!40!white}, leaf]
          [Gaussian Shading \cite{yang2024gaussian}, text width=40em, for tree={fill=magenta!40!white}, leaf]
        ]
        [Static, text width=40em, for tree={fill=green!20!white}
          [Visible \\Watermark, text width=40em, for tree={fill=yellow!50!white}
            [Channel Logo, text width=52em, for tree={fill=brown!50!white}, leaf]
            [Adaptive Visible Watermark \cite{kankanhalli1999adaptive}, text width=52em, for tree={fill=brown!50!white}, leaf]
          ]
          [Invisible \\Watermark, text width=40em, for tree={fill=orange!70!red!10!white}
            [IA-DCT \cite{podilchuk1998image}, text width=52em, for tree={fill=cyan!50!green!30!white}, leaf]
            [IA-W \cite{podilchuk1998image}, text width=52em, for tree={fill=cyan!50!green!30!white}, leaf]
            [Perceptual Image Watermark \cite{wolfgang1999perceptual}, text width=52em, for tree={fill=cyan!50!green!30!white}, leaf]
            [Psychovisual Digital Watermarking \cite{delaigle1998psychovisual}, text width=52em, for tree={fill=cyan!50!green!30!white}, leaf]
            [DwtDctSVD \cite{navas2008dwt}, text width=52em, for tree={fill=cyan!50!green!30!white}, leaf]
          ]
        ]
      ]
    \end{forest}
  }
  \vspace{-2mm}
  \caption{Watermarking techniques are generally classified into two categories: (i) static (i.e., non-learning) watermarking methods and (ii) learning-based (dynamic) watermarking methods. Static watermarking includes both invisible and visible types, while learning-based techniques represent the state-of-the-art. Although static watermarking techniques are mostly outdated and seldom used, we selected the latest method, DwtDctSVD \cite{dctdw}, for comparison. Other static methods are discussed solely for literature review purposes. Learning-based watermarking techniques are more modern, and we tested all the listed methods against visual paraphrase attacks.}
  \label{fig:taxonomy}
\end{figure*}
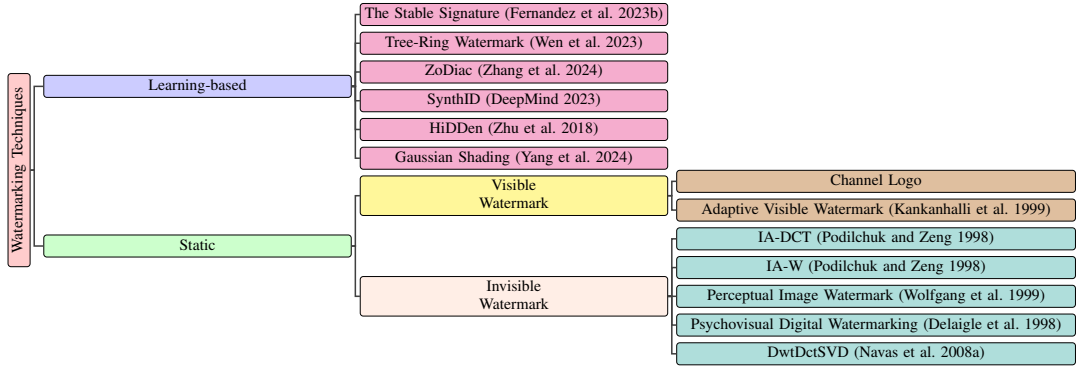

\vspace{-6mm}
 \begin{defin}
 \vspace{-2mm}
 \begin{itemize}
 [leftmargin=1mm]
 \setlength\itemsep{0em}
 \begin{spacing}{0.85}
 \vspace{1mm}
 
    \item[\ding{224}] {\footnotesize 
     {\fontfamily{phv}\fontsize{8}{9}\selectfont
    We introduce the concept of a ``\emph{visual paraphrase attack}" as a method to circumvent existing image watermarking techniques, emphasizing their inherent brittleness.
     } 
     \vspace{1mm}
     \item[\ding{224}] {\footnotesize 
     {\fontfamily{phv}\fontsize{8}{9}\selectfont We present empirical evidence demonstrating that visual paraphrasing attacks are effective against six of the most recent and SoTA watermarking techniques.
    }
     }
%\vspace{1mm}
%     \item {\footnotesize 
%     {\fontfamily{phv}\fontsize{8}{9}\selectfont We establish a framework for \textit{assessing the robustness of watermarking methods against various adversarial attacks}, specifically focusing on their susceptibility to de-watermarking through visual paraphrasing.
%     }
%     }    
\vspace{1mm}
\item[\ding{224}] {\footnotesize 
     {\fontfamily{phv}\fontsize{8}{9}\selectfont We call on the scientific community to prioritize the development of more robust watermarking techniques. Our proposed framework and dataset can serve as a benchmark for testing the robustness of new watermarking methods. 
     }
     }
%\item {\footnotesize 
%     {\fontfamily{phv}\fontsize{8}{9}\selectfont 
%     \vspace{1mm}
%We explore the \textit{impact of image-to-image diffusion processes on watermark integrity}, revealing the trade-offs between preserving image quality and maintaining watermark detection accuracy.
%     }
%     }     
     
\vspace{-5mm}}
\end{spacing}    
 \end{itemize}
\end{defin}

\vspace{-8mm}
\section{Related Work: State-of-the-art Image Watermarking and Detection Methods}

Watermarking techniques are broadly classified into two categories: (i) static (i.e., learning-free) watermarking methods and (ii) learning-based watermarking methods. Static watermarking refers to embedding a watermark into an image in a fixed, unchanging manner. Once the watermark is embedded, it remains the same regardless of any subsequent use or manipulation of the image. Dynamic watermarking, on the other hand, refers to a more flexible approach where the watermark can change or adapt based on certain conditions or during the image's usage. This type of watermarking is often used in scenarios where the watermark needs to convey additional information, such as the time of access, user identity, or location, and can be embedded in real-time. Dynamic watermarking can be more difficult to detect and remove because the watermark isn't static or predictable.

\subsection{Static Watermarking Methods}
% Static watermarking techniques are methods used to embed a permanent, unchangeable watermark into digital content like images, videos, audio, or text. 
% These watermarks are often imperceptible to human senses but detectable by algorithms. The watermark is embedded during content creation or distribution and remains constant regardless of how the content is used or shared. Some of the methods used, include Least Significant Bit (LSB) insertion, Discrete Cosine Transform (DCT), Spread Spectrum techniques and Patchwork. These techniques ensure that the watermark remains embedded even after typical modifications such as resizing, compression, etc. 
The most common way of creating a static watermark is to apply some type of Frequency domain transform and then altering certain frequency coefficients of the image or its image blocks via adding a bit of the watermark. The watermarked image is obtained via inverse transform of this transformed image.
like Discrete Wavelet Transform (DWT) \cite{lai2010digital} to decompose an image into several frequency sub-bands, then applying another transform like the Discrete Cosine Transform (DCT) \cite{yuan2020new} to each block of some of the sub-bands, and finally altering certain frequency coefficients of each block via adding a bit of the watermark. The watermarked image is obtained via inverse transform.  
We won't study these methods further in this work due to these approaches being extremely easy to detect and very outdated, except for DwtDctSVD \cite{dctdw}, included solely for academic comparison.
\subsubsection{DwtDctSVD}
The DwtDctSVD \cite{dctdw} watermarking algorithm uses various techniques to embed a watermark into an image, including Discrete Wavelet Transform (DWT), Discrete Cosine Transform (DCT), and Singular Value Decomposition (SVD). These methods decompose the image into frequency bands, allowing the watermark to be embedded in specific regions that are less prone to common image processing operations. The watermark is embedded in middle-frequency bands to balance robustness and imperceptibility. However, the watermark can be removed or degraded by manipulating the target frequency bands through filtering or compression, altering the singular values obtained from SVD, or applying visual paraphrasing techniques such as random pixel swapping or contrast changes. These methods can destroy or weaken the watermark, rendering it less effective or totally removed.

\subsection{Learning-based Watermarking Methods}
%Here encoders and decoders \cite{8945866} are neural networks and learn via back-propagation.
A typical learning based watermarking method has three key components: watermark ($w$), encoder ($E$), and decoder ($D$). An encoder takes an image $X$ and watermark $w$ as inputs and produces an watermarked image $(X_w)$. So, $X_w = E(X, w)$ and a decoder takes $X_w$ as an input and produces $\hat{w} = D(X_w)$. $\hat{w}_i = [ \hat{w}_i \geq \tau ]$, where $[\cdot]$ represents the indicator function and $\tau$ is a threshold value we decide based on the problem requirements.

The following paragraphs describe the five state of the art learning based watermarking techniques we selected for comparison with visual paraphrasing.

\subsubsection{HiDDen: A Watermarking Method for Images}
The HiDDen paper \cite{zhu2018hidden}  proposes a watermarking technique where an encoder embeds a secret message into a cover image, which is then noised and decoded to retrieve the message. To ensure robustness, the encoder and decoder are trained to minimize losses related to image similarity, message accuracy, and adversarial detection. However, the method has weaknesses that can be exploited, such as the noise layer's impact on the encoded message and the complexity of balancing multiple loss functions. Visual paraphrasing, which alters the image while preserving its semantic content, can manipulate these weaknesses to distort the encoded message or make the watermark undetectable.
% The HiDDen paper \cite{zhu2018hidden} introduces a watermarking technique where an encoder \(E\) takes a secret message \(M\) and a cover image \(I_{co}\) as inputs, producing an encoded image \(I_{en}\). This encoded image is then subjected to a noise layer \(N\), resulting in a noised image \(I_{no}\). The decoder retrieves a predicted message from the noised image. To ensure robustness, an adversary model is trained to detect whether an image has been encoded. The encoder and decoder are trained jointly to minimize three types of losses: \(L_I\), the difference between the cover and encoded images; \(L_M\), the difference between the input and predicted messages; and \(L_G\), which reflects the encoded image being detected by the adversary.

% Despite its robustness, the HiDDen watermarking method has fragile factors that can be exploited. The noise layer introduces variability that can affect the integrity of the encoded message. Additionally, the joint optimization of multiple loss functions presents a complex balancing act, which could lead to suboptimal encoding under certain conditions. Visual paraphrasing, by utilizing image-to-image diffusion, can potentially affect these fragile factors. By altering the image while preserving its semantic content, visual paraphrasing can introduce changes that distort the encoded message or make the watermark undetectable to the adversary. This manipulation leverages the inherent weaknesses in the HiDDen method's reliance on maintaining the integrity of the encoded message through noise and adversarial detection.

\subsubsection{Stable Signature}
The Stable Signature method \cite{fernandez2023stable} 
 introduces a novel watermarking technique for images generated by latent diffusion models (LDMs) \cite{rombach2022high}, building on the process of progressively denoising a latent image representation. Watermarking is achieved by subtly modifying this latent representation in a way that remains invisible to the human eye but can be detected by a pretrained watermark extractor network. The core of the technique lies in refining the LDM decoder to produce images that exhibit a specific signature when analyzed by the watermark extractor. This involves minimizing a loss function that balances the reconstruction loss, which measures the difference between the generated and target images, and the watermark loss, which gauges the discrepancy between the generated image's signature and the desired watermark signature, controlled by a hyperparameter $\lambda$ \cite{gower2019sgd}. The method employs both standard training using SGD and adversarial training to enhance robustness against post-processing.

\subsubsection{Tree Ring Watermark} 
The proposed tree-ring watermarking technique \cite{wen2023treering} embeds a watermark into the frequency domain of the initial noise vector using Fast Fourier Transform (FFT) \cite{heckbert1995fourier}, followed by a diffusion process. To detect the watermark, the inverse diffusion process is applied, and an Inverse Fast Fourier Transform (IFFT) \cite{heckbert1995fourier} is performed. The L1 distance between the inverted noise vector and the key in the Fourier space is then compared to determine if the image is watermarked. Any attempts to disrupt the watermark through frequency manipulation or adversarial attacks result in loss of image details, rendering the image unusable. This approach aims to retain the image's essence while allowing for changes to the pixel values, similar to paraphrasing in text.
% The proposed tree-ring watermarking \cite{wen2023treering} technique involves embedding the watermark into the frequency domain of the initial noise vector using Fast Fourier Transform (FFT)\cite{heckbert1995fourier}, followed by a diffusion process applied to the watermarked latent image. To ascertain whether an image has been watermarked, we utilize the inverse diffusion process to reconstruct the latent image, subsequently performing an Inverse Fast Fourier Transform (IFFT)\cite{heckbert1995fourier}. By comparing the L1 distance between the inverted noise vector and the key in the Fourier space of the watermarked area, we determine if the image is watermarked. Any attempt at frequency manipulation or adversarial attack to disrupt this watermark results in loss of image details, rendering the image unusable as shown in figure \ref{fig:tree ring quality} Thus, akin to paraphrasing in text, we explore an approach aiming to retain the image's essence while not necessarily utilizing the same pixel values.

\subsubsection{ZoDiac Watermarking}
ZoDiac \cite{zhang2024zodiac} is a zero-shot watermarking technique that utilizes pre-trained diffusion models to embed watermarks into images while maintaining visual similarity. The method consists of three main steps: initializing a trainable latent vector using the DDIM inversion process \cite{song2022denoising} to reproduce the original image, encoding a concentric ring-like watermark into the latent vector's Fourier space and refining it using a custom reconstruction loss, and adaptively enhancing the visual quality of the watermarked image by mixing it with the original image to meet a desired quality threshold. Unlike tree-ring watermarking, ZoDiac can be used to watermark existing images, making it a versatile and effective watermarking technique.
% ZoDiac\cite{zhang2024zodiac} is a zero-shot watermarking technique that leverages pre-trained diffusion models to embed watermarks into images while maintaining visual similarity between the watermarked and original images. The method comprises three main steps:

% \begin{enumerate}[label=\Roman*.]
%     \item \textbf{Latent Vector Initialization}: A trainable vector is initialized to be used by the Stable Diffusion model to reproduce the original image \(x_0\). The original image \(x_0\) undergoes the DDIM \cite{song2022denoising} inversion process to generate the latent vector \(Z_T\).
%     \item \textbf{Watermark Encoding}: The latent vector \(Z_T\) is transformed into its Fourier space, where a concentric ring-like watermark is embedded. This watermark is similar to the one used in tree-ring watermarking. To ensure that the final watermarked image \(\hat{x}_0\) closely resembles the original image, ZoDiac iteratively refines the latent vector \(Z_T\) using a custom reconstruction loss.
%     \item \textbf{Adaptive Image Enhancement}: Once the watermarked image \(\hat{x}_0\) is generated, its visual quality is enhanced by adaptively mixing it with the original image \(x_0\) to meet a desired image quality threshold.
% \end{enumerate}
% Unlike tree-ring watermarking, ZoDiac can be used to watermark existing images.

\subsubsection{Gaussian Shading}
The Gaussian Shading watermarking method \cite{yang2024gaussian} offers a performance-lossless approach to embedding watermarks in images generated by diffusion models by operating entirely within the latent space, preserving the statistical distribution of latent representations. The process involves randomizing the watermark $W$ using a stream cipher like ChaCha20 \cite{bernstein2008chacha} to create an encrypted watermark $W'$, which is then embedded into the latent space $z$ during the diffusion process through the equation $z' = z + \sigma \cdot W'$, where $\sigma$ is a scaling factor. This technique ensures that the quality of watermarked images is indistinguishable from non-watermarked ones, supporting high watermark capacity and robustness against attacks such as noise and lossy compression.
% The Gaussian Shading \cite{yang2024gaussian} watermarking method presents a novel, performance-lossless approach to embedding watermarks in images generated by diffusion models. Unlike conventional methods that modify the image directly or require fine-tuning of model parameters, Gaussian Shading operates entirely within the latent space of the diffusion process. The method preserves the statistical distribution of latent representations, ensuring that the quality of watermarked images is indistinguishable from non-watermarked ones.

% The watermark $W$ is first randomized using a stream cipher, $S$, such as ChaCha20\cite{bernstein2008chacha}, to create a securly encrypted watermark $W'$. This randomized watermark is then embedded into the latent space $z$ during the diffusion process. The watermarking process is governed by the  equation $z' = z + \sigma W'$, where $z'$ represetn the watermarked latent representation, and $\sigma$ is a scaling factor that ensures the watermark blends seamlessly with the latent space distribution. 
% % \begin{equation}
% %     z' = z + \sigma W'
% % \end{equation}

% By maintaining the Gaussian distribution of latent representations, the Gaussian Shading method ensures that the image generation process remains unaffected, resulting in high-quality, watermark-embedded images. The method supports a high watermark capacity (e.g., 256 bits) and claims to be robust against various attacks, including noise and lossy compression.

%\subsubsection{SynthID}
In addition to the six techniques previously mentioned, the method proposed in \citet{deepmind_synthid} appears promising. However, due to the unavailability of its code, we are unable to include it in our study.

%is a toolkit developed by Google DeepMind that embeds a robust and imperceptible watermark into AI-generated content, including images, videos, text, and audio. The watermark is created by training two neural networks: one to modify pixels and create a unique pattern, and another to detect this pattern even after post-processing edits. However, due to the unavailability of its code, benchmarking and comparing SynthID with other watermarking techniques is not feasible, and therefore, it is omitted from this study.
% SynthID \cite{deepmind_synthid} is a toolkit from Google DeepMind that watermarks AI-generated content. It utilizes a data-driven watermarking approach, embedding an imperceptible mark during AI-generated content (AIGC) creation. This mark, robust to post-processing edits, persists across different modalities like images, videos, text, and audio. It watermarks AI-generated images by training two neural networks: one to imperceptibly modify pixels, creating a unique pattern, and another to detect this pattern even after edits, ensuring the image can be identified as AI-made despite manipulations.

% However, benchmarking SynthID is not feasible due to the unavailability of its code. This watermarking technique is internally developed by Google for all their AI-generated content, including images, and they have not disclosed the code or provided an API for external testing. Therefore, comparisons with SynthID are omitted from this study, as we lack the necessary access to implement or evaluate its performance. 

\begin{figure*}[ht]
    \centering
    \resizebox{0.75\linewidth}{!}{
        \begin{tabular}{l}
            \begin{minipage}{\linewidth}\centering 
                \includegraphics[width=\linewidth]{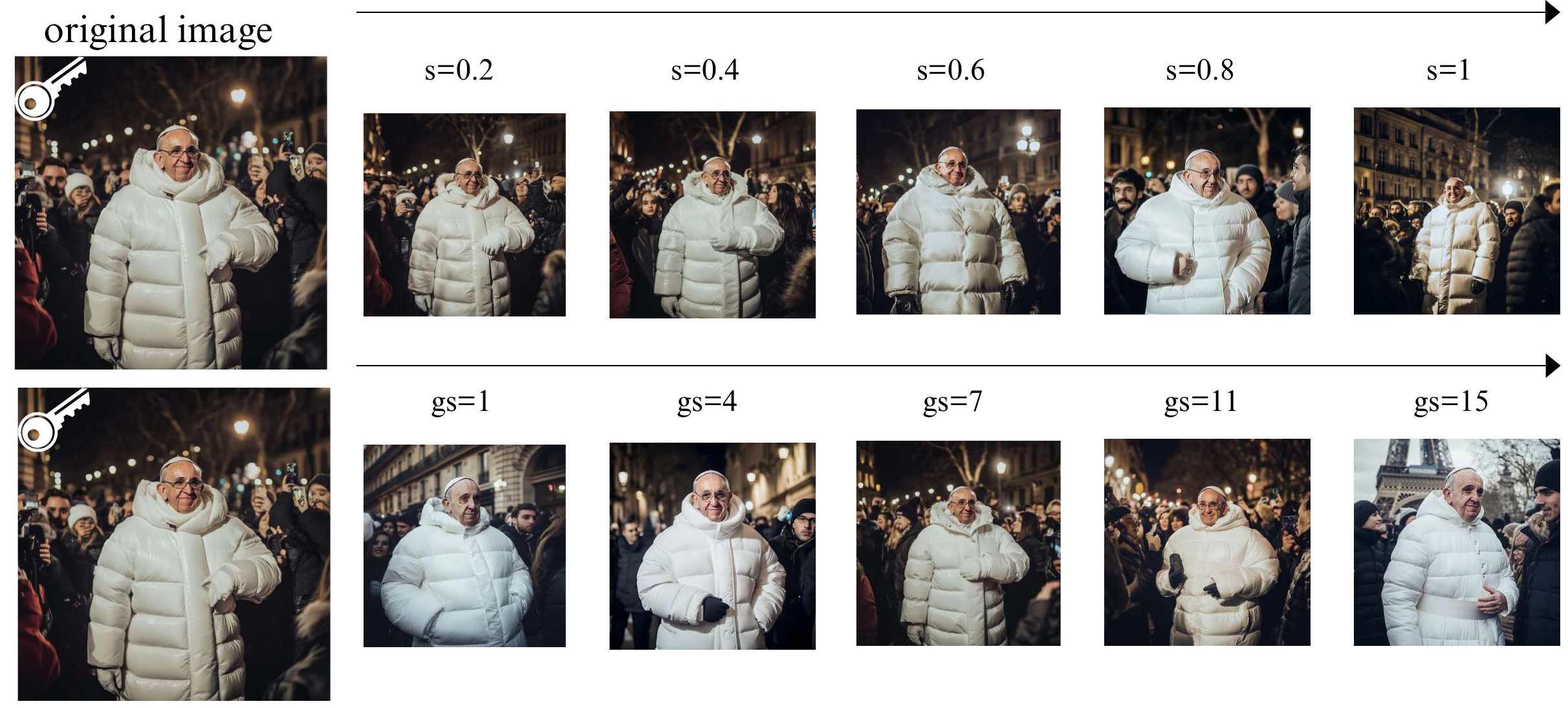}
                \footnotesize % Change \medium to \footnotesize or another appropriate size
                Prompt: Pope Francis, dressed in a white puffer jacket, surrounded by a small crowd before Christmas in Paris
            \end{minipage}
        \end{tabular}
    }
    \vspace{-1mm}
    \caption{Impact of paraphrasing strength ($s$) and guidance scale ($gs$) on Visual Paraphrasing: A higher strength $s$ allows for greater deviation from the original image, while a lower strength preserves more of the original details. The guidance scale $gs$ controls adherence to the text prompt, with higher values enforcing closer alignment to the prompt and lower values permitting more creative variations.}
    \label{fig:strength_var}
    \vspace{-2mm}
\end{figure*}

\subsection{Traditional De-Watermarking Techniques}
In addition to the discussed watermarking methods, certain traditional image alteration techniques can also function as de-watermarking attacks, as explored by previous researchers. We have included the following techniques in our study for comparison purposes.

\noindent
\textbf{Brightness}: 
%\vjil{Add citation: https://arxiv.org/abs/0909.3554} 
Altering the brightness \cite{verma2009robustnessdigitalimagewatermarking} of an image is a simple yet effective method for attempting to reduce the visibility of watermarks. By increasing or decreasing the brightness, the contrast between the watermark and the underlying image can be diminished, making the watermark less noticeable. However, this method can also degrade the overall quality of the image, potentially affecting important visual details. For our experiments, we selected a brightness level increased by a factor of 2.

\noindent
\textbf{Rotation}:
%\vjil{Add citation: https://arxiv.org/pdf/2206.10813} 
Rotating \cite{luo2022lecalearnedapproachefficient} an image is another technique used to obscure watermarks, especially those that are positioned in a fixed location. By rotating the image, the watermark may be repositioned to an area where it is less visible or more easily cropped out. While rotation can effectively reduce watermark visibility, it can also distort the original image content, particularly if the rotation angle is significant. For our experiments, the images were rotated by ±45\textdegree.

\noindent
\textbf{JPEG Compression}: 
%\vjil{Add citation: https://arxiv.org/abs/2108.08211} 
JPEG compression \cite{Jia_2021} is a common technique that reduces the file size of an image by discarding some of its data, which can incidentally affect the visibility of watermarks. The lossy nature of JPEG compression can blur or distort the watermark, making it less discernible. However, this technique may also lead to a loss of image quality, particularly when high compression levels are used. For our experiments, we set the quality setting to a reduced level of 50.

\noindent
\textbf{Gaussian Noise}: 
%\vjil{cite https://arxiv.org/html/2407.01301v1}
Adding Gaussian noise \cite{li2024gaussianstegogeneralizablestenographypipeline} to an image is a method that introduces random variations in pixel intensity, which can help in reducing the clarity of watermarks. The noise can obscure the fine details of the watermark, blending it into the background. While this approach can be effective, it may also degrade the visual quality of the image, making it appear grainy or less sharp. In our experiments, noise with a standard deviation of 0.05 was added to the images.

\section{Visual Paraphrasing} \label{sec:approach}

Paraphrasing is a well-established area of research within natural language processing (NLP). For instance, sentences such as ``What is your age?" and ``How old are you?" convey identical meanings despite their differing linguistic structures, thus constituting paraphrases of each other. In contrast, the concept of visual paraphrasing has not been as extensively explored, likely due to the recent emergence of text-to-image generation systems such as Stable Diffusion and Midjourney. These systems are capable of producing slight variations of a given image that maintain the same semantic content while differing in visual presentation. A related concept is visual entailment, which concerns image-sentence pairs where the image serves as the premise, as opposed to a sentence in traditional Visual Entailment tasks \cite{xie2019visual}. The objective in visual entailment is to determine whether the image semantically supports the text. However, given the significant differences between visual entailment and visual paraphrasing, this discussion will not explore visual entailment further. For example, as illustrated in Figure \ref{fig:strength_var}, all generated images are visual paraphrases of the input image. 

The process of visual paraphrasing begins with the generation of a caption for the image, followed by the application of image-to-image diffusion techniques. This two-step approach ensures that the output images retain the semantic integrity of the original while allowing for variations in visual presentation. The effectiveness of visual paraphrasing is governed by adjusting two key parameters: paraphrase strength and guiding scale, as described below.

\subsubsection{Generating Caption}
When an image encountered in the wild is suspected to have been generated by AI, the original prompt used to create it is typically unavailable. To address this challenge, we employed KOSMOS 2 \cite{peng2023kosmos2} to generate a textual description or a brief caption of the image. KOSMOS 2, along with other image captioning models \cite{you2016image}, is particularly effective at producing detailed textual descriptions of images. This generated caption then serves as the textual conditioning input for the image-to-image diffusion models, which are discussed in the following section. By utilizing the extracted textual context as guidance, the diffusion model reconstructs the image while preserving its semantic content, thereby achieving visual paraphrasing.

%Conversely, in the Black Box scenario, access to the original prompt is unavailable, necessitating an alternative approach to visual paraphrasing. In this context, the image is passed through an image caption generator, such as KOSMOS 2 \cite{peng2023kosmos2}, to obtain a textual representation of its content. The generated caption serves as text conditioning for the watermarked image, facilitating its processing through image-to-image diffusion models. By utilizing the extracted textual context as a guiding force, the diffusion model reconstructs the image while preserving its semantic content, effectively achieving visual paraphrasing even in the absence of direct access to the original prompt.

\subsubsection{Image-to-Image Diffusion}
At the core of visual paraphrasing lies the image-to-image diffusion process \cite{1021076}. This technique, employed in generative models, transforms images while maintaining their underlying structure and semantic information. The diffusion process involves two key stages: the forward diffusion process and the reverse diffusion process. In the forward diffusion process, an image is gradually corrupted by adding noise, eventually reaching a state of complete noise. Mathematically, this process is described as follows: $x_t = \sqrt{\alpha_t} x_{t-1} + \sqrt{1 - \alpha_t} \epsilon_t,$, where \( x_t \) is the image at time step \( t \), \( \alpha_t \) is a noise scaling factor, and \( \epsilon_t \) is the noise sampled from a Gaussian distribution. In the reverse diffusion process, the model attempts to remove the noise step by step, reconstructing the original image from the noisy version. This is achieved using a learned denoising function \( \epsilon_\theta \): $x_{t-1} = \frac{1}{\sqrt{\alpha_t}} \left( x_t - \sqrt{1 - \alpha_t} \epsilon_\theta(x_t, t) \right)$. This iterative denoising continues until the model produces an image that closely resembles the original in both visual and semantic terms. In this context, two controls are utilized: (i) the original image and (ii) the generated caption. The number of inference steps, denoted by \( T \), is a critical factor in this process. Increasing the number of steps generally results in more refined reconstructions, yielding higher-quality images, albeit with greater computational demands. In this scenario, we employed the default setting of 50 inference steps.

% \begin{figure*}[ht!]
% \centering
%     \scriptsize
%     \newcommand{\imwidth}{0.24\textwidth}
%     \resizebox{\linewidth}{!}{
%     \begin{tabular}{cccc}
%     \toprule
%     \multicolumn{2}{c}{\textbf{Tree Ring}} & \multicolumn{2}{c}{\textbf{Stable Signature}} \\
%     \midrule
%        \includegraphics[width=\imwidth]{img/variation_plots/tr_guidance_scale_vs_cmmd_gs.png} &
%        \includegraphics[width=\imwidth]{img/variation_plots/tr_guidance_scale_vs_det_gs.png} &
%        \includegraphics[width=\imwidth]{img/variation_plots/stable_guidance_scale_vs_cmmd_gs.png} &
%        \includegraphics[width=\imwidth]{img/variation_plots/stable_guidance_scale_vs_det_gs.png}  \\

%        \includegraphics[width=\imwidth]{img/variation_plots/tr_strength_vs_cmmd_s.png} &
%        \includegraphics[width=\imwidth]{img/variation_plots/tr_strength_vs_det_s.png} &
%        \includegraphics[width=\imwidth]{img/variation_plots/stable_strength_vs_cmmd_s.png} &
%        \includegraphics[width=\imwidth]{img/variation_plots/stable_strength_vs_det_s.png} \\
%        \bottomrule
%     \end{tabular}
%     }
    
% \caption{This figure shows the variation of CMMD \cite{jayasumana2024rethinking} and detectability of visual paraphrases with respect to strength and guidance scale. The images were watermarked using Tree Ring Watermarking \cite{wen2023treering} and Stable Signature \cite{fernandez2023stable}.}
% \label{fig:strength_gs_plots}
% \end{figure*}

\begin{figure*}[ht!]
\centering
    \scriptsize
    \newcommand{\imwidth}{0.24\textwidth}
    \resizebox{0.76\linewidth}{!}{
    \begin{tabular}{cccc}
    \toprule
    \multicolumn{2}{c}{\textbf{Tree Ring}} & \multicolumn{2}{c}{\textbf{Stable Signature}} \\
    \midrule
       \includegraphics[width=\imwidth]{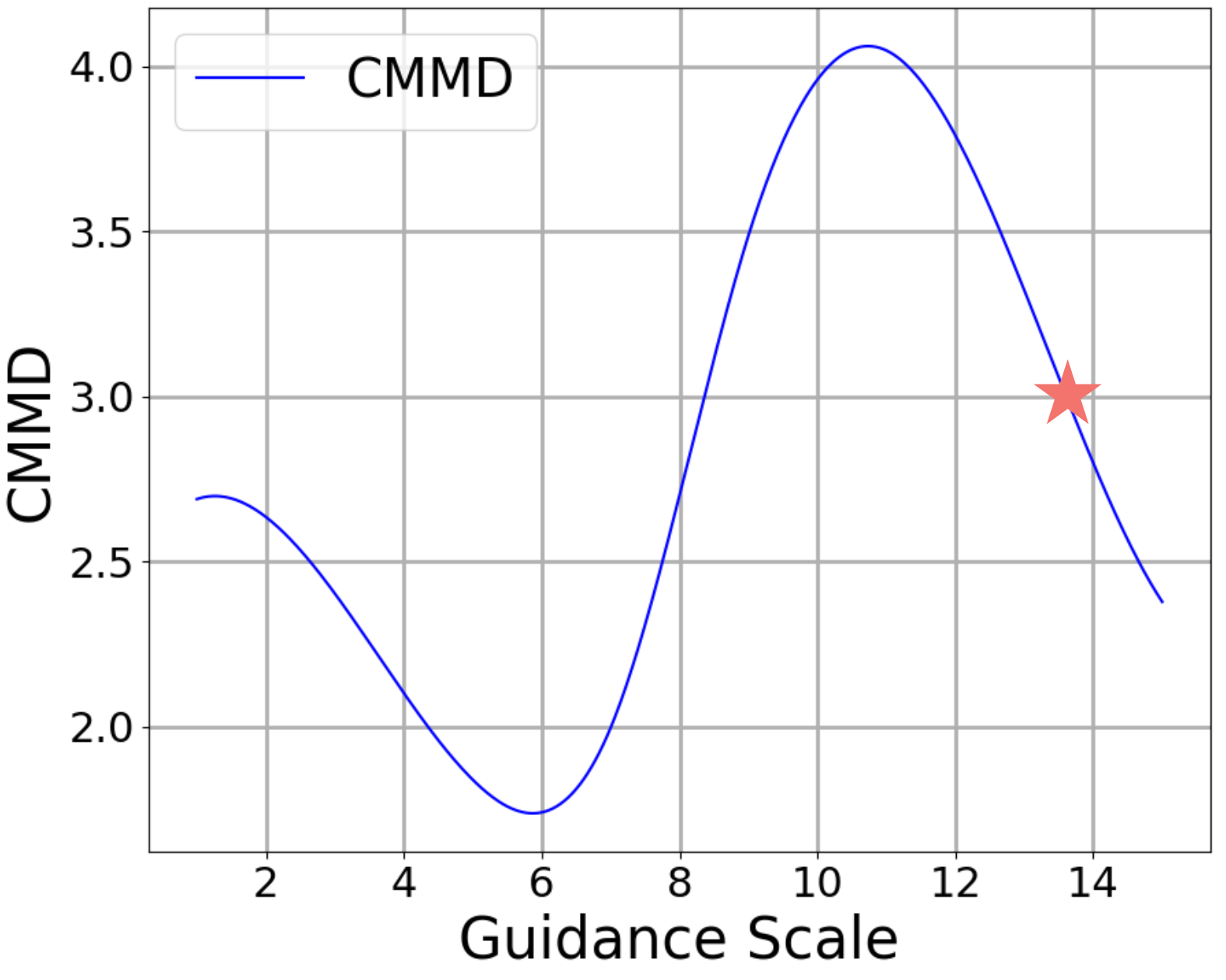} &
       \includegraphics[width=\imwidth]{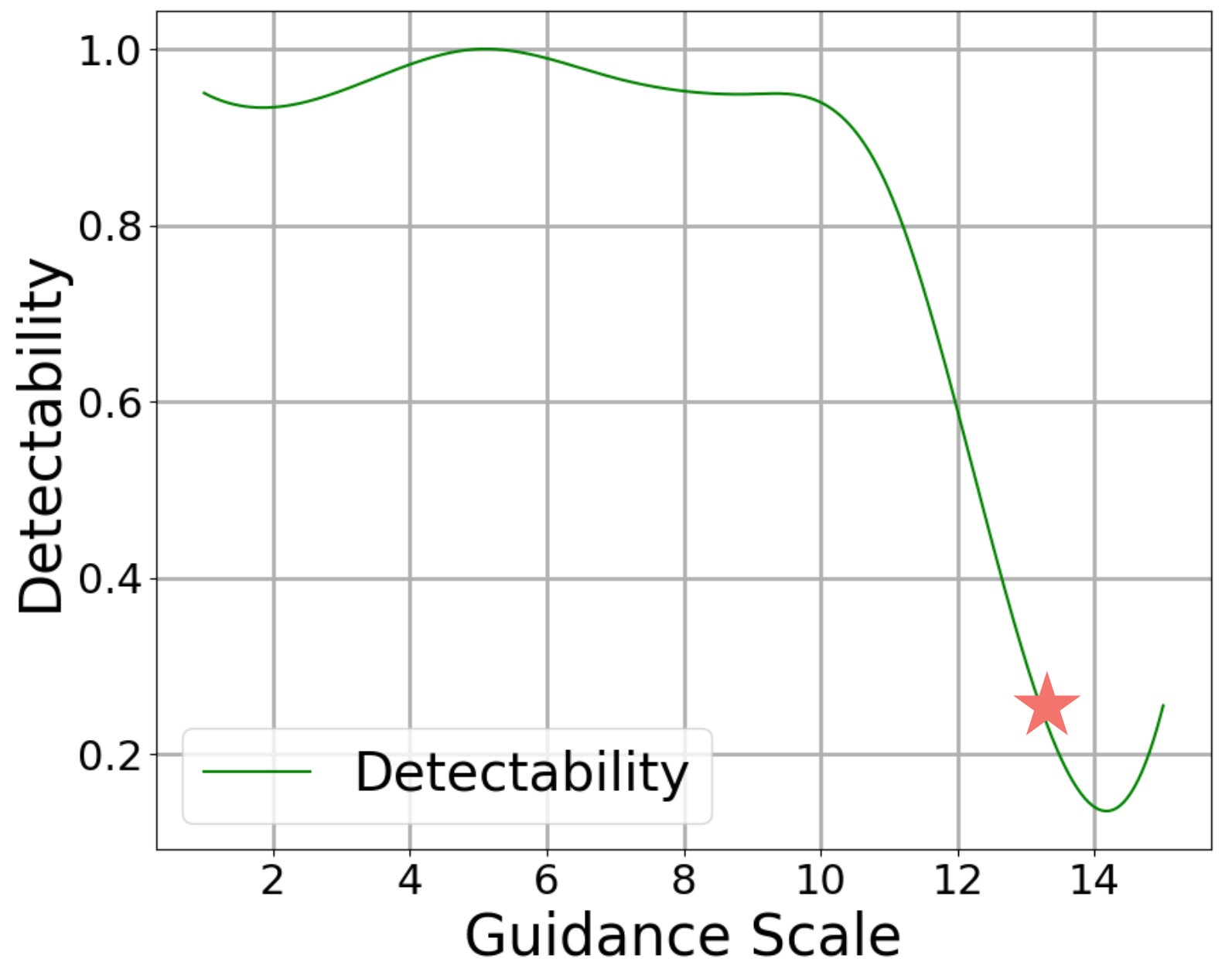} &
       \includegraphics[width=\imwidth]{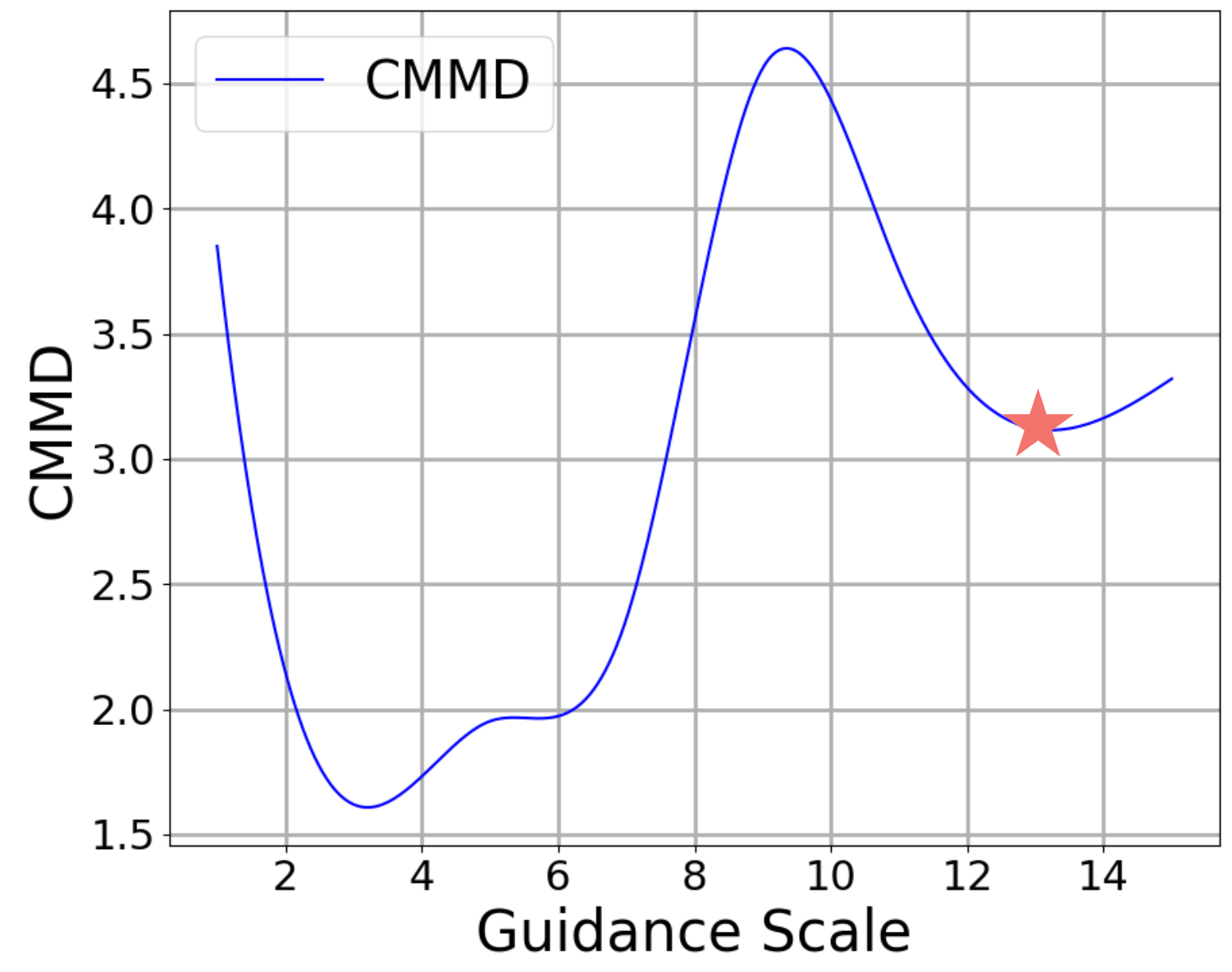} &
       \includegraphics[width=\imwidth]{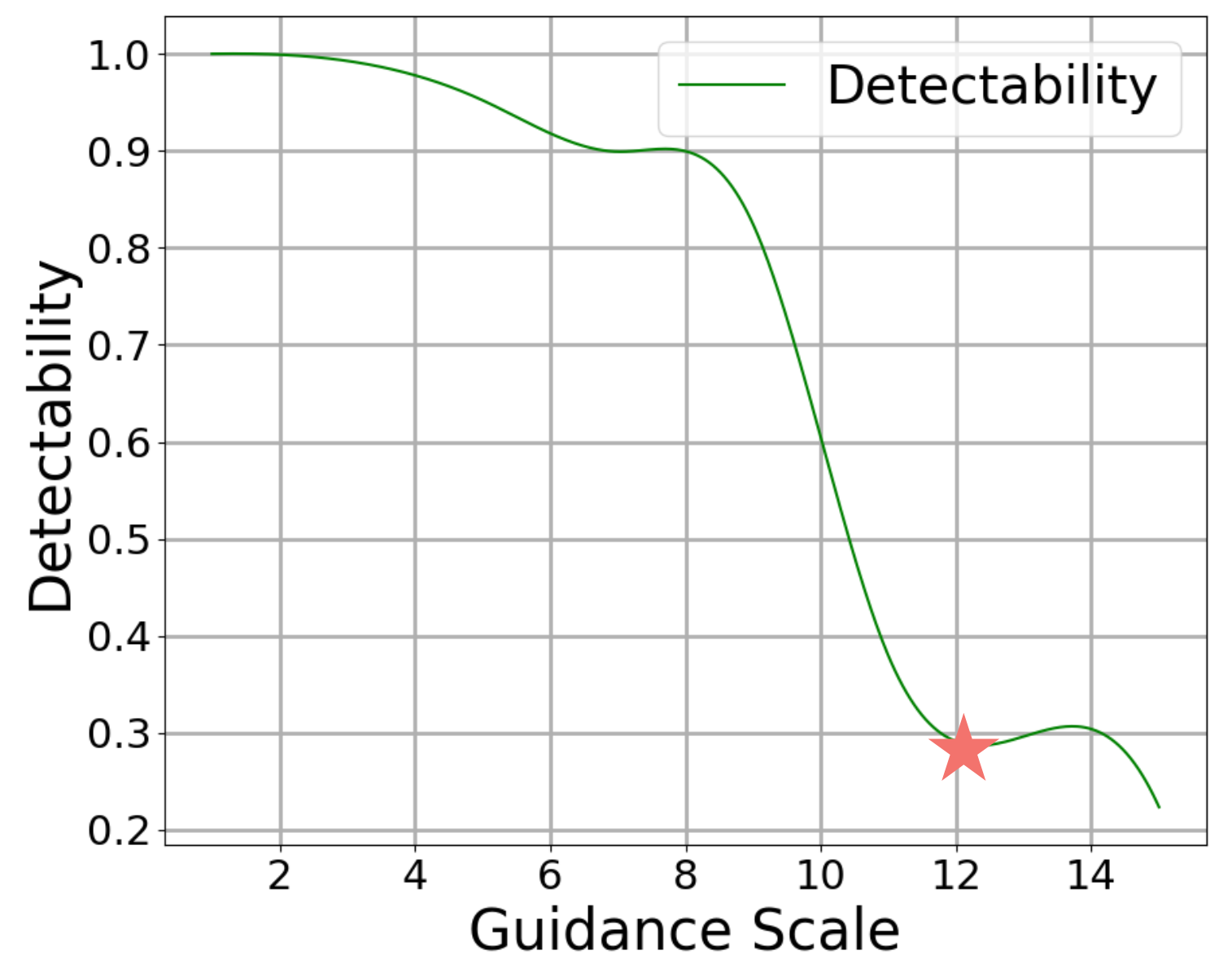}  \\

       \includegraphics[width=\imwidth]{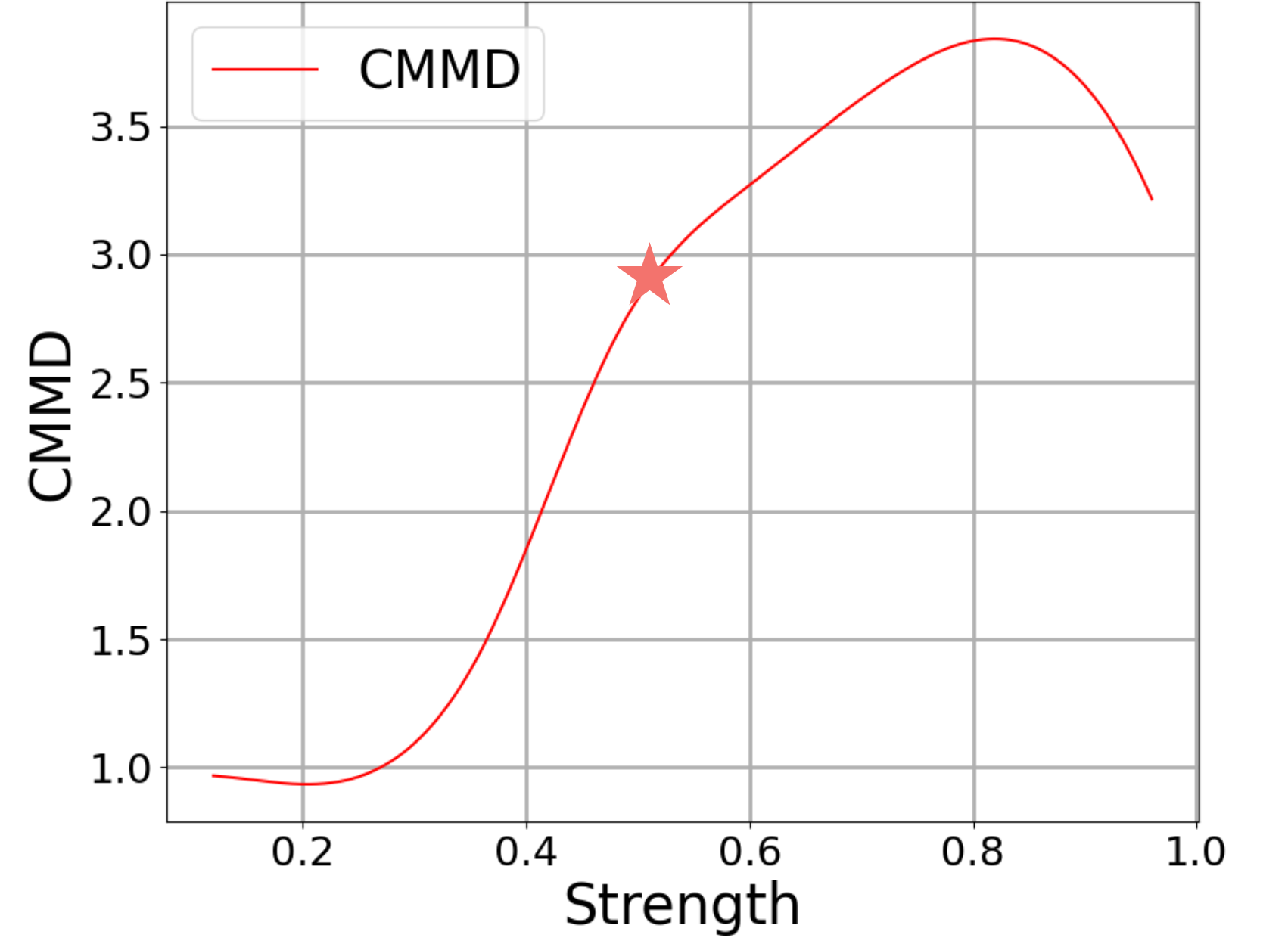} &
       \includegraphics[width=\imwidth]{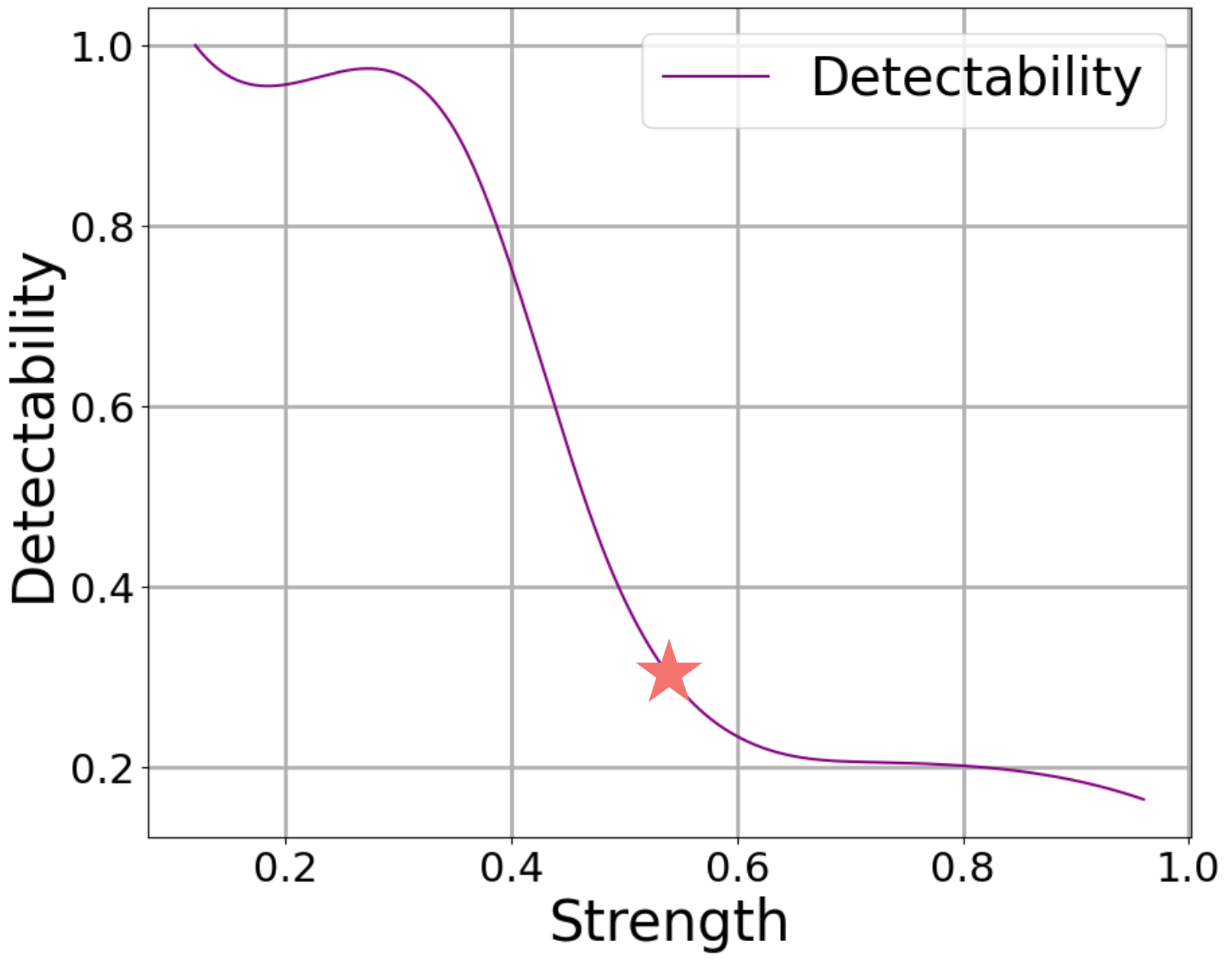} &
       \includegraphics[width=\imwidth]{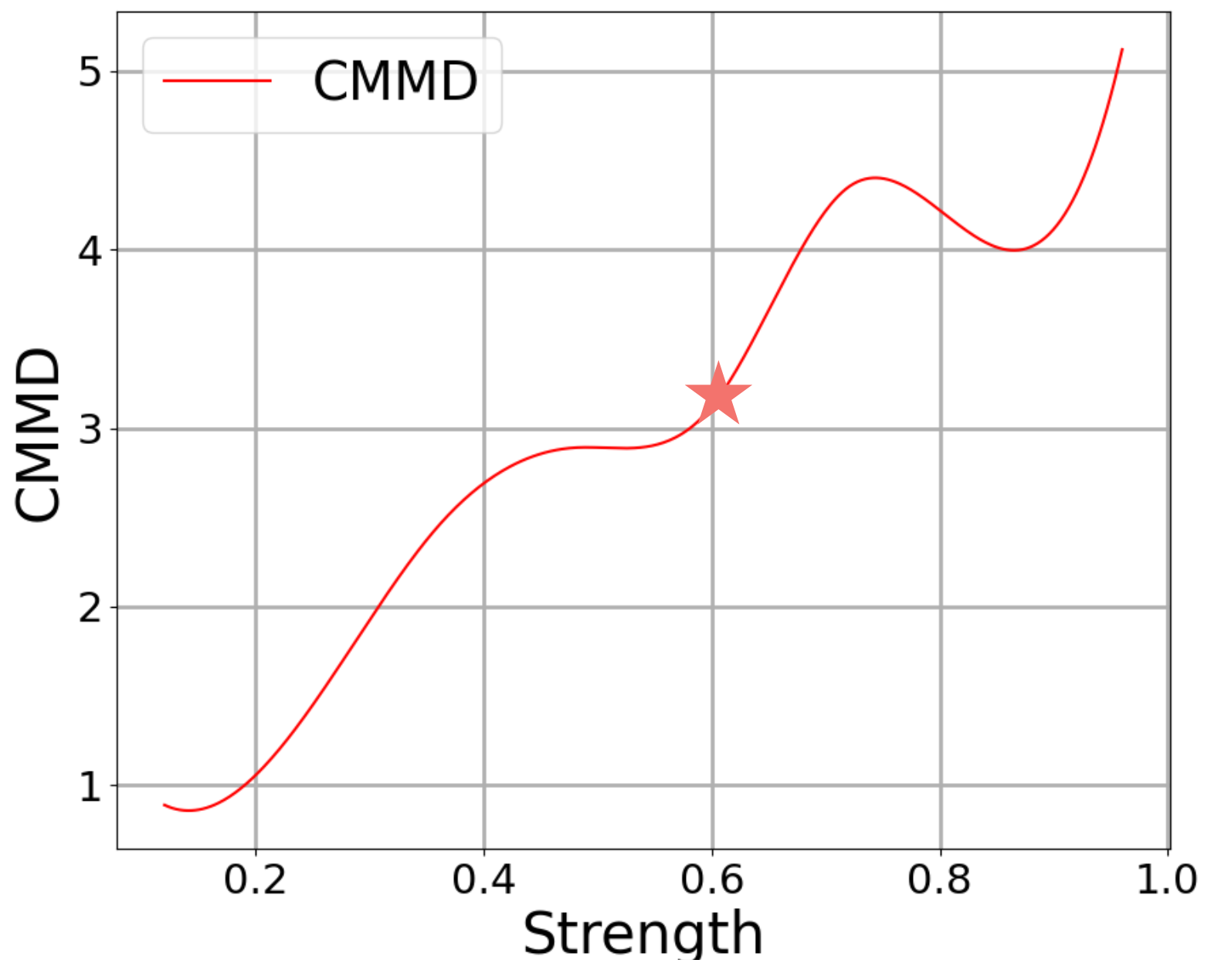} &
       \includegraphics[width=\imwidth]{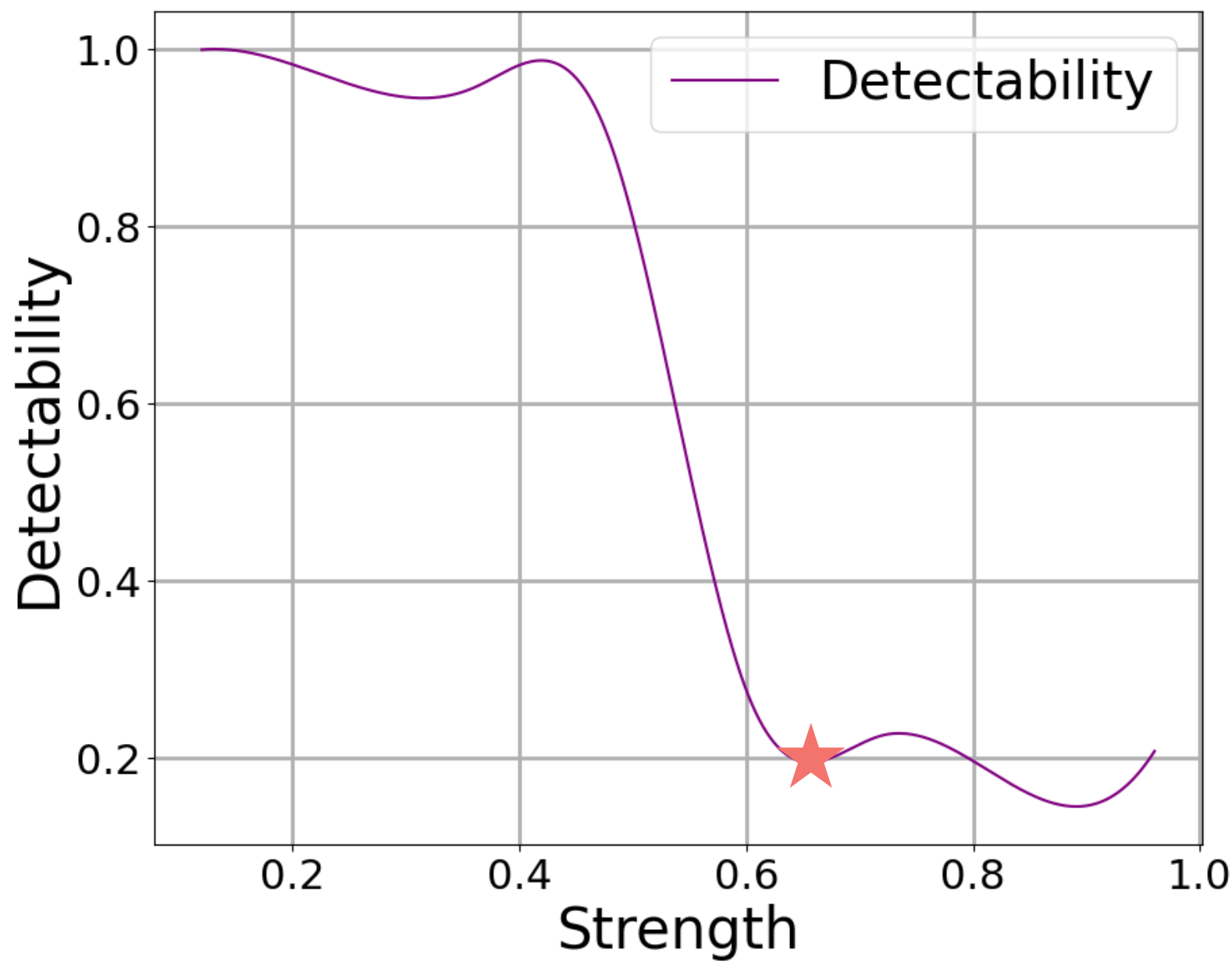} \\
       \\
       \multicolumn{2}{c}{
\begin{tcolorbox}[enhanced,attach boxed title to top left={yshift=-1mm,yshifttext=-1mm,xshift=8pt},left=1pt,right=1pt,top=1pt,bottom=1pt,colback=orange!5!white,colframe=orange!70!black,colbacktitle=orange!70!black,title=Observations,fonttitle=\ttfamily\bfseries\scshape\fontsize{8}{9}\selectfont,boxed title style={size=small,colframe=violet!50!black},width=0.5\textwidth]
\begin{itemize}
    \item[\ding{224}] The least semantic distortion occurs for low strength values ($s<0.4$) and $gs$ in the range of $4-7$. 
    \item[\ding{224}] The detectability decreases with a strength value over $0.4$ and guidance scale value over $12$. 
\end{itemize}
\end{tcolorbox}
    } & 
       \multicolumn{2}{c}{
\begin{tcolorbox}[enhanced,attach boxed title to top left={yshift=-1mm,yshifttext=-1mm,xshift=8pt},left=1pt,right=1pt,top=1pt,bottom=1pt,colback=orange!5!white,colframe=orange!70!black,colbacktitle=orange!70!black,title=Observations,fonttitle=\ttfamily\bfseries\scshape\fontsize{8}{9}\selectfont,boxed title style={size=small,colframe=violet!50!black},width=0.5\textwidth]
\begin{itemize}
    \item[\ding{224}] The least semantic distortion occurs at low strength values ($s = 0.1-0.3$) and $gs$ and guidance scale values around $3-6$. 
    \item[\ding{224}] The detectability decreases with strength values greater than $0.6$ and guidance scale values above $10$. 
\end{itemize}
\end{tcolorbox}
    }
    \\
       \bottomrule
    \end{tabular}
    }
  \vspace{-1mm}
\caption{This figure shows the variation of CMMD \cite{jayasumana2024rethinking} and detectability of visual paraphrases with respect to strength and guidance scale. \includegraphics[width=3mm]{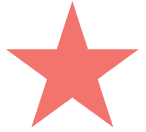} represents the optimal $s$ and $gs$ value for the particular technique. The images were watermarked using Tree Ring Watermarking \cite{wen2023treering} and Stable Signature \cite{fernandez2023stable}.}
\label{fig:strength_gs_plots}
\end{figure*}

\subsubsection{Strength of Paraphrase}

The strength of paraphrasing in visual paraphrasing, ranging from \(0\) to \(1\), determines the extent to which the original image's features are preserved versus the introduction of new variations. Achieving the right balance is crucial to ensure that the paraphrased image remains semantically consistent with the original while varying certain attributes effectively, as outlined in the following points:

\begin{itemize}
    \item A higher strength value allows the model greater creative latitude, enabling it to produce an image that significantly deviates from the original. At a strength value of 1.0, the original image is largely disregarded, resulting in a completely transformed output.
    \item Conversely, a lower strength value maintains closer fidelity to the original image, preserving much of its details.
\end{itemize}

%In image-to-image diffusion, there are two key hyperparameters that control the strength of paraphrase: (i) Strength and (ii) Guidance Scale. These parameters allow for precise adjustments in the degree of transformation applied to the image, ensuring the desired level of alteration while maintaining the integrity of the original content.

%\paragraph{Strength} The strength parameter is one of the most critical factors in visual paraphrasing. It dictates how much the generated image will resemble the initial image. Specifically:

%This parameter is essential when the goal is to subtly alter an image while preserving its core attributes, or conversely, to create a heavily paraphrased version that still aligns with the original context.

\paragraph{Guidance Scale}
The guidance scale parameter determines the extent to which the generated image aligns with the details specified in the text prompt. This parameter plays a crucial role when the paraphrasing process is guided by textual descriptions, as it regulates the balance between strict adherence to the prompt and permitting creative variations, as demonstrated by the following points:

\begin{itemize}
    \item A higher guidance scale value ensures that the generated image closely follows the prompt, resulting in an output that is strongly influenced by the provided text.
    \item A lower guidance scale value allows for greater flexibility, enabling the model to deviate from the prompt and introduce more creative variations in the generated image.
\end{itemize}

\definecolor{Mycolor1}{HTML}{FFCE93}
\definecolor{Mycolor2}{HTML}{CBCEFB}

\begin{table*}[ht]
    \centering
    \renewcommand{\arraystretch}{1.2}
    \setlength{\extrarowheight}{1pt}
    \resizebox{\textwidth}{!}{% 
    
    \begin{tabular}{lccccccccccc}
    \hline
    \multirow{3}{*}{\parbox{2cm}{Watermarking\\ Method}} & \multicolumn{10}{c}{Watermark Detection Rate ($\eta$)} \\
    \cmidrule{2-12}
     & \multirow{2}{*}{Pre-Attack} &  \multicolumn{8}{c}{Post-Attack} \\
    \cmidrule{3-12}
      &  & Brightness & Rotation & JPEG Compression & Gaussian Noise & \multicolumn{5}{c}{\textbf{Visual Paraphrase (Ours)}} \\
      \midrule
        & & & & & & $s=0.2$ & $s=0.4$ & $s=0.6$ & $s=0.8$ & $s=1.0$  \\
    \hline
    DwtDctSVD  & 0.99 & 0.84 & 0.96 & 0.88 & 0.89 & 0.226 & 0.185 & 0.117 & 0.082 & 0.029 \\
    HiDDen  & 1.00 & 0.95 & 0.93 & 0.88 & 0.91 & 0.298 & 0.215 & 0.154 & 0.096 & 0.041  \\
    Stable Signature & 1.00 & 0.931 & 0.98 & 0.85 & 0.90 & 0.319 & 0.225 & 0.176 & 0.107 & 0.059\\
    Tree Ring & \cellcolor[HTML]{CBCEFB}1.00 & \cellcolor[HTML]{CBCEFB}0.98 & \cellcolor[HTML]{CBCEFB}0.92 & \cellcolor[HTML]{CBCEFB}0.97 & \cellcolor[HTML]{CBCEFB}0.98 & \cellcolor[HTML]{CBCEFB}0.473 & \cellcolor[HTML]{CBCEFB}0.394 \(\textcolor{red}{(16\%\Downarrow)}\) & \cellcolor[HTML]{CBCEFB}0.255 \(\textcolor{red}{(35\%\Downarrow)}\) & \cellcolor[HTML]{CBCEFB}0.156 \(\textcolor{red}{(39\%\Downarrow)}\) & \cellcolor[HTML]{CBCEFB}0.097 \(\textcolor{red}{(38\%\Downarrow)}\) \\
    ZoDiac & 1.00 & 0.961 & 0.91 & 0.90 & 0.91 & 0.457 & 0.335 & 0.219 & 0.14 & 0.065  \\
    Gaussian Shading & \cellcolor[HTML]{FFCE93}1.00 & \cellcolor[HTML]{FFCE93}0.99 & \cellcolor[HTML]{FFCE93}0.93 & \cellcolor[HTML]{FFCE93}0.94 & \cellcolor[HTML]{FFCE93}0.93 & \cellcolor[HTML]{FFCE93}0.517 & \cellcolor[HTML]{FFCE93}0.384 \(\textcolor{red}{(26\%\Downarrow)}\) & \cellcolor[HTML]{FFCE93}0.221 \(\textcolor{red}{(42\%\Downarrow)}\) & \cellcolor[HTML]{FFCE93}0.157 \(\textcolor{red}{(28\%\Downarrow)}\) & \cellcolor[HTML]{FFCE93}0.119 \(\textcolor{red}{(24\%\Downarrow)}\)\\
    \hline
    \end{tabular}

    }    
    \caption{Watermark detection rates (\(\eta\)) for various methods on the COCO Dataset \cite{lin2015microsoft} are shown, both pre-attack and post-attack, under common image distortions like brightness adjustment, rotation, JPEG compression, Gaussian noise, and Visual Paraphrase. The Visual Paraphrase attack is tested at five strength levels (\(s=0.2, 0.4, 0.6, 0.8, 1.0\)), with higher strengths causing more significant alterations. As Visual Paraphrase strength increases, detection rates decrease across all methods. However, \colorbox{Mycolor1}{Gaussian Shading} (1\textsuperscript{st}) and \colorbox{Mycolor2}{Tree Ring} (2\textsuperscript{nd}) are the most resilient (relatively) against visual paraphrase attacks. }
    \label{tab:watermark-detection-rate}
    \vspace{-2mm}        
\end{table*}

\section{Performance with De-Watermarking}
After visually paraphrasing a watermarked image, the next crucial step in evaluation involves answering two key questions: (i) To what extent has the visually paraphrased image distorted the original content? Is the distortion too severe to be acceptable, or does it remain within an acceptable range? (ii) How effectively has the paraphrased image removed the watermark from the original image?

\subsubsection{Semantic Distortion}
Semantic distortion refers to the extent to which visual paraphrasing alters the original meaning or content of an image. To quantify this, we employed the continuous Metric Matching Distance (CMMD) score \cite{jayasumana2024rethinking}, which measures the similarity between the original and paraphrased images. Figure \ref{fig:strength_gs_plots} includes a comparison of CMMD scores across various paraphrasing strengths and guidance scale values, illustrating the trade-off between de-watermarking effectiveness and semantic preservation. Our analysis reveals a complex relationship: low-strength paraphrasing typically results in minimal semantic distortion but is less effective at removing watermarks. As paraphrasing strength increases, we observe more successful watermark removal but at the cost of increased semantic distortion. The optimal balance point varies depending on the specific image content and watermarking technique employed. An extended version of Figure \ref{fig:strength_gs_plots}, which includes all discussed watermarking techniques, is provided in the appendix as Figure 8.

\subsubsection{Detectability Rate:} 
The detectability rate is a crucial metric in assessing the effectiveness of watermark detection methods after visual paraphrasing. Our experiments reveal a clear inverse relationship between the strength of visual paraphrasing and the detectability of watermarks. As the intensity of paraphrasing increases, we observe a significant decline in the ability to detect and extract the original watermarks. This trend is consistent across various watermarking techniques, though some algorithms demonstrate more resilience than others. Detail results are presented in Table \ref{tab:watermark-detection-rate}.

\begin{comment}
\subsection{Experiment Setup}
We selected state-of-the-art watermarking methods that have demonstrated strong performance in recent literature. To evaluate the robustness of these methods, we conducted experiments on a subset of images from the COCO dataset, subjecting them to both the visual paraphrase attack and a series of traditional adversarial attacks. The attacks applied, along with their parameters, are as follows:
\begin{itemize}
    \item \textbf{Brightness Enhancement:} Brightness increased by a factor of 2.
    \item \textbf{JPEG Compression:} Quality setting reduced to 50.
    \item \textbf{Gaussian Noise:} Noise with a standard deviation of 0.05 added to the images.
    \item \textbf{Rotation:} Images rotated by $\pm 45$ degrees.
\end{itemize}
\end{comment}

\textbf{Experiment Setup}: For each attack, we report the watermark probability post-attack. Additionally, we determine the success of watermark detection by applying a threshold on the obtained probability. These threshold values were derived from the original publications of each watermarking method. The results are reported in Table \ref{tab:watermark-detection-rate}. For methods that embed the watermark in the image generation process, such as Tree-Ring and Gaussian Shading, the given captions in the subset were used to generate new watermarked images using Stable Diffusion XL \cite{podell2023sdxlimprovinglatentdiffusion}.
%Furthermore, we extend our evaluation to subsets of the DiffusionDB and WikiArt datasets \cite{saleh2015wikiart}, with results provided in the Appendix. For the visual paraphrase attack, we used the captions provided in the datasets for the DiffusionDB and COCO subsets. In case of WikiArt, the caption for images were generated using Kosmos-2. For the WikiArt subset, the visual paraphrase was generated using an empty prompt. 
All watermarking methods were tested at their default settings as specified in the original publications.

\textbf{Datasets}: For our experiments, we utilize three distinct datasets: MS COCO \cite{lin2015microsoft}, DiffusionDB \cite{wang2023diffusiondb}, and WikiArt \cite{saleh2015wikiart}. By utilizing these three distinct datasets, we aim to ensure that our results are generalized and not biased towards any particular image type or source.

\subsection{Visual Paraphrasing vs. Information Loss}
% \vspace{-5mm}
While we have already discussed measuring semantic distortion using the CMMD score, we critically contend that CMMD may have limitations in capturing significant information loss. With this consideration in mind, we designed a human annotation task. The objective of this task is to obtain annotations from human users regarding the acceptability of these automatically paraphrased images. Furthermore, as previously discussed, there are two controlling factors in visual paraphrasing, namely, the strength of the paraphrase and the guiding scale. 
A pertinent question arises: Are there upper limits on these two parameters that should not be exceeded, beyond which the generated paraphrases start to exhibit excessive distortion?

\begin{figure}[H]
    \centering
    \begin{subfigure}[t]{0.46\linewidth} 
        \centering
        \includegraphics[width=\linewidth]{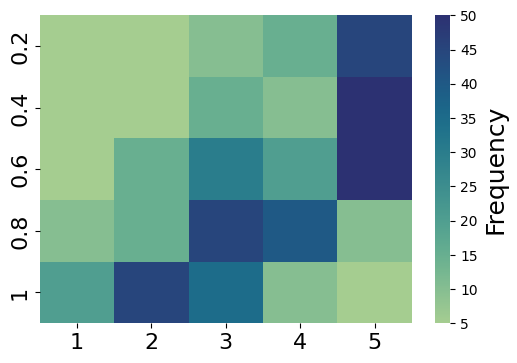}
        \caption{Paraphrase Strength ($s$) vs. Semantic Distortion}
        \label{fig:mos_strength}
    \end{subfigure}
    \hfill
    \begin{subfigure}[t]{0.46\linewidth} 
        \centering
        \includegraphics[width=\linewidth]{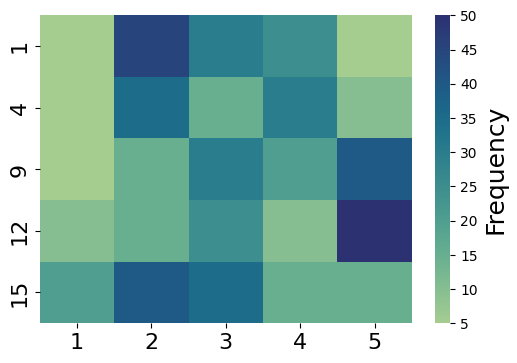}
        \caption{Guidance Scale ($gs$) vs. Semantic Distortion}
        \label{fig:mos_guidance}
    \end{subfigure}
\vspace{-2mm}        
    \caption{These heatmaps illustrate the MOS Scores from human annotations, showing the impact of varying Strength and Guidance Scale on content distortion caused by visual paraphrasing.}
    \label{fig:mos_combined}
\end{figure}
 To investigate this, we generated 1,000 paraphrased images, equally distributed based on paraphrase strength and guiding scale. The Mean Opinion Scores (MOS) of five annotators are reported in Figures \ref{fig:mos_strength} and \ref{fig:mos_guidance}, corresponding to paraphrase strength and guiding scale, respectively. 

Our research indicates that for strength, the modal MOS is observed at a value of 0.4, signifying optimal acceptability. Conversely, the lowest acceptability is recorded at a MOS of 0.8, where the paraphrases are deemed least acceptable. Regarding the guidance scale, the highest MOS frequencies are noted at values of 1 and 3, which suggests that these settings yield the most acceptable results. In contrast, a guidance scale of 13 results in the least acceptable paraphrases. These observations highlight the essential role of adjusting both strength and guidance scale to enhance the acceptability of paraphrases. Detailed visual examples and further analysis can be found in the appendix, as illustrated in Figure 9.

\section{Conclusion}
\vspace{-1mm}    
In this study, we empirically demonstrate that existing image watermarking techniques are fragile and susceptible to circumvention via visual paraphrase attacks. To facilitate further research, we are releasing the first-of-its-kind visual paraphrase dataset, along with the accompanying code for all state-of-the-art watermarking methods. This work underscores the urgent need for the scientific community to prioritize the development of more robust watermarking strategies. We anticipate that this research will serve as a benchmark for future efforts to create watermarking methods resilient to visual paraphrase attacks.

%In conclusion, the imperative for robust AI-generated image detection mechanisms has never been more pressing. The proliferation of highly realistic generative models poses significant challenges to content ownership and the dissemination of misinformation. By exploring watermarking methods, visual paraphrasing scenarios, and the control of information loss, this paper has underscored the critical role of technological innovation in safeguarding the integrity of digital imagery. Moving forward, continued research and development in AI-generated image detection promise to fortify the foundations of a trustworthy digital ecosystem, empowering individuals and organizations to navigate the complexities of the information age with confidence and discernment. 

\section{Ethical Considerations}

The development of visual paraphrasing methods that can bypass state-of-the-art watermarking techniques raises important ethical considerations. While our research aims to advance image processing and improve watermarking resilience, we acknowledge the potential for misuse, such as unauthorized removal of watermarks from copyrighted images. To mitigate these risks, we will responsibly disclose our findings to stakeholders, restrict access to our methodologies and tools to legitimate entities, and advocate for the establishment of ethical guidelines for the use of visual paraphrasing tools. Our goal is to conduct research that aligns with the highest ethical standards, promotes collaborative improvements in watermarking technologies, and respects intellectual property rights and broader societal values.
\bibliography{aaai25}
\clearpage
\newpage
\onecolumn
\section*{Frequently Asked Questions (FAQs)}\label{sec:FAQs}

\begin{itemize}
[leftmargin=4mm]
\setlength\itemsep{0em}
    \item[\ding{93}] {\fontfamily{lmss} \selectfont \textbf{How did you determine the optimal combination of paraphrase strength ($s$) and guiding scale ($gs$), given the multiple possibilities, such as higher $s$ with higher $gs$, or other variations?}}
    \begin{description}
    \item[\ding{224}] 
     We conducted a series of rigorous experiments to explore various combinations of paraphrase strength (s) and guiding scale (gs), shown in Figure \ref{fig:strength_gs_plots_all}. By systematically varying these parameters, we were able to identify the configurations that produced the highest Mean Opinion Scores (MOS) for paraphrase acceptability. 
    \end{description}

    %\item[\ding{93}] {\fontfamily{lmss} \selectfont \textbf{\cite{chakraborty2023possibilities} presents an opposing view on the subject matter, how do we compare our study with theirs?}} 
    %\vspace{-1mm}
    %\begin{description}
    %\item[\ding{224}]  It is important to note that a recent study \cite{chakraborty2023possibilities} contradicts our findings and claims otherwise.
    %The study postulates that given enough sample points, whether the output was derived from a human vs an LLM is detectable, irrespective of the LLM used for AI-generated text. The sample size of this dataset is a function of the difference in the distribution of human text vs AI-text, with a smaller sample size enabling detection if the distributions show significant differences. Furthermore, the authors propose that employing techniques such as watermarking can change the distributions of AI text, making it more separable from human-text distribution and thus detectable. However, the main drawback of this argument is that given a single text snippet (say, an online article or a written essay), detecting whether it is AI-generated is not possible. Also, the proposed technique may not be cost-efficient compute-wise, especially as new LLMs emerge. 
    %\end{description}    

    \item[\ding{93}] {\fontfamily{lmss} \selectfont \textbf{Is the optimal combination of paraphrase strength ($s$) and guiding scale ($gs$) dependent on the model?}}
    \vspace{-1mm}
    \begin{description}
    \item[\ding{224}]  Yes, the optimal combination of paraphrase strength (s) and guiding scale (gs) can vary depending on the model. Different models have unique architectures and training data, which influence how they respond to variations in these parameters. Therefore, fine-tuning these settings for each specific model is crucial to achieving the best balance between maintaining image semantics and ensuring high visual quality.
    \end{description}

    \item[\ding{93}] {\fontfamily{lmss} \selectfont \textbf{Why Gaussian Shading is the most resilient towards Visual Paraphrase attack? What we learn from it?}}
    \vspace{-1mm}
    \begin{description}
    \item[\ding{224}] Gaussian Shading is the most resilient to visual paraphrase attacks because it smooths out high-frequency details and textures in an image, which are typically exploited in such attacks to alter the visual appearance while preserving recognizability. By applying Gaussian shading, the image becomes less susceptible to small perturbations and subtle modifications, which are commonly used in visual paraphrase attacks to create misleading variations. This resilience teaches us that the robustness of image processing techniques can be significantly enhanced by focusing on reducing the sensitivity to fine details and focusing on the broader, less granular features of the image, thus improving the security and reliability of image recognition systems.
    \end{description}

    \item[\ding{93}] {\fontfamily{lmss} \selectfont \textbf{ Why did you compare only six methods? }} 
    \vspace{-1mm}
    \begin{description}
    \item[\ding{224}] 
    We have focused on methods that have demonstrated strong performance in recent literature, particularly emphasizing dynamic approaches. While we have mostly omitted older static watermarking methods, we have included results on the DwtdctSvd technique due to its popularity and relevance compared to others within the same category.
    \end{description}

    \item[\ding{93}] {\fontfamily{lmss} \selectfont \textbf{ On average, at what values of strength and guidance scale does the generated image deviate significantly from the original image? }} 
    \vspace{-1mm}
    \begin{description}
    \item[\ding{224}] 
    The generated image begins to deviate significantly from the original when the strength value exceeds 0.8. Similarly, notable deviations occur when the guidance scale is set to values below 4 or above 13. These settings allow the model more flexibility, leading to greater alterations in the image's appearance while potentially straying from the original content and context.
    
    \end{description}

    \item[\ding{93}] {\fontfamily{lmss} \selectfont \textbf{ You use KOSMOS-2 for caption generation. How would the performance of the visual paraphrasing attack be affected if a different captioning model was used, especially one with varying levels of detail and accuracy?}} 
    \vspace{-1mm}
    \begin{description}
    \item[\ding{224}] 
     While KOSMOS-2 is a strong performer, different captioning models could indeed influence the attack's effectiveness. A less detailed caption might lead to more significant semantic distortion during the visual paraphrasing process, potentially hindering the removal of the watermark. Conversely, a highly accurate and detailed caption could improve the attack by providing more precise guidance to the image-to-image diffusion model, leading to better preservation of semantic content while still removing the watermark. Further research could explore the impact of various captioning models with varying levels of accuracy and detail on the success of visual paraphrasing attacks.
    
    \end{description}

    \item[\ding{93}] {\fontfamily{lmss} \selectfont \textbf{ The paper focuses on diffusion-based watermarking techniques. How do you think your visual paraphrasing attack would perform against GAN-based watermarking methods, or those that employ steganographic techniques in the spatial domain?}} 
    \vspace{-1mm}
    \begin{description}
    \item[\ding{224}] 
     In this paper our focus was on diffusion models due to their prominence in current watermarking research. GAN-based or spatial domain watermarking techniques might exhibit different vulnerabilities. GAN-based methods could be more robust due to their adversarial training nature, potentially making it harder to generate paraphrases that both remove the watermark and maintain image fidelity. Spatial domain techniques might be vulnerable to subtle pixel manipulations introduced during visual paraphrasing. Further investigation is needed to assess the effectiveness of our attack against these alternative watermarking approaches.
    
    \end{description}

    \item[\ding{93}] {\fontfamily{lmss} \selectfont \textbf{ You primarily evaluate the attack based on CMMD and detectability. Are there other metrics, especially those focused on perceptual similarity or specific watermarking features, that could provide a more comprehensive evaluation?}} 
    \vspace{-1mm}
    \begin{description}
    \item[\ding{224}] 
     CMMD and detectability provide a good starting point, but other metrics could enhance the evaluation. Perceptual similarity metrics like LPIPS (Learned Perceptual Image Patch Similarity) could capture subtle differences in visual appearance missed by CMMD. Analyzing specific watermarking features, like frequency distribution changes or alterations in specific latent space dimensions, could offer more granular insights into the attack's impact. Incorporating these additional metrics would provide a richer understanding of the attack's efficacy.
    
    \end{description}

    \item[\ding{93}] {\fontfamily{lmss} \selectfont \textbf{ You mention the potential for adversarial training to improve watermarking robustness. Can you elaborate on how adversarial training could be specifically tailored to defend against visual paraphrasing attacks?}} 
    \vspace{-1mm}
    \begin{description}
    \item[\ding{224}] 
     Adversarial training could be a powerful defense mechanism. We envision training the watermarking encoder and decoder against a dataset of visually paraphrased images. This would expose the model to the types of perturbations introduced by our attack, forcing it to learn more robust embedding strategies. The training process could involve generating paraphrases using different strengths and guidance scales to ensure generalization across a variety of attack parameters.
    
    \end{description}

    \item[\ding{93}] {\fontfamily{lmss} \selectfont \textbf{ The paper acknowledges the ethical implications of visual paraphrasing. What specific measures, beyond responsible disclosure, can be taken to prevent the misuse of this technique for malicious purposes like copyright infringement?}} 
    \vspace{-1mm}
    \begin{description}
    \item[\ding{224}] 
     Beyond responsible disclosure, we could explore incorporating "detection mechanisms" within the visual paraphrasing tool itself. This could involve training a classifier to identify watermarked images and either prevent their paraphrasing or add a persistent notification indicating potential copyright protection. Another avenue could be developing a collaborative platform where researchers can share newly developed watermarking techniques and test their resilience against visual paraphrasing, fostering a continuous improvement cycle in watermarking robustness.
    
    \end{description}

    \item[\ding{93}] {\fontfamily{lmss} \selectfont \textbf{ The paper claims that visual paraphrasing is a novel approach to remove watermarks. However, image editing techniques like inpainting and masking have been around for a while. How does visual paraphrasing differ from these existing techniques?}} 
    \vspace{-2mm}
    \begin{description}
    \item[\ding{224}] 
     While inpainting and masking can be used to remove visible watermarks, they often leave noticeable artifacts or require precise manual intervention. Visual paraphrasing, on the other hand, leverages the capabilities of image-to-image diffusion models to generate visually similar images guided by a text caption. This process aims to preserve the semantic content while subtly altering the image, making it more challenging to detect and remove the watermark. It can achieve a higher level of realism and detail compared to inpainting and masking while being less susceptible to detection.
    
    \end{description}

    \item[\ding{93}] {\fontfamily{lmss} \selectfont \textbf{ The paper mainly focuses on visual paraphrasing with Stable Diffusion. Have the authors explored the efficacy of other image-to-image diffusion models or other generative AI models for this task?}} 
    \vspace{-1mm}
    \begin{description}
    \item[\ding{224}] 
     Our paper primarily uses Stable Diffusion for its established capabilities and accessibility. While we acknowledge the potential of other image-to-image diffusion models and generative AI systems for visual paraphrasing, we haven't yet extensively tested them. This is a potential area for future research, examining the effectiveness of different models for watermark removal and the potential impact of different model architectures on the results.
    
    \end{description}

\end{itemize}
\newpage

\clearpage
% \section{Experiments}
\section{Appendix}
This section provides supplementary material in the form of additional examples, implementation details, etc. to bolster the reader’s understanding of the concepts presented in this work.
\subsection{Stripping Metadata}
\label{sec:metadata-removal}
While attaching metadata to images is one proposed method for identifying AI-generated content, this approach is vulnerable to simple removal techniques. Here we demonstrate how easily metadata can be stripped from image files, rendering this method ineffective for long-term content attribution.
\subsubsection{Removing Metadata Using ExifTool}
To illustrate the simplicity of metadata removal, we'll use ExifTool \cite{ExifTool}, a popular and freely available command-line application for reading, writing, and editing metadata in various file types, including images.

1. Consider an AI-generated image with embedded metadata identifying its source.

\begin{verbatim}
$ exiftool sample_image.jpg
File Name                  : sample_image.jpg
File Size                  : 2.5 MB
File Type                  : JPEG
AI Generator               : Midjourney v5
Creation Time              : 2023:08:15 14:30:22
Image Width                : 1024
Image Height               : 1024
\end{verbatim}

2. Using the \texttt{-all=} ExifTool command, we can strip all metadata from the image:

\begin{verbatim}
$ exiftool -all= sample_image.jpg
1 image files updated
\end{verbatim}

3. Checking the image again, we see that all metadata has been removed:

\begin{verbatim}
$ exiftool sample_image.jpg
File Name                     : sample_image.jpg
File Size                     : 2.5 MB
File Type                     : JPEG
Image Width                   : 1024
Image Height                  : 1024
\end{verbatim}

This demonstration shows that with a single command, all identifying metadata can be eliminated from an image file. The process is quick, requires no specialized knowledge, and can be easily automated for batch processing.

\clearpage

\subsection{Examples on Strength Variation}

\begin{figure*}[h!]
    \centering
    \resizebox{\linewidth}{!}{
    \begin{tabular}{l}
        \begin{minipage}{\linewidth}\centering \includegraphics[width=\textwidth,height=\textheight,keepaspectratio]{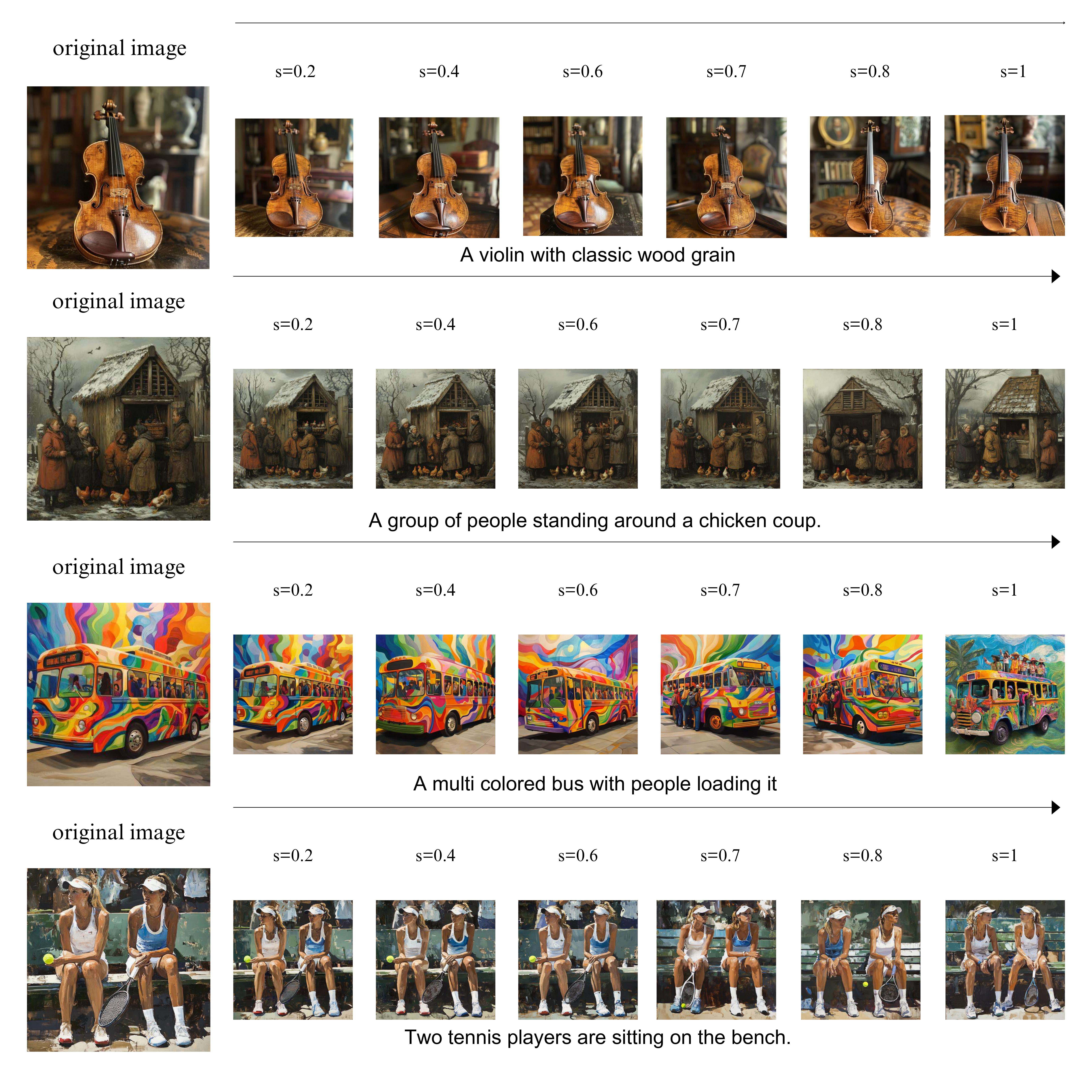}
        \end{minipage}
        \label{fig:strength_variation}
    \end{tabular}

} 

\caption{ 
Examples of Visual Paraphrasing with varying levels of strengths.
}

\label{fig:strength-variation-examples}
\end{figure*}

The figure \ref{fig:strength-variation-examples} showcases examples of visual paraphrasing at different strength levels. The images illustrate how varying the strength parameter impacts the degree of transformation applied to the original image. Lower strength values result in paraphrases that closely resemble the original, while higher strength values introduce more significant alterations. Figure \ref{fig:strength-variation-examples}.

\vspace{-4mm}
\clearpage

% \begin{figure*}[h]
%     \centering
%     \resizebox{\linewidth}{!}{
%     \begin{tabular}{l}
%         \begin{minipage}{\linewidth}\centering \includegraphics[width=\linewidth]{img/img2img_violin.pdf}\end{minipage}
%     \end{tabular}
%     } 

% \caption{This figure illustrates the image-to-image diffusion process \cite{1021076}. The top row demonstrates the forward diffusion process, where the original image progressively becomes more noisy. The bottom row shows the denoising process, where noise is incrementally removed from the noisy image, guided by text conditioning to generate the final, Visual Paraphrased image.
% }

% \label{fig:img2img}
% \end{figure*}

\subsection{Dewatermarking Across Three Datasets}

The table \ref{tab:watermark-detection} presents the detection rates, denoted as $\eta$, for various watermarking techniques when subjected to different types of attacks. This data highlights the effectiveness of each watermarking method in maintaining its integrity and being detected under varying conditions, offering insights into the robustness of these techniques against adversarial manipulations. The comparison across different methods and attack scenarios provides a comprehensive overview of each technique's resilience.

\begin{table*}[h]
    \centering
    \renewcommand{\arraystretch}{1.2}
    \setlength{\extrarowheight}{1pt}
\resizebox{\textwidth}{!}{%
    \begin{tabular}{lccccccccccc}
    \hline
    \multirow{3}{*}{\parbox{2cm}{Watermarking\\ Method}} & \multicolumn{10}{c}{Watermark Detection Rate ($\eta$)} \\
    \cmidrule{2-12}
     & \multirow{2}{*}{Pre-Attack} &  \multicolumn{8}{c}{Post-Attack} \\
    \cmidrule{3-12}
      &  & Brightness & Rotation & JPEG Compression & Gaussian Noise & \multicolumn{5}{c}{\textbf{Visual Paraphrase (Ours)}} \\
      \midrule
        & & & & & & $s=0.2$ & $s=0.4$ & $s=0.6$ & $s=0.8$ & $s=1.0$  \\
    \hline
    DwtDctSVD  & 0.99 & 0.84 & 0.96 & 0.88 & 0.89 & 0.226 & 0.185 & 0.117 & 0.082 & 0.029 \\
    HiDDen  & 1.00 & 0.95 & 0.93 & 0.88 & 0.91 & 0.298 & 0.215 & 0.154 & 0.096 & 0.041  \\
    Stable Signature & 1.00 & 0.931 & 0.98 & 0.85 & 0.90 & 0.319 & 0.225 & 0.176 & 0.107 & 0.059\\
    Tree Ring & \cellcolor[HTML]{CBCEFB}1.00 & \cellcolor[HTML]{CBCEFB}0.98 & \cellcolor[HTML]{CBCEFB}0.92 & \cellcolor[HTML]{CBCEFB}0.97 & \cellcolor[HTML]{CBCEFB}0.98 & \cellcolor[HTML]{CBCEFB}0.473 & \cellcolor[HTML]{CBCEFB}0.394 \(\textcolor{red}{(16\%\Downarrow)}\) & \cellcolor[HTML]{CBCEFB}0.255 \(\textcolor{red}{(35\%\Downarrow)}\) & \cellcolor[HTML]{CBCEFB}0.156 \(\textcolor{red}{(39\%\Downarrow)}\) & \cellcolor[HTML]{CBCEFB}0.097 \(\textcolor{red}{(38\%\Downarrow)}\) \\
    ZoDiac & 1.00 & 0.961 & 0.91 & 0.90 & 0.91 & 0.457 & 0.335 & 0.219 & 0.14 & 0.065  \\
    Gaussian Shading & \cellcolor[HTML]{FFCE93}1.00 & \cellcolor[HTML]{FFCE93}0.99 & \cellcolor[HTML]{FFCE93}0.93 & \cellcolor[HTML]{FFCE93}0.94 & \cellcolor[HTML]{FFCE93}0.93 & \cellcolor[HTML]{FFCE93}0.517 & \cellcolor[HTML]{FFCE93}0.384 \(\textcolor{red}{(26\%\Downarrow)}\) & \cellcolor[HTML]{FFCE93}0.221 \(\textcolor{red}{(42\%\Downarrow)}\) & \cellcolor[HTML]{FFCE93}0.157 \(\textcolor{red}{(28\%\Downarrow)}\) & \cellcolor[HTML]{FFCE93}0.119 \(\textcolor{red}{(24\%\Downarrow)}\)\\
    \midrule
        \multicolumn{12}{c}{\textit{DiffusionDB \cite{wang2023diffusiondb}}} \\
        \midrule
    
    DwtDctSVD  & 0.99 & 0.93 & 0.91 & 0.87 & 0.81 & 0.215 & 0.176 & 0.145 & 0.062 & 0.037 \\
    HiDDen  & 0.99 & 0.94 & 0.95 & 0.85 & 0.82 & 0.314 & 0.267 & 0.153 & 0.103 & 0.052  \\
    Stable Signature & 1.00 & 0.98 & 0.97 & 0.91 & 0.86 & 0.325 & 0.245 & 0.164 & 0.116 & 0.057\\
    Tree Ring & \cellcolor[HTML]{CBCEFB}1.00 & \cellcolor[HTML]{CBCEFB}0.99 & \cellcolor[HTML]{CBCEFB}0.99 & \cellcolor[HTML]{CBCEFB}0.94 & \cellcolor[HTML]{CBCEFB}0.92 & \cellcolor[HTML]{CBCEFB}0.452 & \cellcolor[HTML]{CBCEFB}0.351 \(\textcolor{red}{(22\%\Downarrow)}\) & \cellcolor[HTML]{CBCEFB}0.227 \(\textcolor{red}{(35\%\Downarrow)}\) & \cellcolor[HTML]{CBCEFB}0.171 \(\textcolor{red}{(24\%\Downarrow)}\) & \cellcolor[HTML]{CBCEFB}0.108 \(\textcolor{red}{(37\%\Downarrow)}\) \\
    ZoDiac & 1.00 & 0.98 & 0.97 & 0.93 & 0.94 & 0.412 & 0.324 & 0.264 & 0.162 & 0.084  \\
    Gaussian Shading & \cellcolor[HTML]{FFCE93}1.00 & \cellcolor[HTML]{FFCE93}0.99 & \cellcolor[HTML]{FFCE93}0.98 & \cellcolor[HTML]{FFCE93}0.93 & \cellcolor[HTML]{FFCE93}0.91 & \cellcolor[HTML]{FFCE93}0.493 & \cellcolor[HTML]{FFCE93}0.357 \(\textcolor{red}{(28\%\Downarrow)}\) & \cellcolor[HTML]{FFCE93}0.285 \(\textcolor{red}{(20\%\Downarrow)}\) & \cellcolor[HTML]{FFCE93}0.193 \(\textcolor{red}{(32\%\Downarrow)}\) & \cellcolor[HTML]{FFCE93}0.124 \(\textcolor{red}{(36\%\Downarrow)}\)\\

     \midrule
        \multicolumn{12}{c}{\textit{WikiArt \cite{saleh2015wikiart}}} \\
        \midrule
    
    DwtDctSVD  & 1.00 & 0.95 & 0.93 & 0.86 & 0.81 & 0.194 & 0.152 & 0.105 & 0.073 & 0.035 \\
    HiDDen  & 1.00 & 0.97 & 0.94 & 0.87 & 0.85 & 0.278 & 0.235 & 0.193 & 0.103 & 0.039  \\
    Stable Signature & 1.00 & 0.98 & 0.98 & 0.89 & 0.87 & 0.342 & 0.251 & 0.146 & 0.091 & 0.061\\
    Tree Ring & \cellcolor[HTML]{CBCEFB}1.00 & \cellcolor[HTML]{CBCEFB}0.98 & \cellcolor[HTML]{CBCEFB}0.99 & \cellcolor[HTML]{CBCEFB}0.94 & \cellcolor[HTML]{CBCEFB}0.93 & \cellcolor[HTML]{CBCEFB}0.413 & \cellcolor[HTML]{CBCEFB}0.296 \(\textcolor{red}{(28\%\Downarrow)}\) & \cellcolor[HTML]{CBCEFB}0.197 \(\textcolor{red}{(33\%\Downarrow)}\) & \cellcolor[HTML]{CBCEFB}0.116 \(\textcolor{red}{(41\%\Downarrow)}\) & \cellcolor[HTML]{CBCEFB}0.082 \(\textcolor{red}{(29\%\Downarrow)}\) \\
    ZoDiac & 1.00 & 0.98 & 0.97 & 0.93 & 0.94 & 0.382 & 0.285 & 0.214 & 0.137 & 0.074  \\
    Gaussian Shading & \cellcolor[HTML]{FFCE93}1.00 & \cellcolor[HTML]{FFCE93}0.99 & \cellcolor[HTML]{FFCE93}0.98 & \cellcolor[HTML]{FFCE93}0.92 & \cellcolor[HTML]{FFCE93}0.91 & \cellcolor[HTML]{FFCE93}0.466 & \cellcolor[HTML]{FFCE93}0.341 \(\textcolor{red}{(26\%\Downarrow)}\) & \cellcolor[HTML]{FFCE93}0.253 \(\textcolor{red}{(26\%\Downarrow)}\) & \cellcolor[HTML]{FFCE93}0.178 \(\textcolor{red}{(30\%\Downarrow)}\) & \cellcolor[HTML]{FFCE93}0.101 \(\textcolor{red}{(43\%\Downarrow)}\)\\
        \bottomrule
    
    \end{tabular}
}    
    \caption{Watermark detection rates ($\eta$) for various methods on the COCO \cite{lin2015microsoft}, DiffusionDB \cite{wang2023diffusiondb} and WikiArt \cite{wang2023diffusiondb} datasets are shown, both pre-attack and post-attack, under common image distortions like brightness adjustment, rotation, JPEG compression, Gaussian noise, and Visual Paraphrase. The Visual Paraphrase attack is tested at five strength levels ($s=0.2, 0.4, 0.6, 0.8, 1.0$), with higher strengths causing more significant alterations. As Visual Paraphrase strength increases, detection rates decrease across all methods. However, \colorbox{Mycolor1}{Gaussian Shading} (1\textsuperscript{st}) and \colorbox{Mycolor2}{Tree Ring}  (2\textsuperscript{nd}) are the most resilient (relatively) against visual paraphrase attacks.}
    \label{tab:watermark-detection}
\end{table*}
\clearpage

\subsection{Impact of Strength and Guidance Scale on Watermark Detectability and Quality}
\begin{figure*}[h!]
\centering
    \scriptsize
    \newcommand{\imwidth}{0.18\textwidth}
    \resizebox{!}{10cm}{
    \begin{tabular}{cccc}
    \toprule
    \multicolumn{2}{c}{\textbf{Tree Ring}} & \multicolumn{2}{c}{\textbf{Stable Signature}} \\
    \midrule
       \includegraphics[width=\imwidth]{img/variation_plots/tr_guidance_scale_vs_cmmd_gs.png} &
       \includegraphics[width=\imwidth]{img/variation_plots/tr_guidance_scale_vs_det_gs.png} &
       \includegraphics[width=\imwidth]{img/variation_plots/stable_guidance_scale_vs_cmmd_gs.png} &
       \includegraphics[width=\imwidth]{img/variation_plots/stable_guidance_scale_vs_det_gs.png}  \\

       \includegraphics[width=\imwidth]{img/variation_plots/tr_strength_vs_cmmd_s.png} &
       \includegraphics[width=\imwidth]{img/variation_plots/tr_strength_vs_det_s.png} &
       \includegraphics[width=\imwidth]{img/variation_plots/stable_strength_vs_cmmd_s.png} &
       \includegraphics[width=\imwidth]{img/variation_plots/stable_strength_vs_det_s.png} \\
\multicolumn{2}{c}{
\begin{tcolorbox}[enhanced,attach boxed title to top left={yshift=-1mm,yshifttext=-1mm,xshift=8pt},left=1pt,right=1pt,top=1pt,bottom=1pt,colback=orange!5!white,colframe=orange!70!black,colbacktitle=orange!70!black,title=Observations,fonttitle=\ttfamily\bfseries\scshape\fontsize{8}{9}\selectfont,boxed title style={size=small,colframe=violet!50!black},width=0.5\textwidth]
\begin{itemize}
    \item For Tree Ring, the least semantic distortion occurs for low strength values ($<0.4$) and guidance scale values in the range of 4-7.
    \item The detectability decreases with a strength value over 0.4 and guidance scale value over 12.
\end{itemize}
\end{tcolorbox}
    } & 
\multicolumn{2}{c}{
\begin{tcolorbox}[enhanced,attach boxed title to top left={yshift=-1mm,yshifttext=-1mm,xshift=8pt},left=1pt,right=1pt,top=1pt,bottom=1pt,colback=orange!5!white,colframe=orange!70!black,colbacktitle=orange!70!black,title=Observations,fonttitle=\ttfamily\bfseries\scshape\fontsize{8}{9}\selectfont,boxed title style={size=small,colframe=violet!50!black},width=0.5\textwidth]
\begin{itemize}
    \item The least semantic distortion for Stable Signature occurs at low strength values (0.1-0.3) and guidance scale values around 3-6.
    \item The detectability decreases with strength values greater than 0.6 and guidance scale values above 10.
\end{itemize}
\end{tcolorbox}
    }
    \\
    \midrule
    \multicolumn{2}{c}{\textbf{Zodiac}} & \multicolumn{2}{c}{\textbf{Gaussian Shading}} \\
    \midrule
        \includegraphics[width=\imwidth]{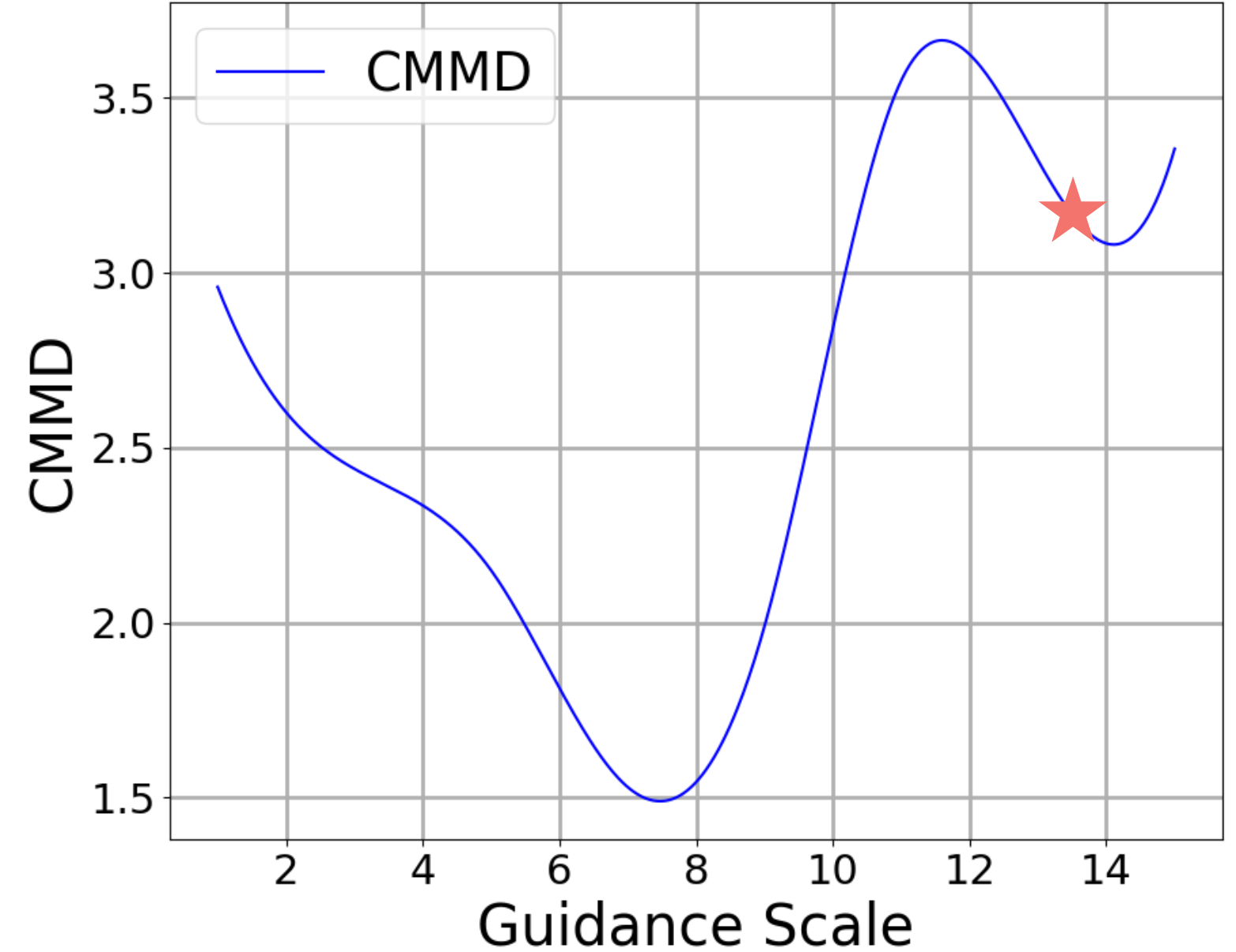} &
       \includegraphics[width=\imwidth]{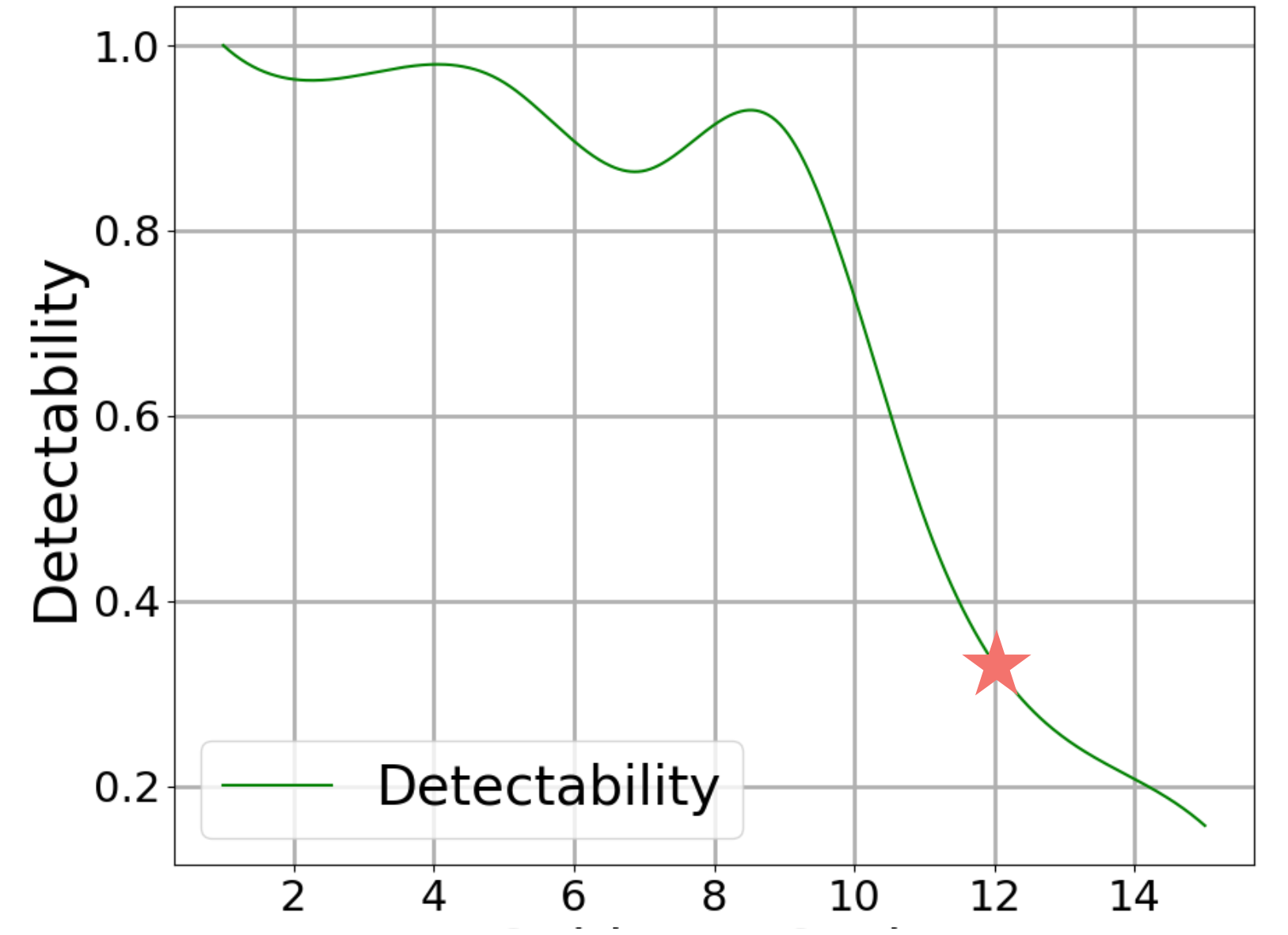} &
       \includegraphics[width=\imwidth]{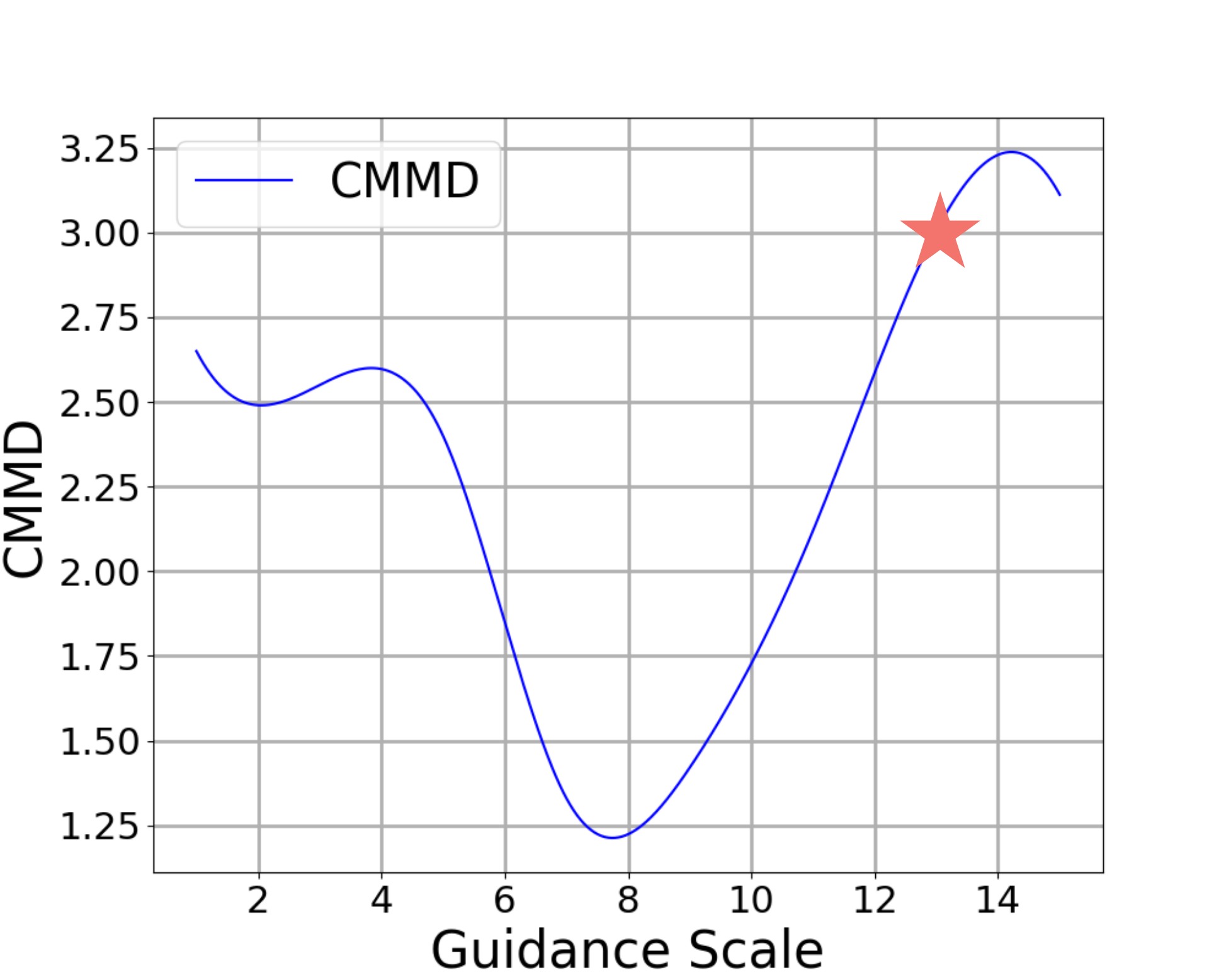} &
       \includegraphics[width=\imwidth]{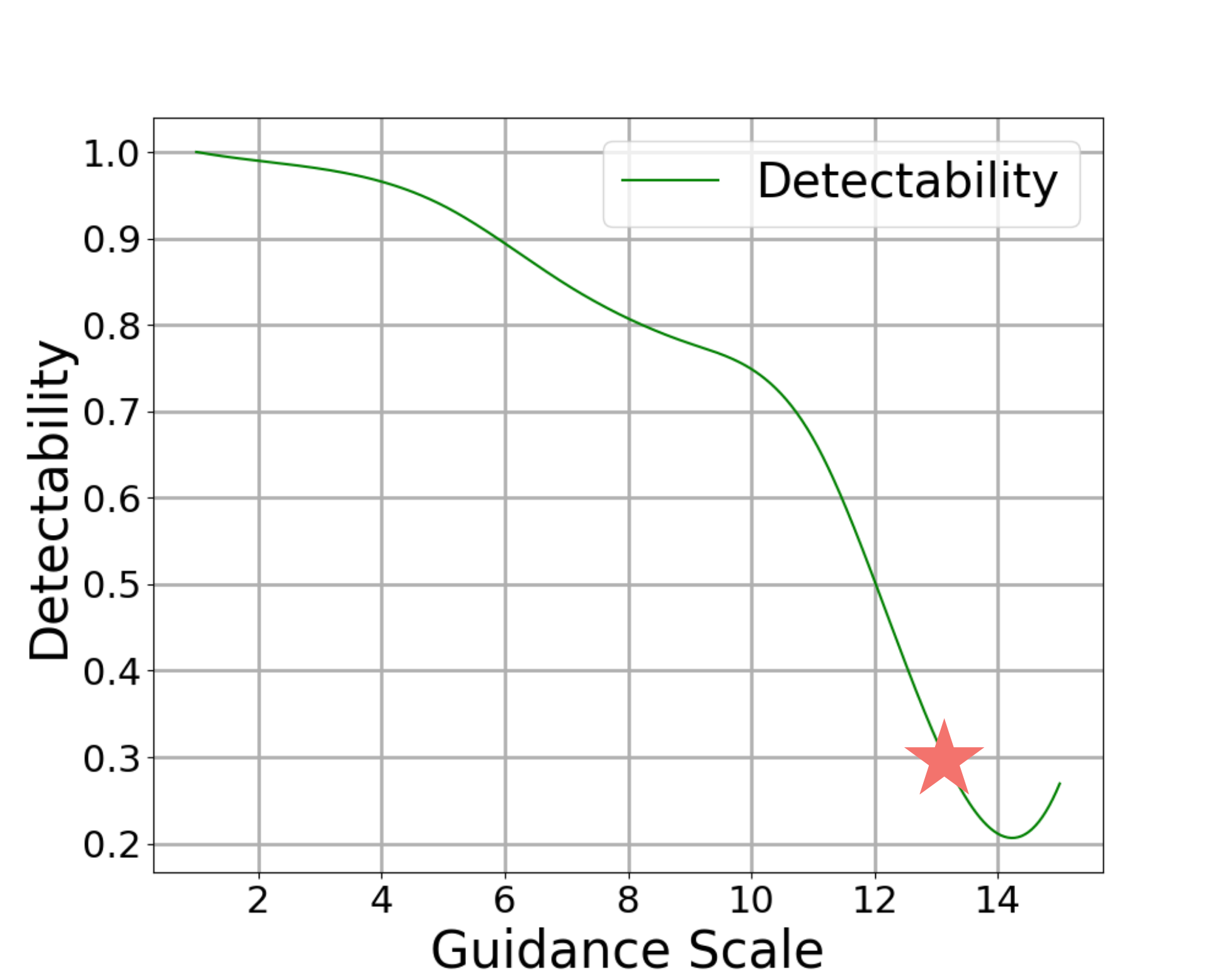}  \\

       \includegraphics[width=\imwidth]{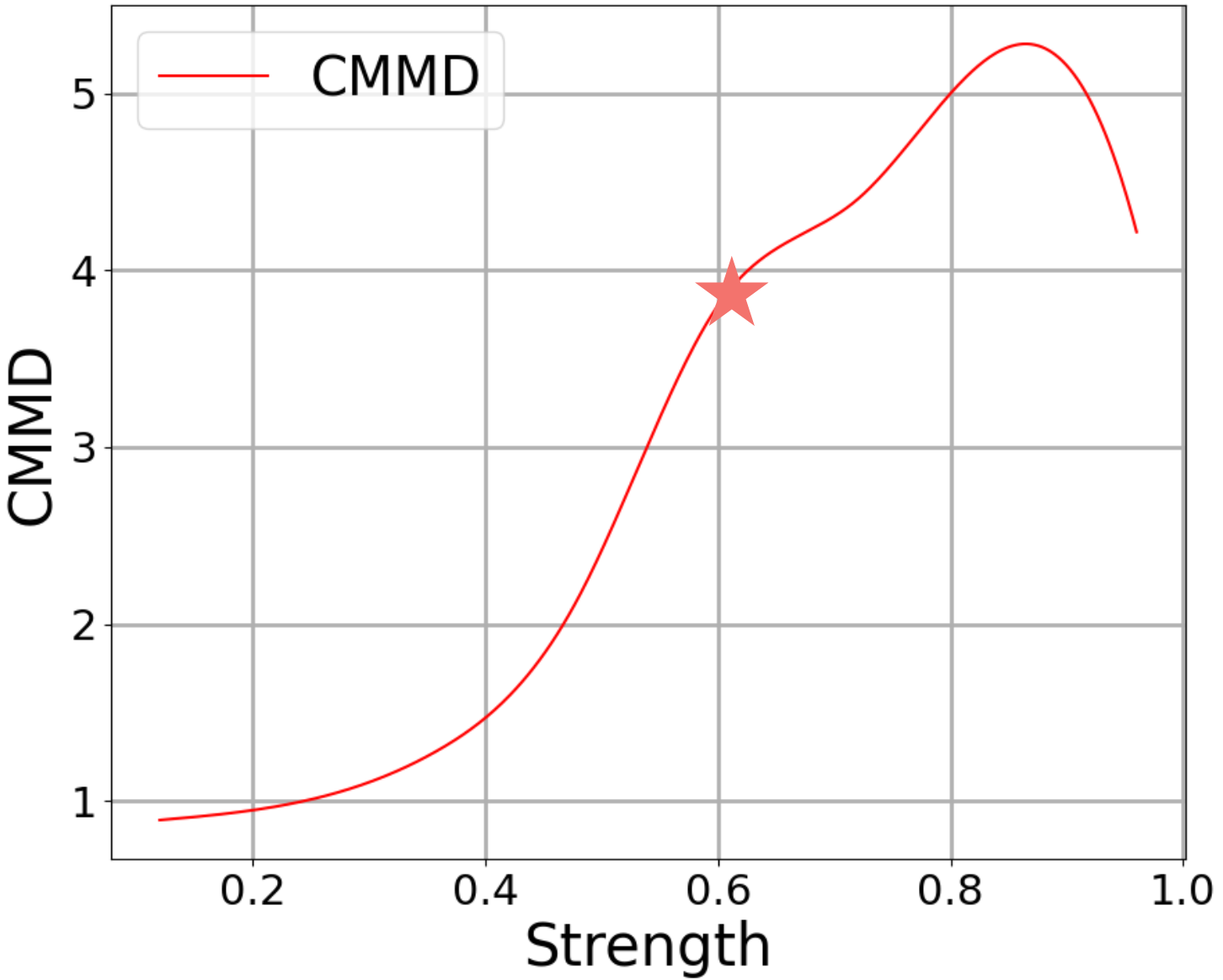} &
       \includegraphics[width=\imwidth]{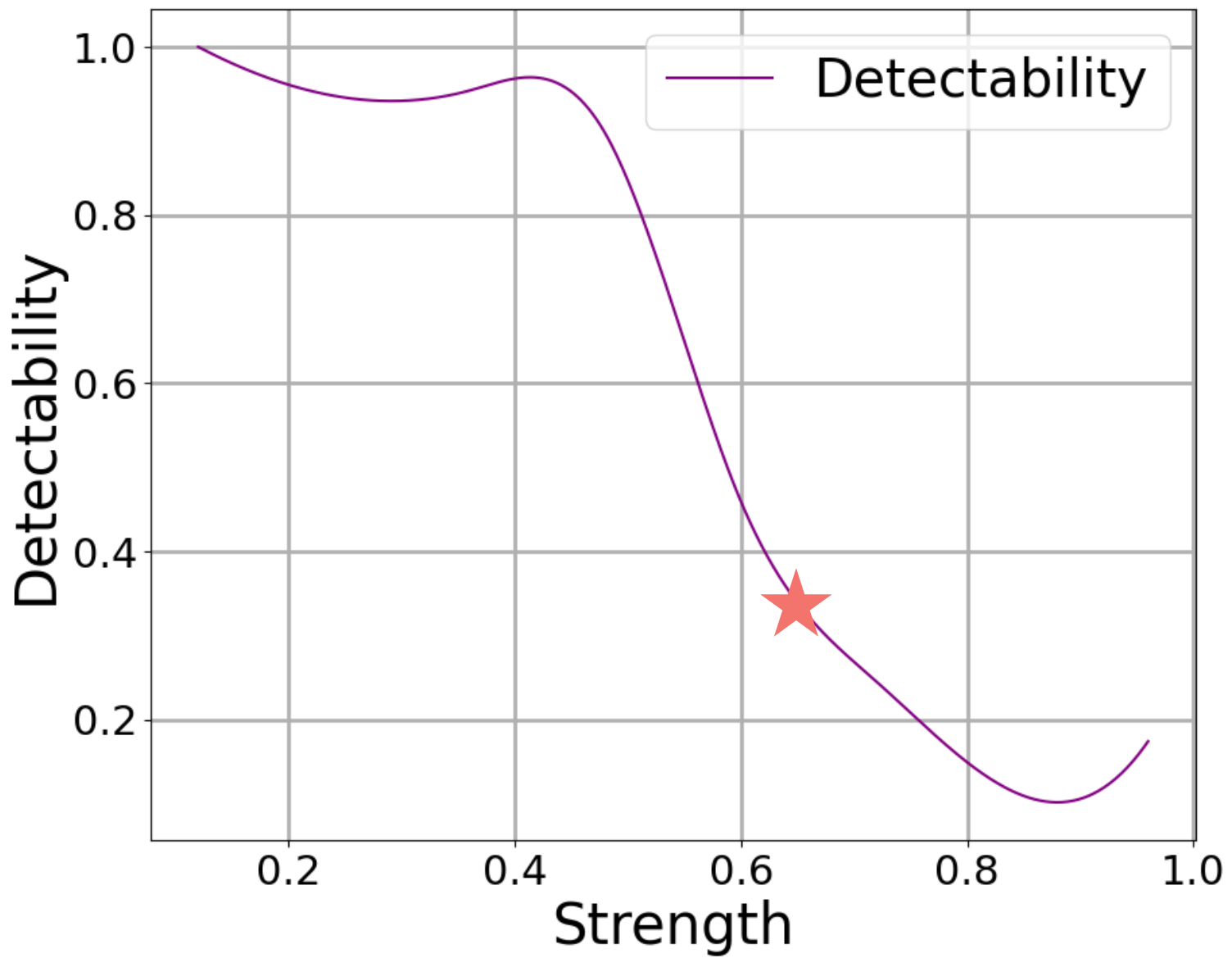} &
       \includegraphics[width=\imwidth]{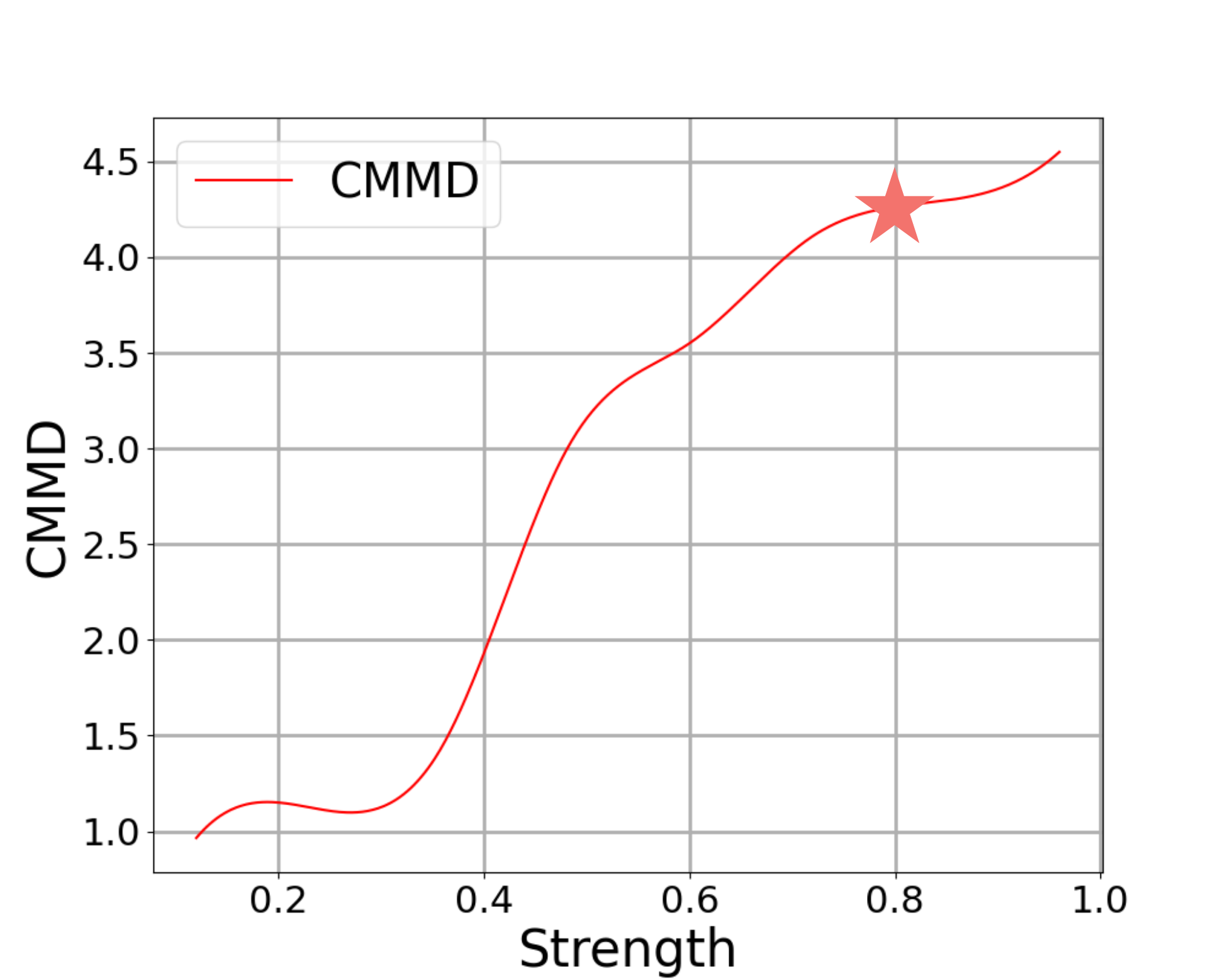} &
       \includegraphics[width=\imwidth]{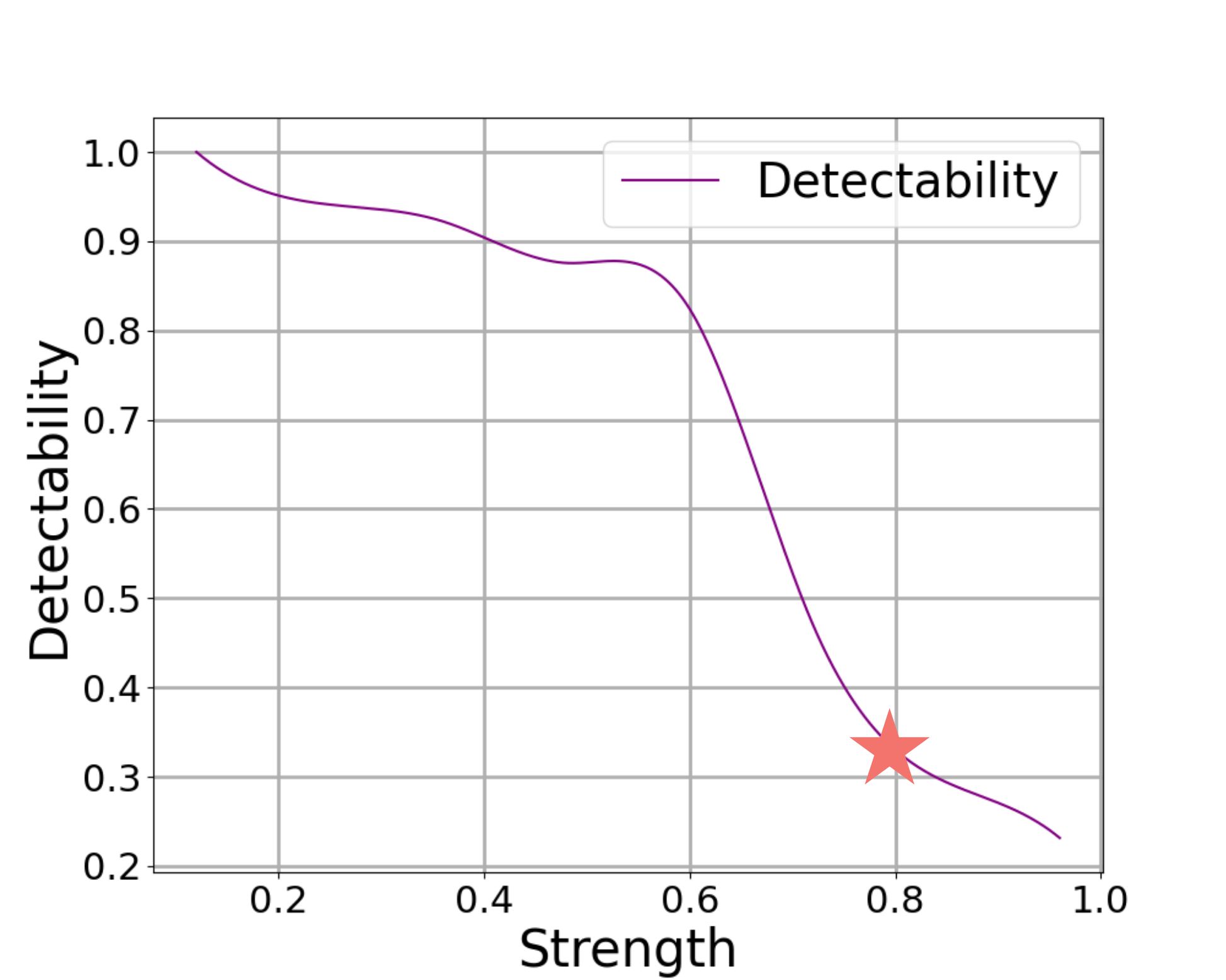} \\
       \multicolumn{2}{c}{
\begin{tcolorbox}[enhanced,attach boxed title to top left={yshift=-1mm,yshifttext=-1mm,xshift=8pt},left=1pt,right=1pt,top=1pt,bottom=1pt,colback=orange!5!white,colframe=orange!70!black,colbacktitle=orange!70!black,title=Observations,fonttitle=\ttfamily\bfseries\scshape\fontsize{8}{9}\selectfont,boxed title style={size=small,colframe=violet!50!black},width=0.5\textwidth]
\begin{itemize}
    \item For Zodiac, the least semantic distortion is observed at low strength values (below 0.5) and guidance scale values in the range of 5-9.
    \item The detectability significantly decreases as strength exceeds 0.5 and guidance scale surpasses 10.
\end{itemize}
\end{tcolorbox}
    } & 
    \multicolumn{2}{c}{
\begin{tcolorbox}[enhanced,attach boxed title to top left={yshift=-1mm,yshifttext=-1mm,xshift=8pt},left=1pt,right=1pt,top=1pt,bottom=1pt,colback=orange!5!white,colframe=orange!70!black,colbacktitle=orange!70!black,title=Observations,fonttitle=\ttfamily\bfseries\scshape\fontsize{8}{9}\selectfont,boxed title style={size=small,colframe=violet!50!black},width=0.5\textwidth]
\begin{itemize}
    \item The least semantic distortion in Gaussian Shading occurs for strength values under 0.3 and guidance scale values between 2-5.
    \item Detectability decreases with strength values over 0.6 and guidance scale values greater than 11.
\end{itemize}
\end{tcolorbox}
    } 
    \\
    \midrule
    \multicolumn{2}{c}{\textbf{DwtDctSVD}} & \multicolumn{2}{c}{\textbf{HiDDen}} \\
    \midrule
        \includegraphics[width=\imwidth]{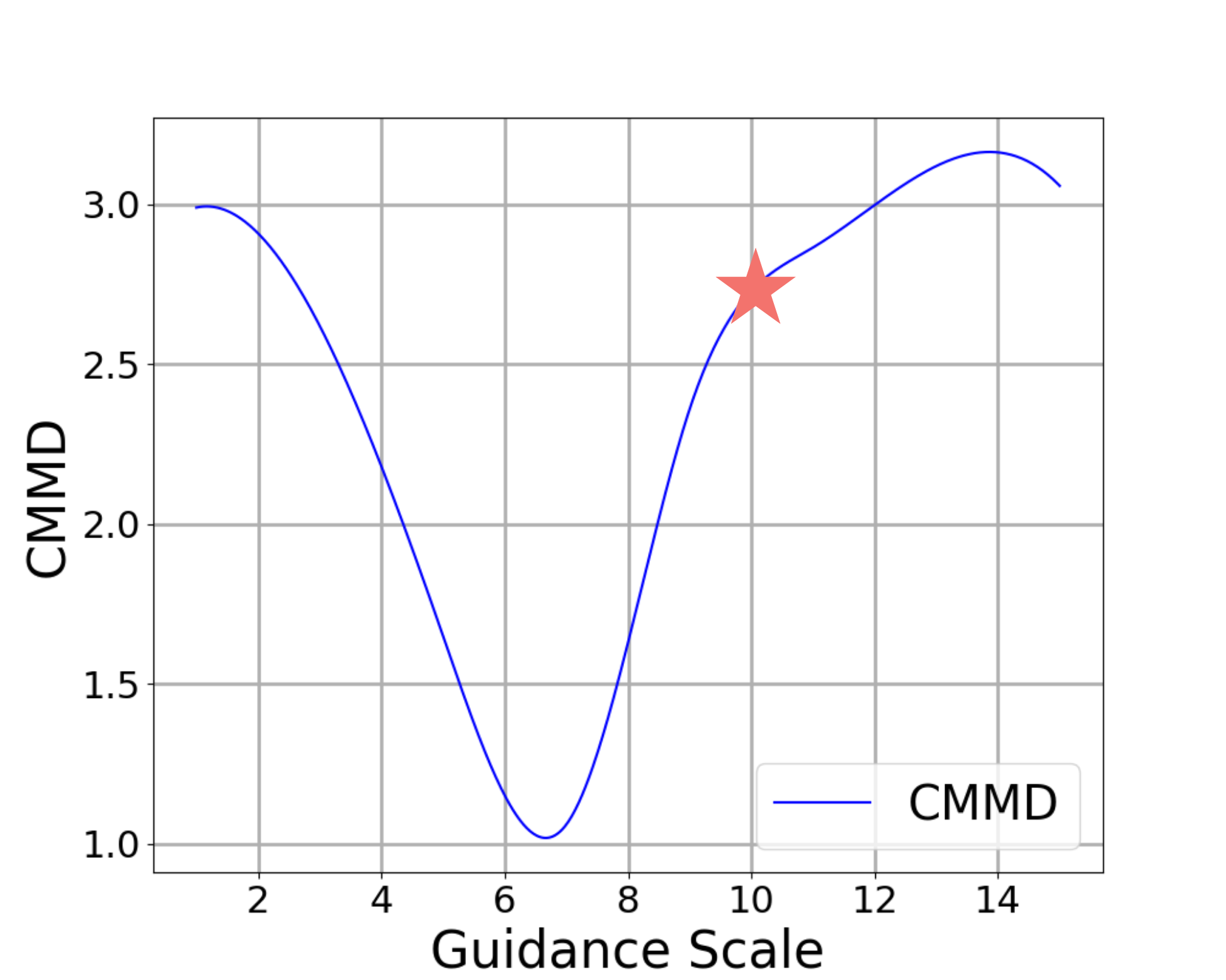} &
       \includegraphics[width=\imwidth]{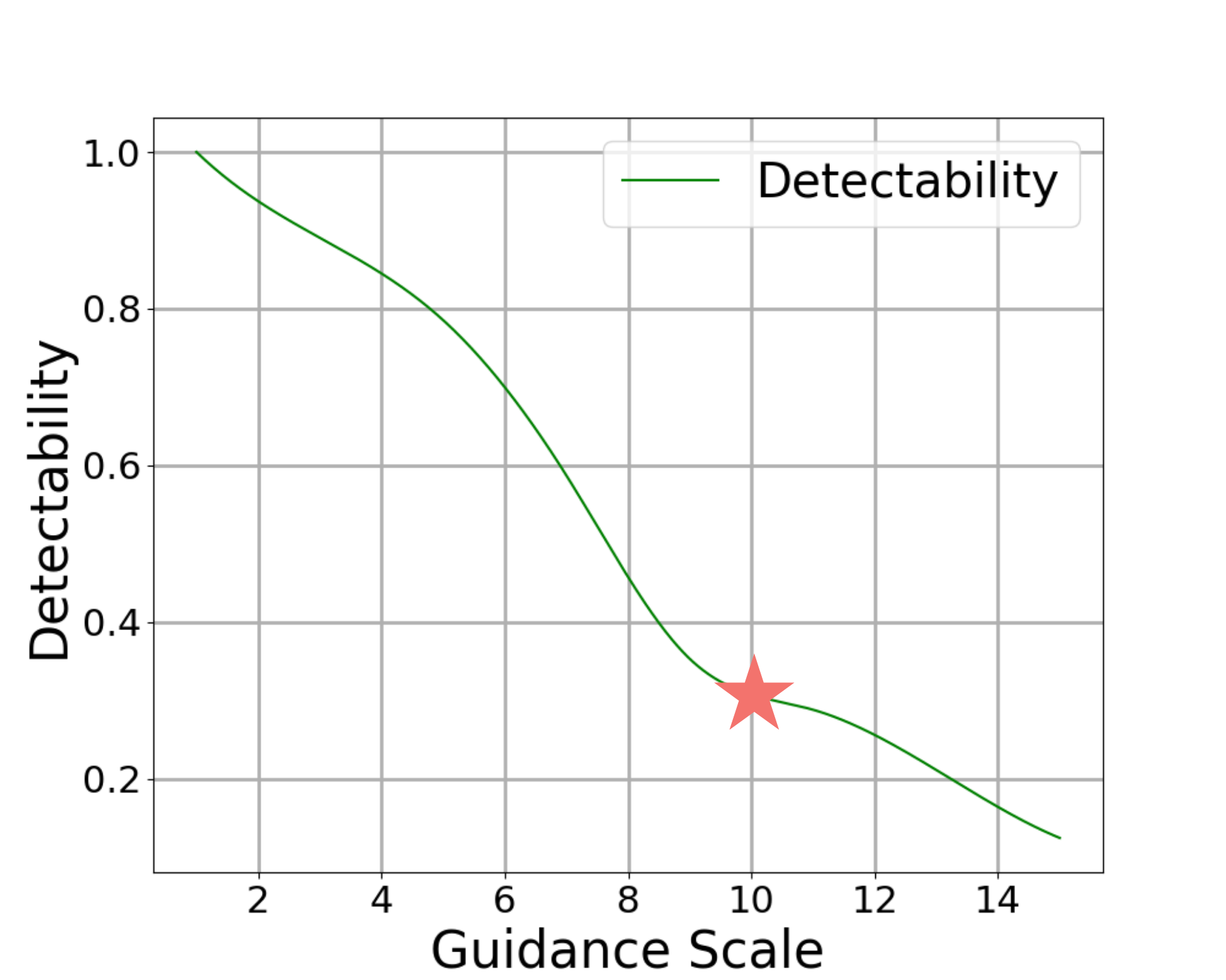} &
       \includegraphics[width=\imwidth]{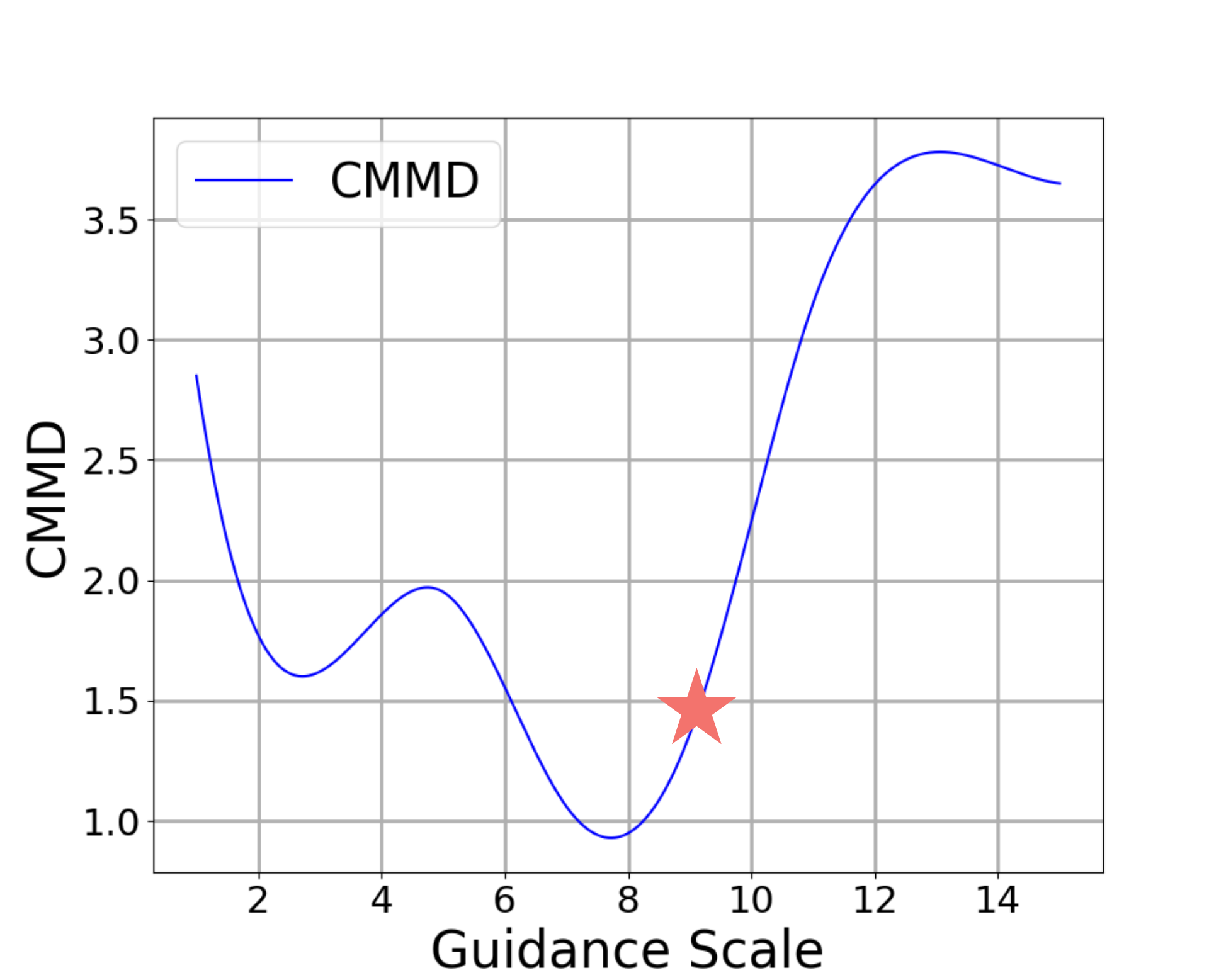} &
       \includegraphics[width=\imwidth]{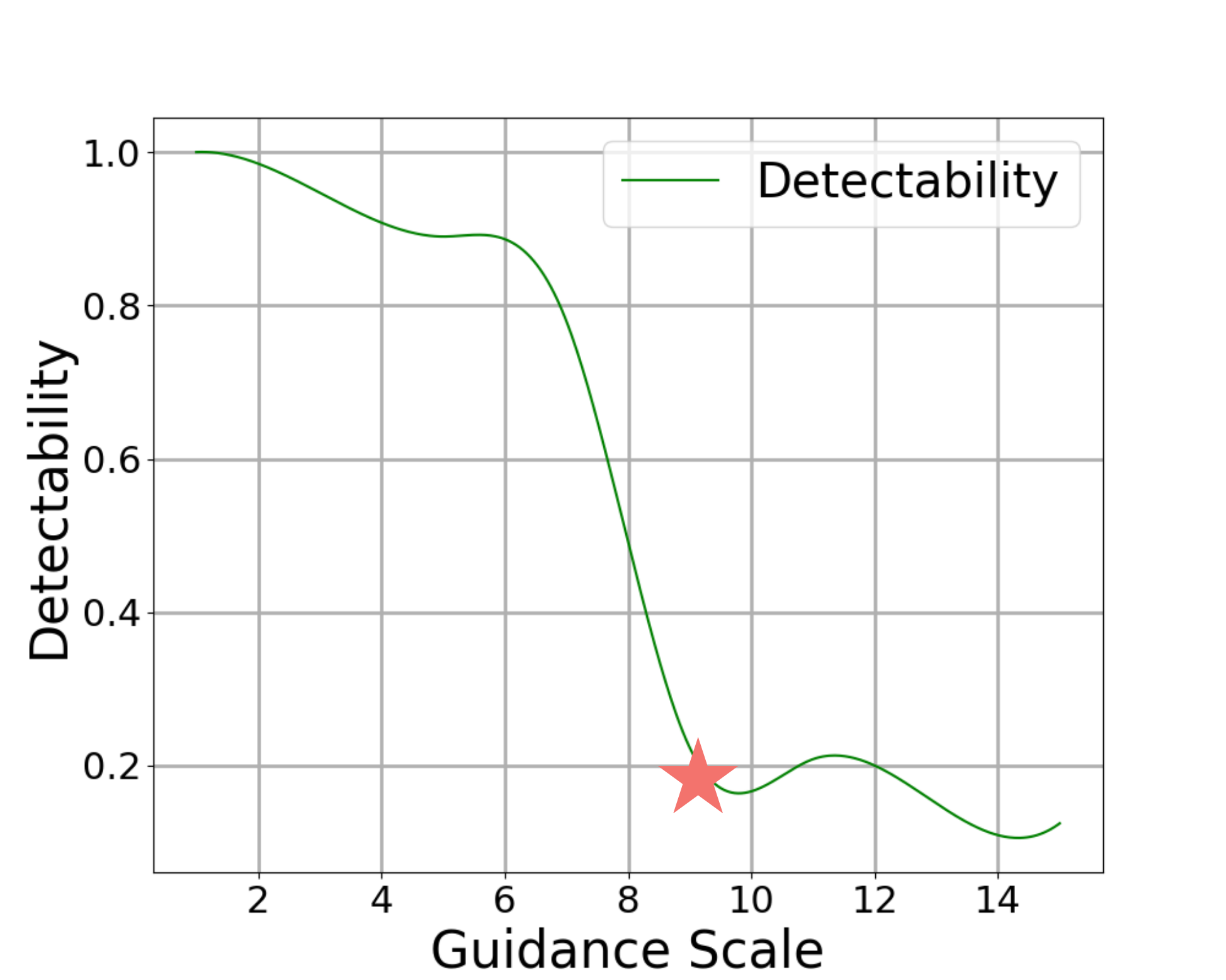}  \\

       \includegraphics[width=\imwidth]{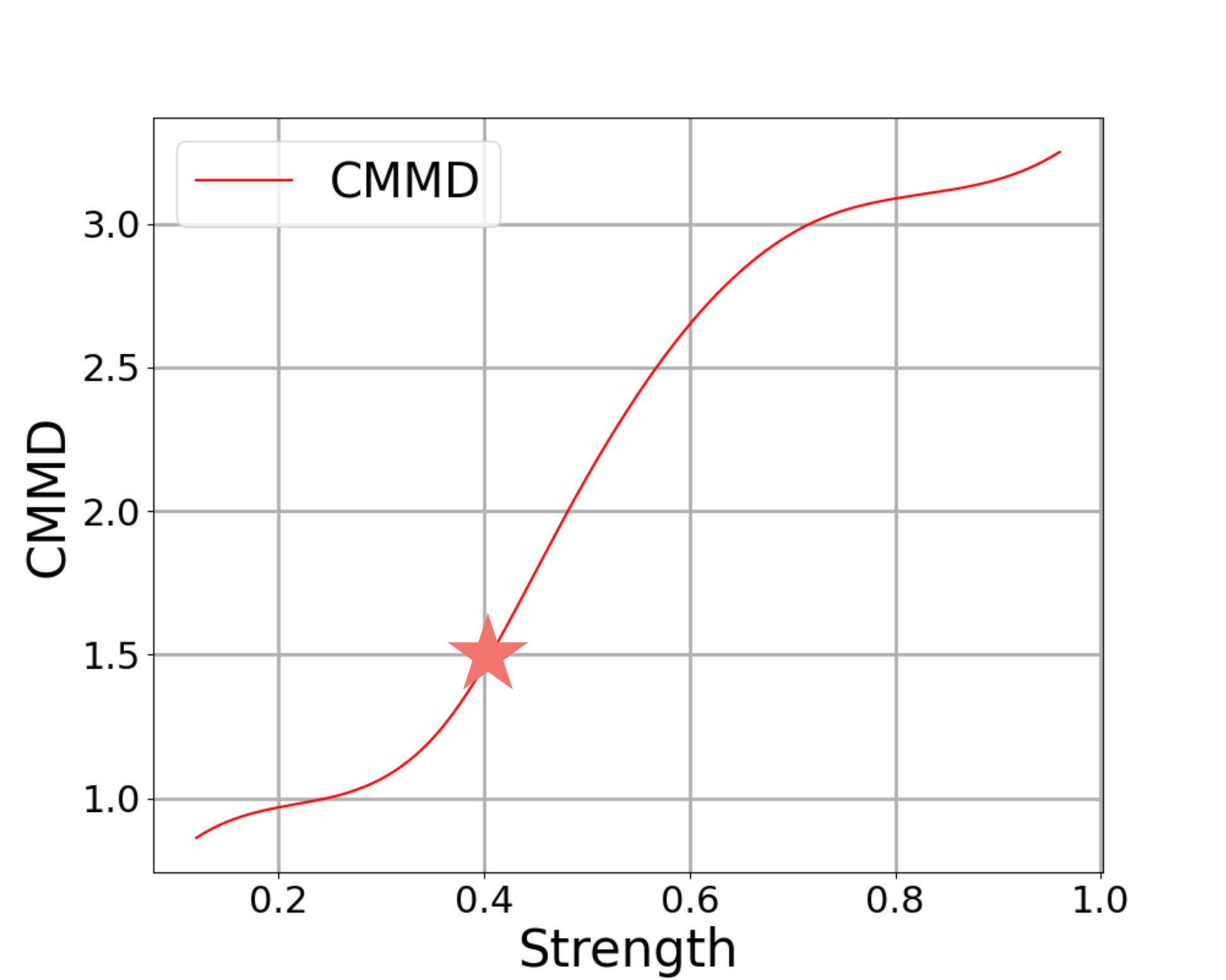} &
       \includegraphics[width=\imwidth]{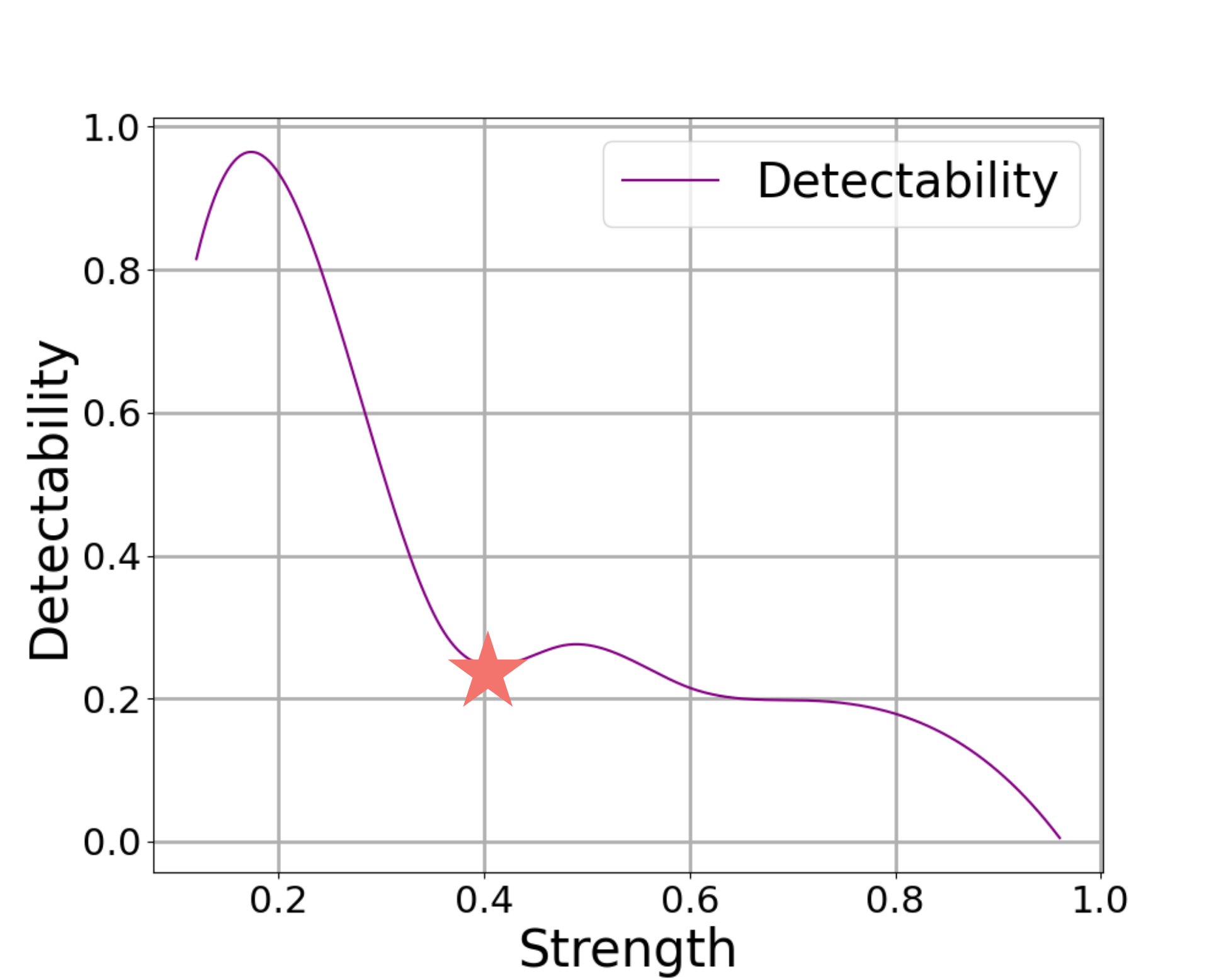} &
       \includegraphics[width=\imwidth]{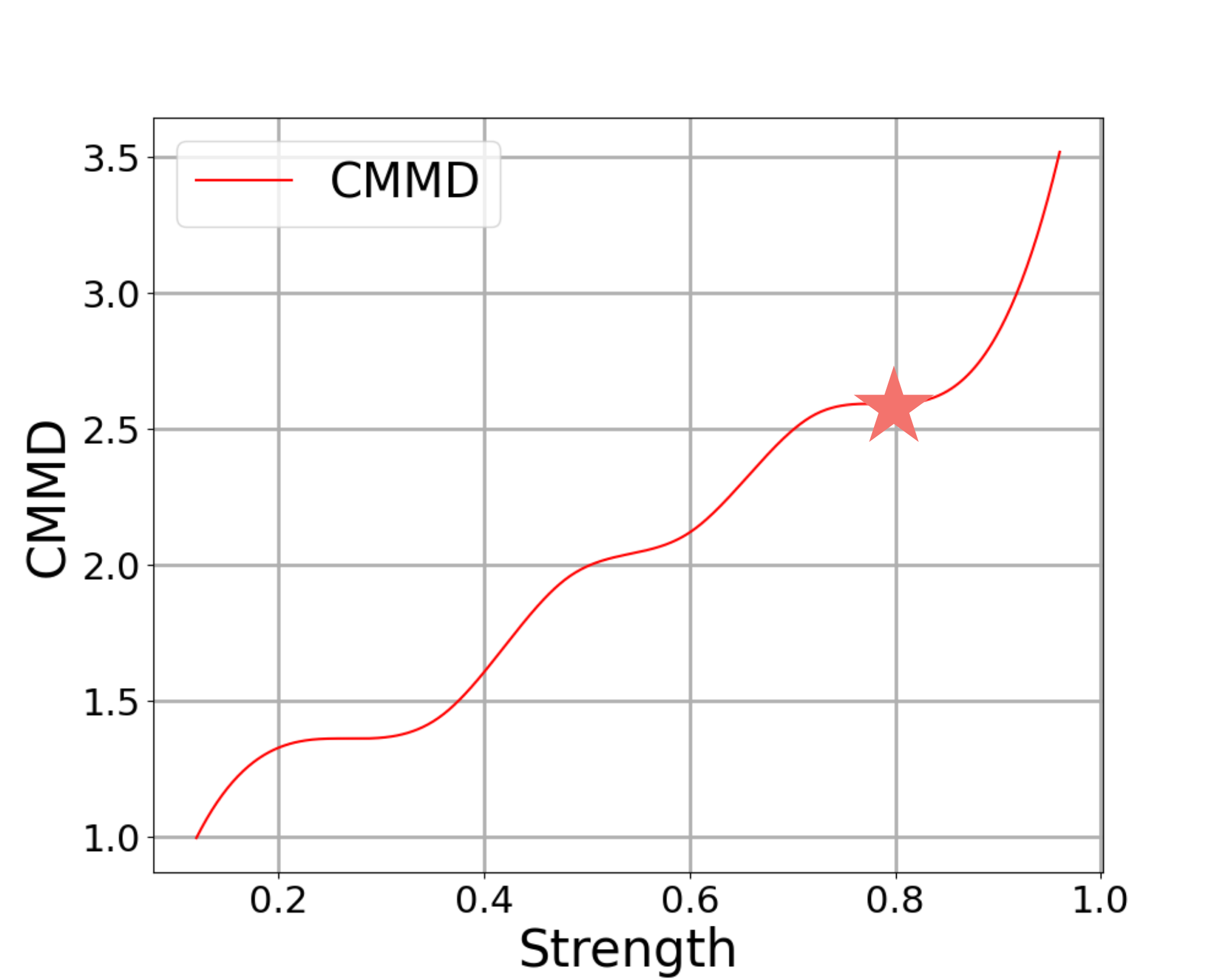} &
       \includegraphics[width=\imwidth]{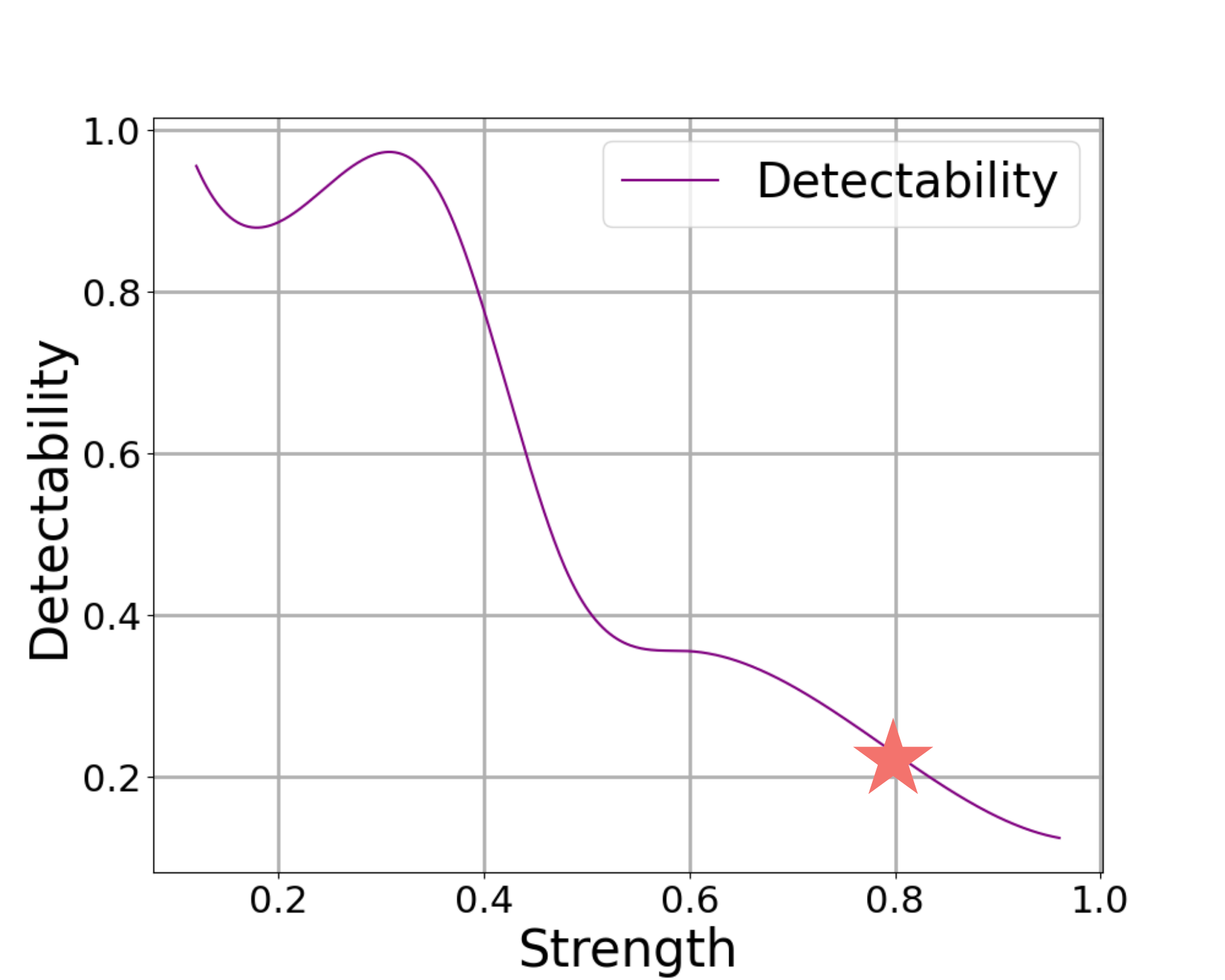} \\
       \multicolumn{2}{c}{
\begin{tcolorbox}[enhanced,attach boxed title to top left={yshift=-1mm,yshifttext=-1mm,xshift=8pt},left=1pt,right=1pt,top=1pt,bottom=1pt,colback=orange!5!white,colframe=orange!70!black,colbacktitle=orange!70!black,title=Observations,fonttitle=\ttfamily\bfseries\scshape\fontsize{8}{9}\selectfont,boxed title style={size=small,colframe=violet!50!black},width=0.5\textwidth]
\begin{itemize}
    \item The least semantic distortion for DwtDctSVD is noted at strength values below 0.5 and guidance scale values between 5-8.
    \item Detectability decreases as strength crosses 0.3 and guidance scale goes beyond 7.
\end{itemize}
\end{tcolorbox}
    } & 
       \multicolumn{2}{c}{
\begin{tcolorbox}[enhanced,attach boxed title to top left={yshift=-1mm,yshifttext=-1mm,xshift=8pt},left=1pt,right=1pt,top=1pt,bottom=1pt,colback=orange!5!white,colframe=orange!70!black,colbacktitle=orange!70!black,title=Observations,fonttitle=\ttfamily\bfseries\scshape\fontsize{8}{9}\selectfont,boxed title style={size=small,colframe=violet!50!black},width=0.5\textwidth]
\begin{itemize}
    \item For HiDDeN, the least semantic distortion occurs at strength values under 0.3 and guidance scale values around 5-9.
    \item Detectability decreases when strength exceeds 0.4 and guidance scale values go above 8.
\end{itemize}
\end{tcolorbox}
    }
    \\
       \bottomrule    
    \end{tabular}
    }
    
\caption{
This figure shows the variation of CMMD \cite{jayasumana2024rethinking} and detectability of visual paraphrases with respect to strength and guidance scale. \includegraphics[width=3mm]{img/star_pointer.png} represents the optimal $s$ and $gs$ value for the particular technique. 
% The images were watermarked using Tree Ring Watermarking \cite{wen2023treering}, Stable Signature \cite{fernandez2023stable}, Zodiac \cite{zhang2024zodiac}, Gaussian Shading \cite{yang2024gaussian}, DwtDctSVD \cite{navas2008dwt} and HiDDen \cite{zhu2018hidden}.
}
\label{fig:strength_gs_plots_all}
\end{figure*}

Figure \ref{fig:strength_gs_plots_all} illustrates the relationship between the CLIP Maximum Mean Discrepancy (CMMD) score and the detectability of visual paraphrases as influenced by variations in strength and guidance scale. 
% It highlights how changes in these parameters impact both the perceptual quality and the robustness of watermark detection.

% \begin{figure*}[h]
%     \centering
%     \resizebox{\linewidth}{!}{
%     \begin{tabular}{l}
%         \begin{minipage}{\linewidth}\centering \includegraphics[width=0.9\textwidth,height=\textheight,keepaspectratio]{img/comparison_vp_watermark.pdf}
%         \end{minipage}
%     \end{tabular}

% } 

% \caption{ 
% This figure illustrates the critical differences between a watermarked image and its visual paraphrased counterpart. The watermark detectability scores are provided for key areas of the image, demonstrating how visual paraphrasing can significantly reduce the detectability of the watermark
% }
% \label{fig:Mos_visual}
% \end{figure*}

\clearpage
\subsection{Paraphrase Acceptability in MOS Evaluation}

\begin{figure*}[h]
    \centering
    \resizebox{\linewidth}{!}{
    \begin{tabular}{l}
        \begin{minipage}{\linewidth}\centering \includegraphics[width=0.8\textwidth,height=\textheight,keepaspectratio]{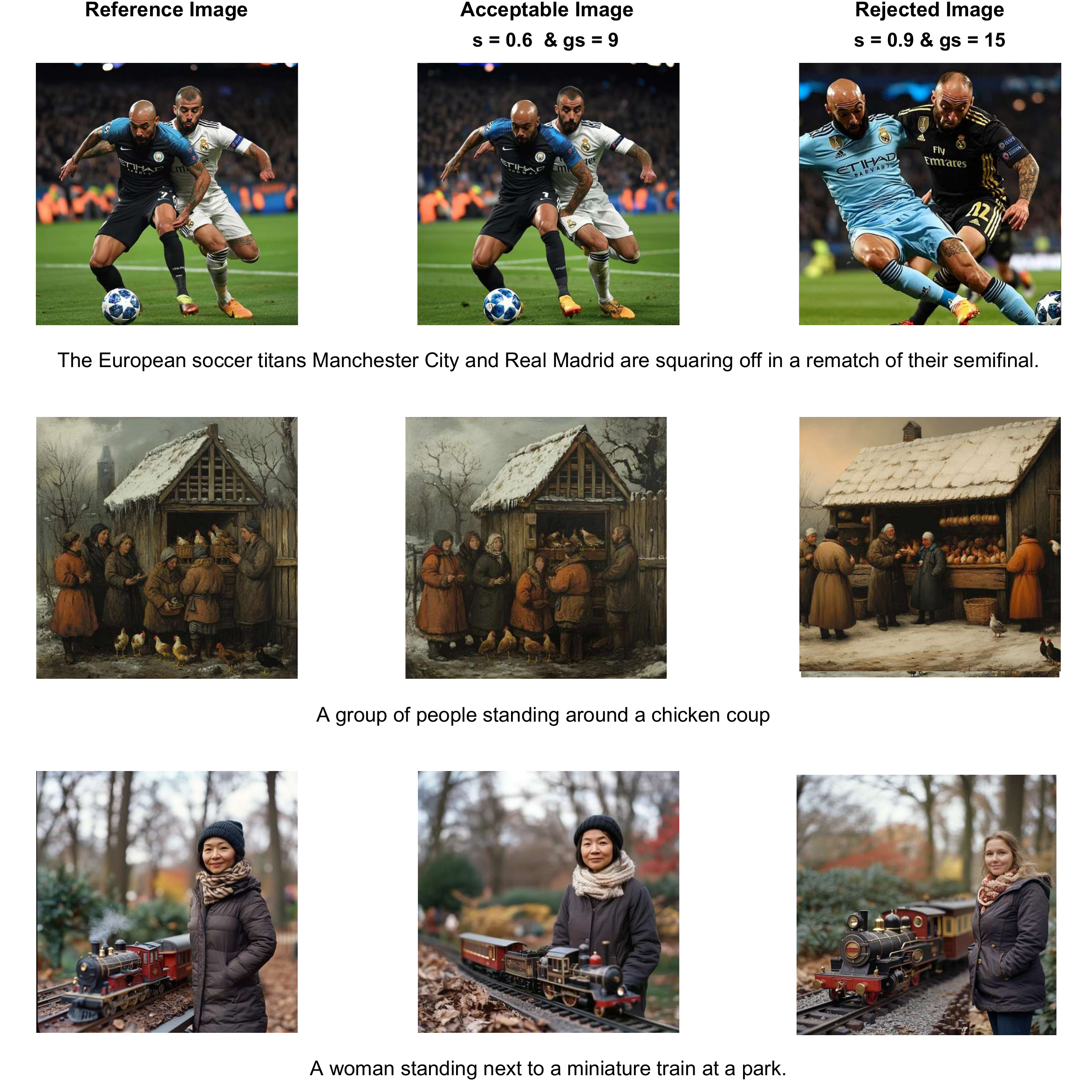}
        \end{minipage}
    \end{tabular}

} 

\caption{ 
Examples of acceptable and rejected Visual Paraphrasing during MOS evaluation.
}
\label{fig:Mos_visual}
\end{figure*}

Figure \ref{fig:Mos_visual} presents a set of visual examples illustrating both accepted and rejected paraphrases during the MOS (Mean Opinion Score) evaluation. These examples highlight the differences in image quality and semantic consistency that led to their respective ratings. Accepted paraphrases maintain a high degree of similarity to the original image while preserving key visual and contextual elements. In contrast, rejected paraphrases exhibit significant deviations that detract from the original image's meaning or visual quality, resulting in lower MOS ratings. This comparison underscores the criteria used by human evaluators to assess the acceptability of visual paraphrases.

\clearpage

\subsection{Watermark Robustness Under Various Attacks}
    This section provides a comparative analysis of watermarked images subjected to various attacks, including Brightness Adjustment, Rotation, JPEG Compression, and Gaussian Noise, as well as our Visual Paraphrase method. The accompanying figure illustrates the impact of each attack on the integrity and detectability of the watermark, with $\eta$ comparisons (watermark detection scores) presented for Stable Signature, ZoDiac, and HiDDeN. Other techniques are not discussed here, as they cannot watermark an already generated image. These comparisons underscore the resilience of the watermark under different conditions and demonstrate the effectiveness of our Visual Paraphrase method in altering the image while potentially preserving or bypassing the watermark. Tables \ref{fig:attack_vp_comparison_1} through \ref{fig:attack_vp_comparison_6} provide a more detailed analysis, offering deeper insights into how various attacks influence watermark robustness and detection.
\begin{table*}[h!]
\centering
    \scriptsize
    \newcommand{\imwidth}{0.165\textwidth}
        \setlength{\tabcolsep}{0pt}
        \resizebox{!}{8.5cm}{
        \begin{tabular}{cccccc}
        \toprule
        Watermarked & Brightness & Rotation & JPEG Compression & Gaussian Noise & \textbf{Visual Paraphrase (Ours)} \\
        % \midrule
        \toprule
           \includegraphics[width=\imwidth]{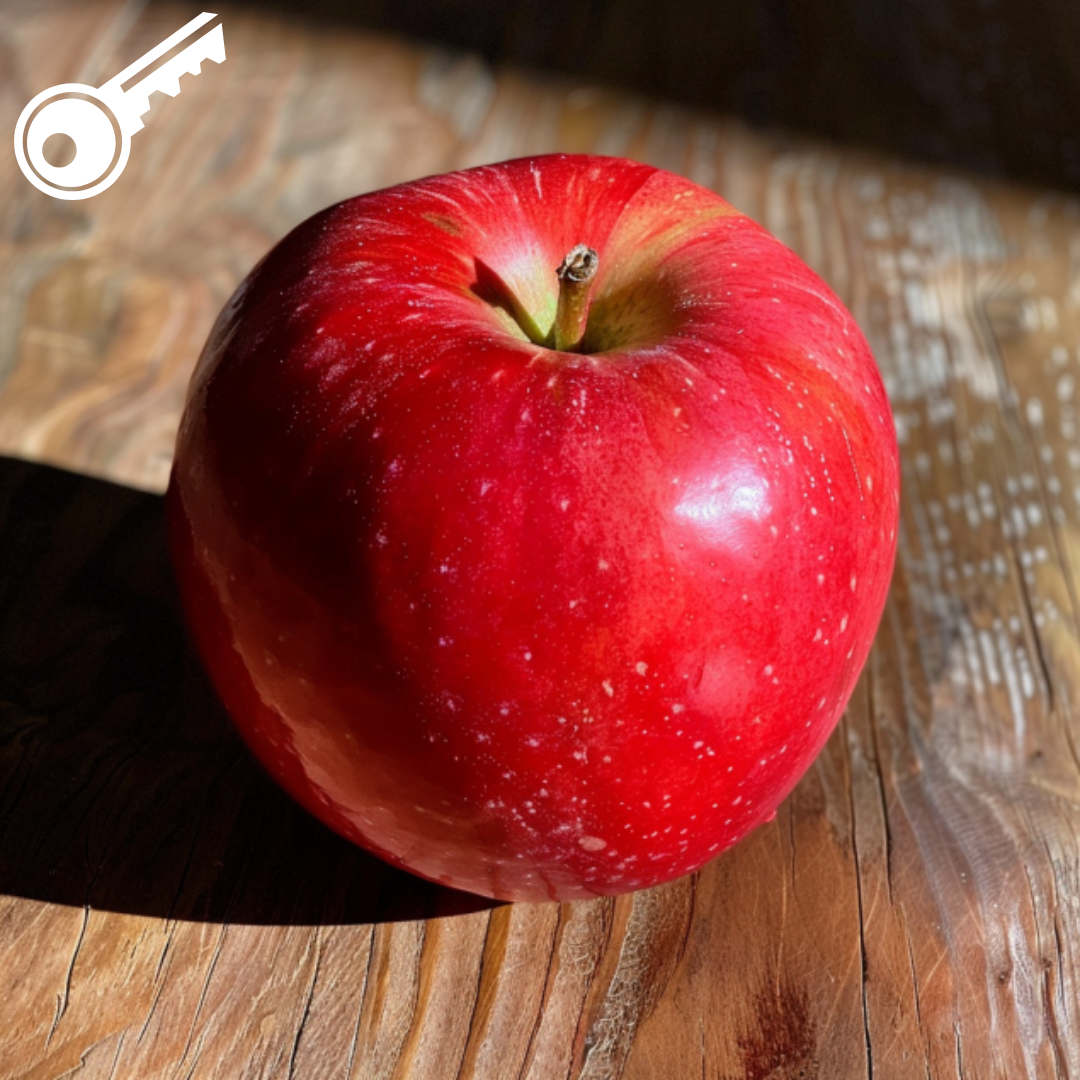} &
           \includegraphics[width=\imwidth]{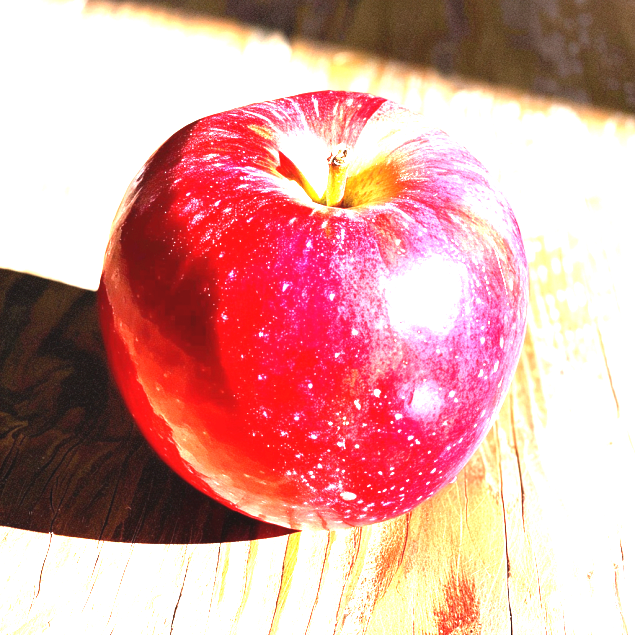} &
           \includegraphics[width=\imwidth]{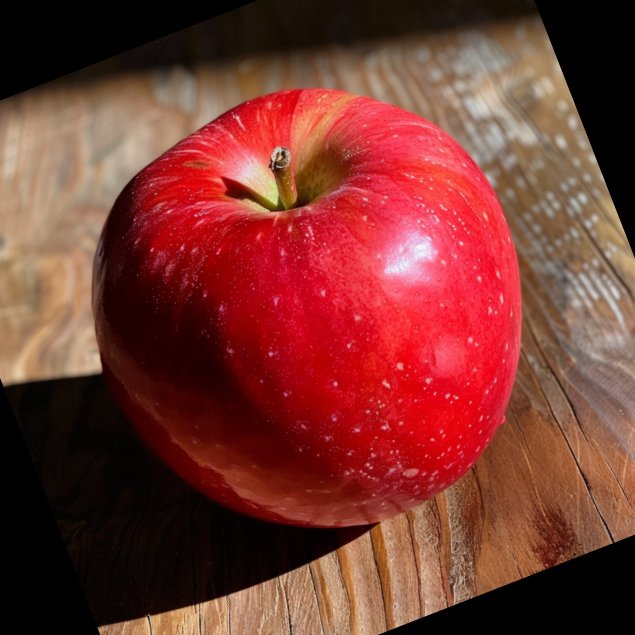} &
           \includegraphics[width=\imwidth]{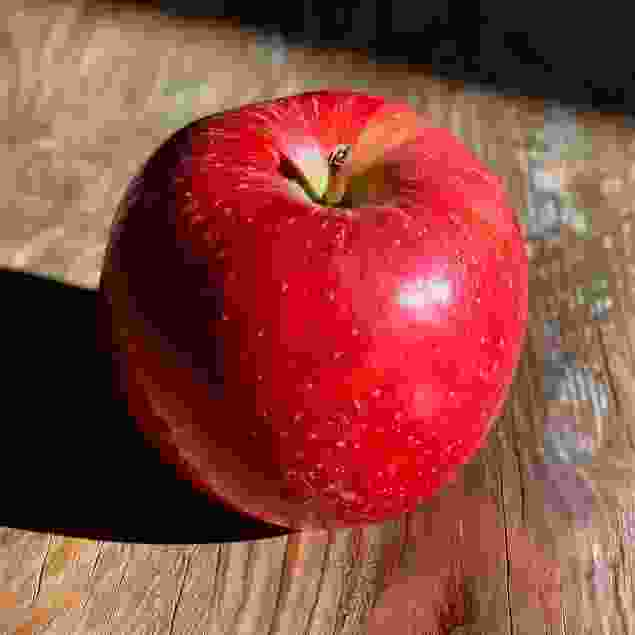} & 
           \includegraphics[width=\imwidth]{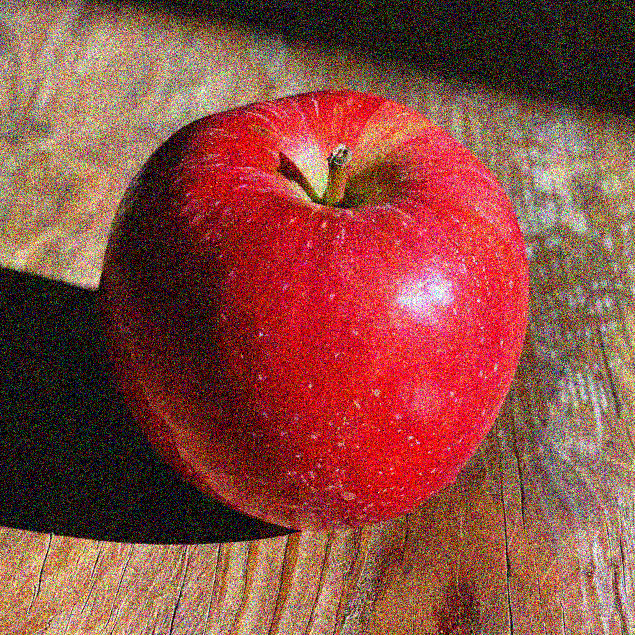} &
           \includegraphics[width=\imwidth]{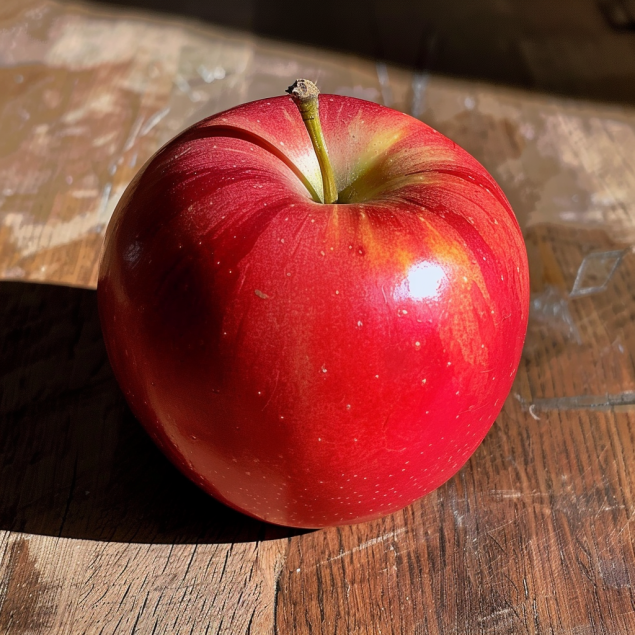} \\
            $\eta$ = 1 & $\eta$ = 0.989 & $\eta$ = 0.841 & $\eta$ = 0.624 & $\eta$ = 0.671 & $\eta$ = 0.263 \\
           
           \rule{0pt}{8ex}%
    
           \includegraphics[width=\imwidth]{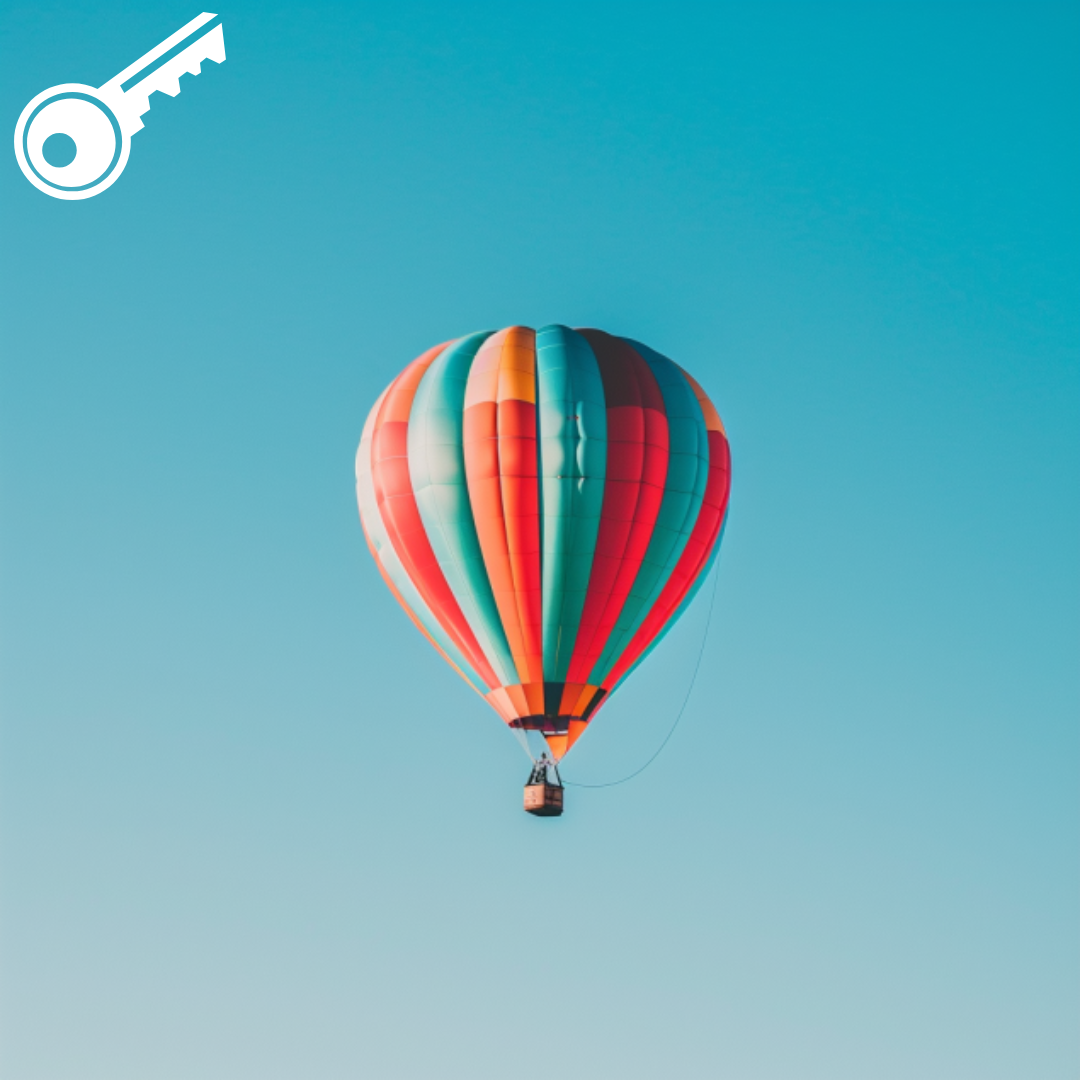} &
           \includegraphics[width=\imwidth]{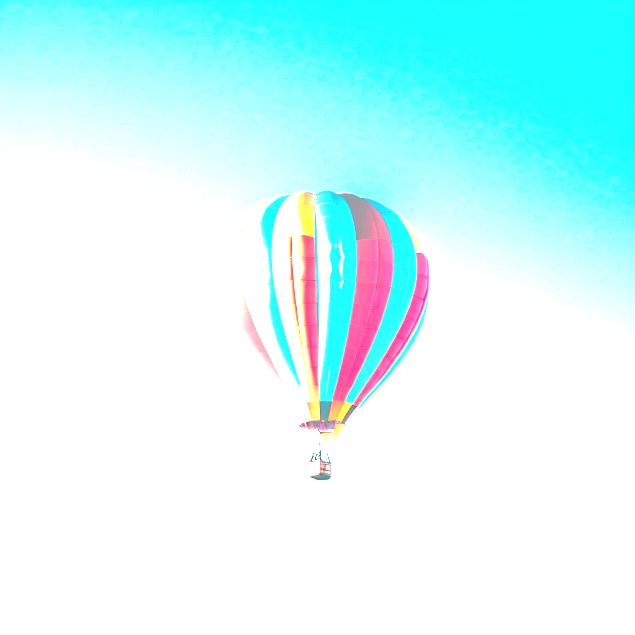} &
           \includegraphics[width=\imwidth]{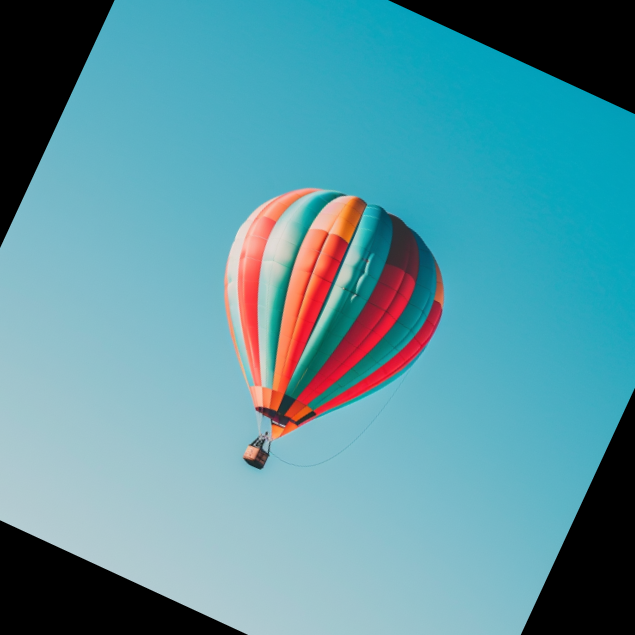} &
           \includegraphics[width=\imwidth]{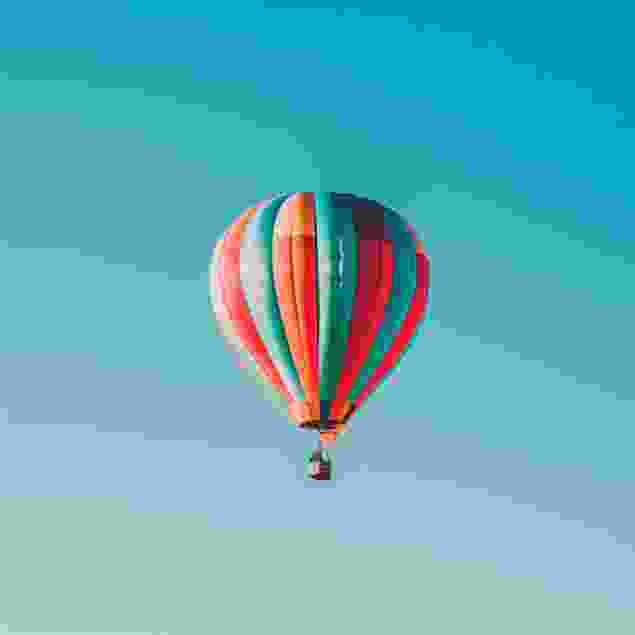} &
           \includegraphics[width=\imwidth]{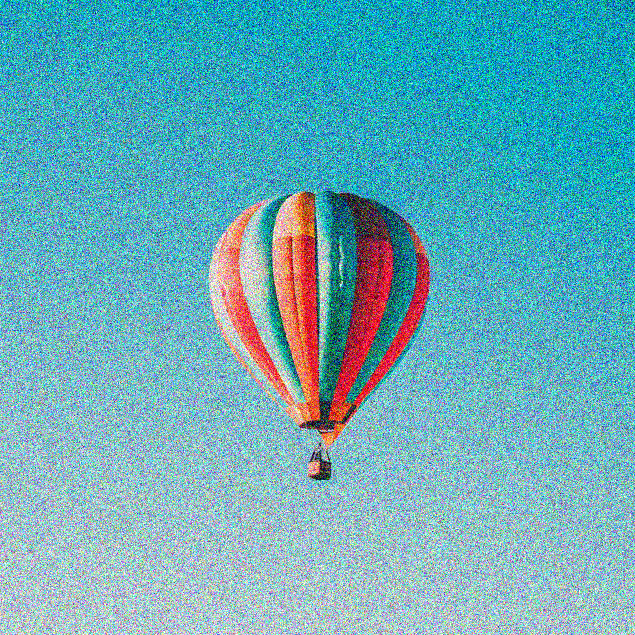} &
           \includegraphics[width=\imwidth]{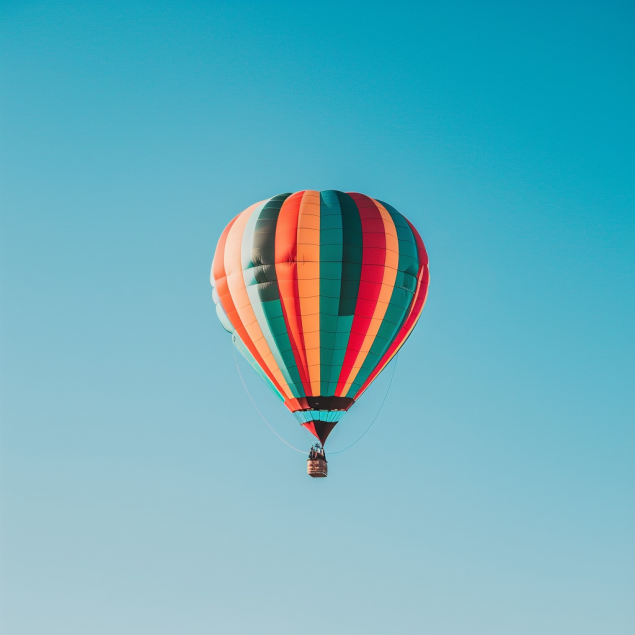} \\
            $\eta$ = 1 & $\eta$ = 0.991 & $\eta$ = 0.813 & $\eta$ = 0.611 & $\eta$ = 0.633 & $\eta$ = 0.334 \\
           \rule{0pt}{8ex}%
    
           \includegraphics[width=\imwidth]{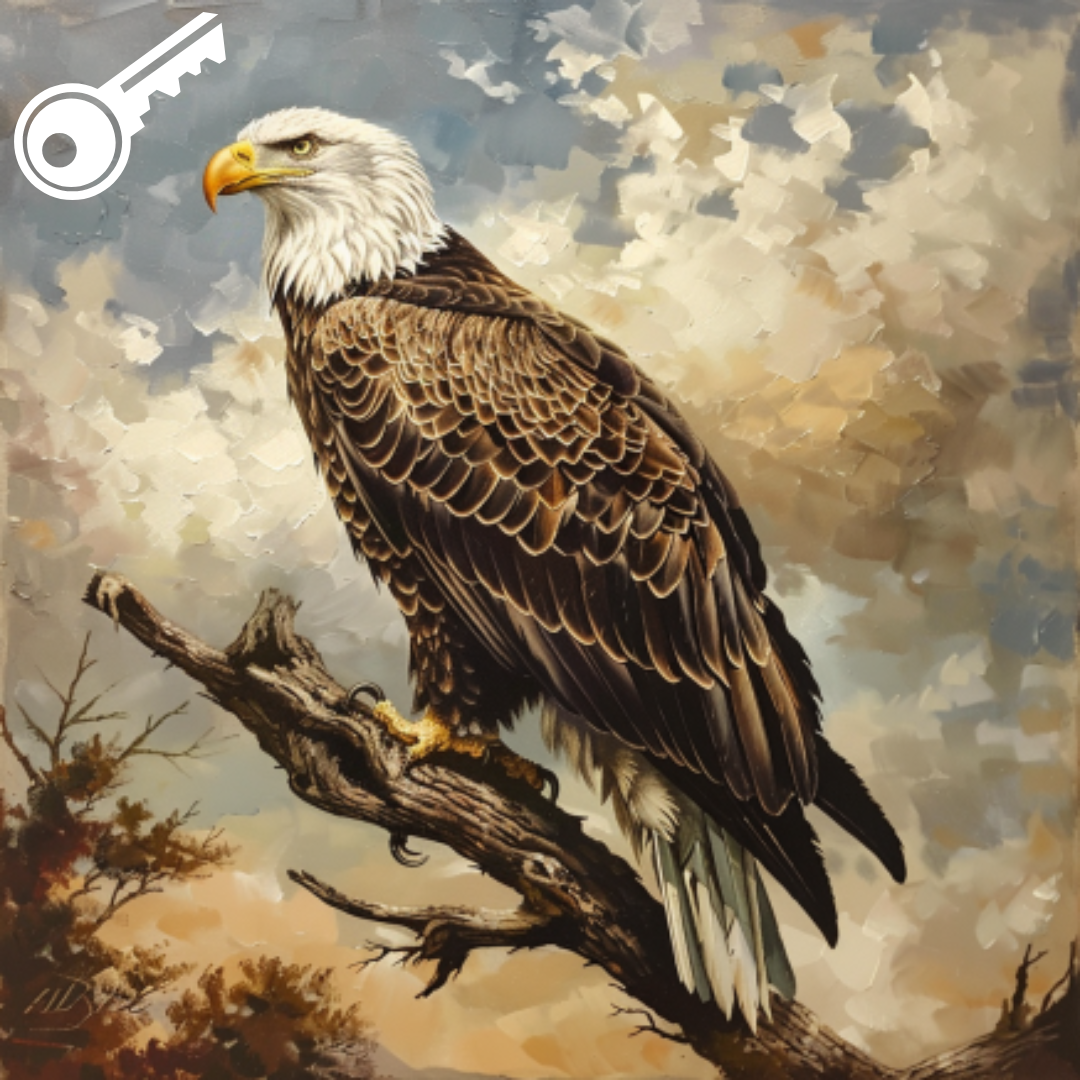} &
           \includegraphics[width=\imwidth]{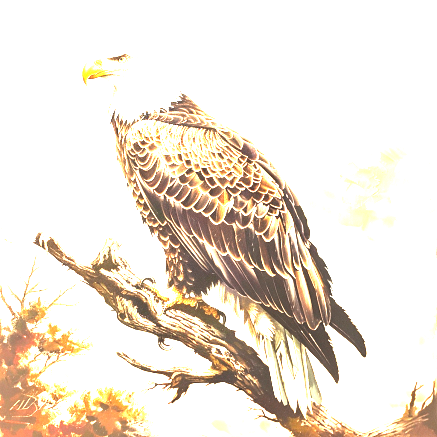} &
           \includegraphics[width=\imwidth]{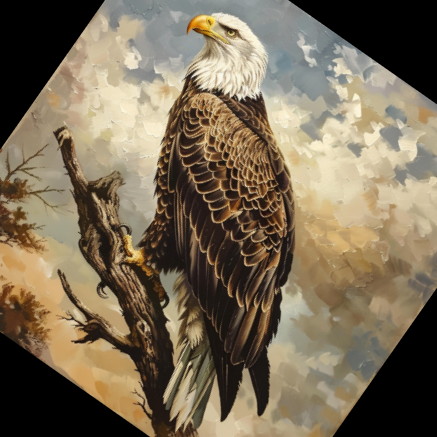} &
           \includegraphics[width=\imwidth]{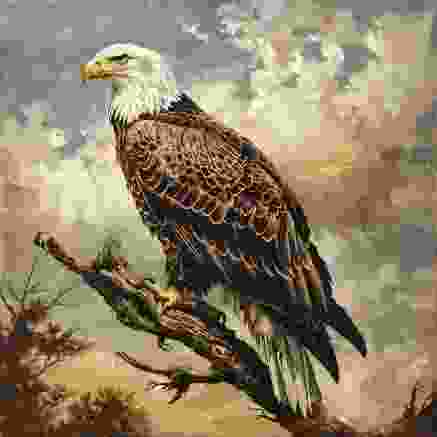} &
           \includegraphics[width=\imwidth]{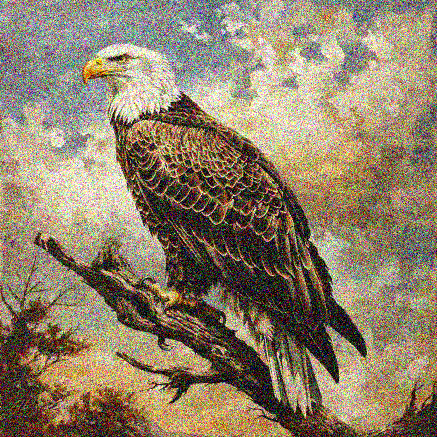} &
           \includegraphics[width=\imwidth]{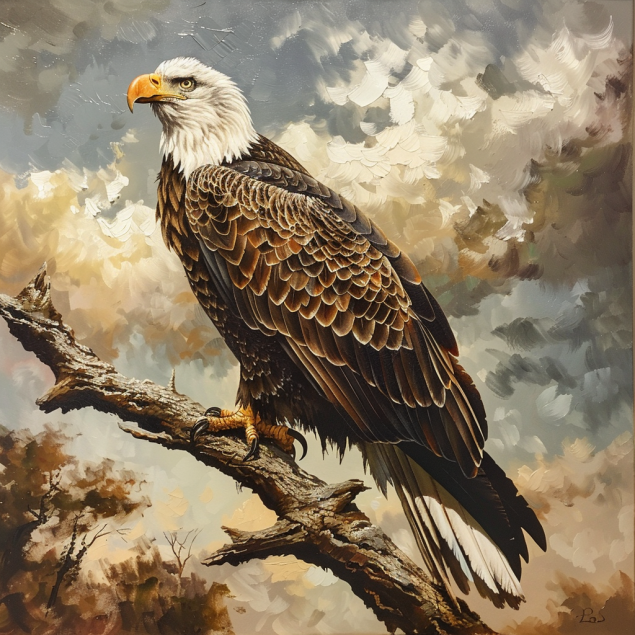} \\
            $\eta$ = 1 & $\eta$ = 0.984 & $\eta$ = 0.837 & $\eta$ = 0.656 & $\eta$ = 0.603 & $\eta$ = 0.297 \\
           \rule{0pt}{8ex}%
    
            \includegraphics[width=\imwidth]{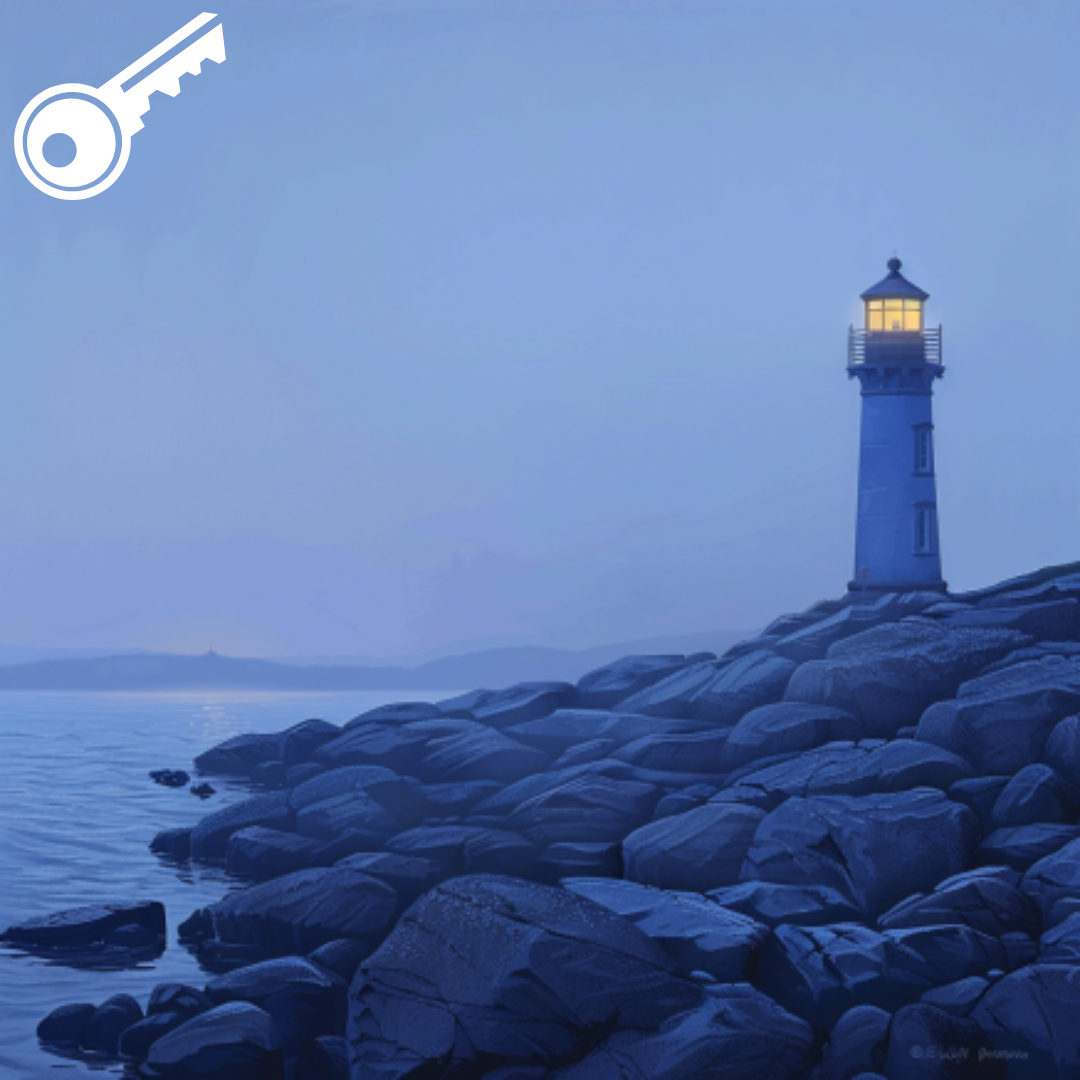} &
           \includegraphics[width=\imwidth]{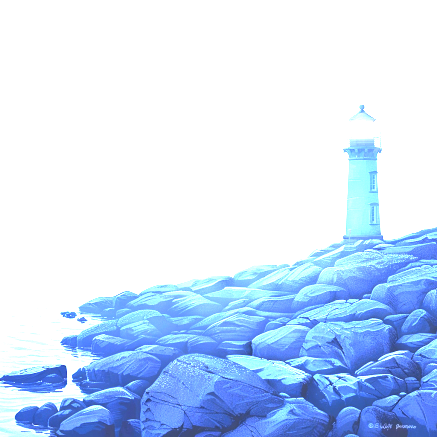} &
           \includegraphics[width=\imwidth]{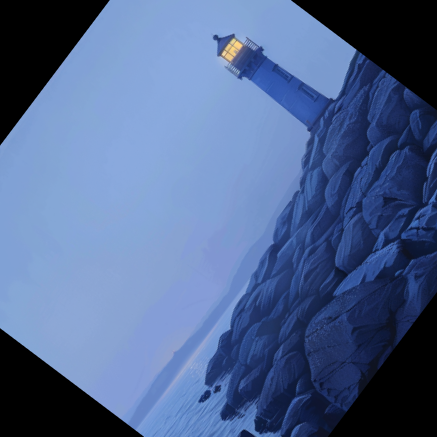} &
           \includegraphics[width=\imwidth]{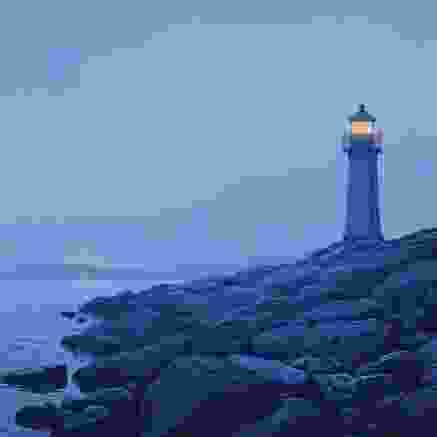} & 
           \includegraphics[width=\imwidth]{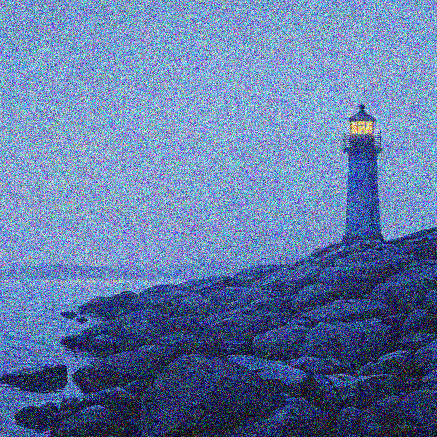} &
           \includegraphics[width=\imwidth]{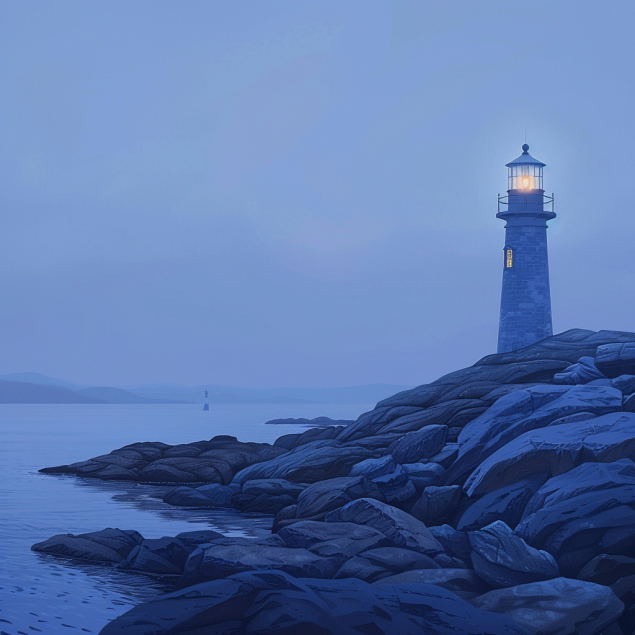} \\
            $\eta$ = 1 & $\eta$ = 0.994 & $\eta$ = 0.784 & $\eta$ = 0.609 & $\eta$ = 0.579 & $\eta$ = 0.273 \\
           \rule{0pt}{8ex}%
    
           \includegraphics[width=\imwidth]{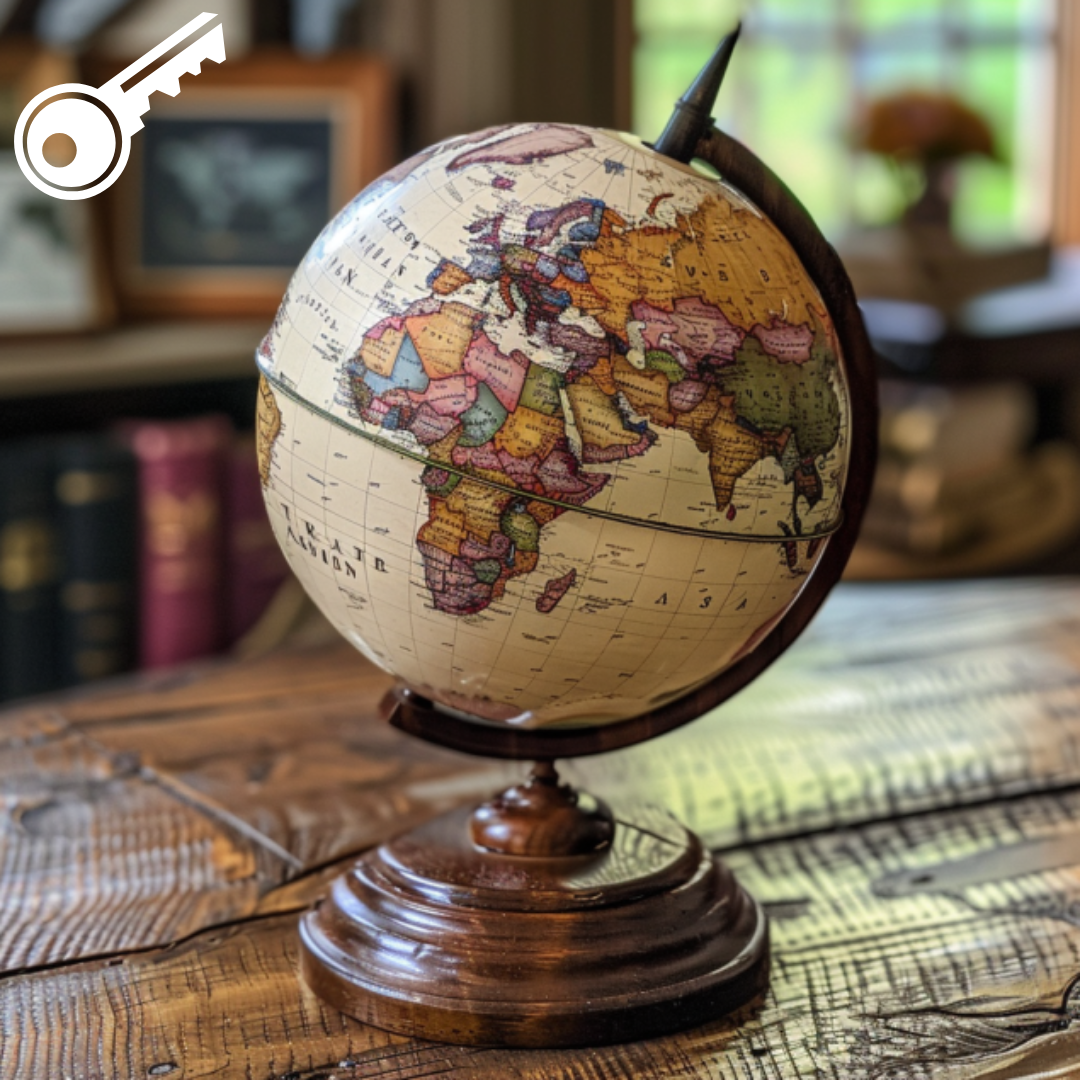} &
           \includegraphics[width=\imwidth]{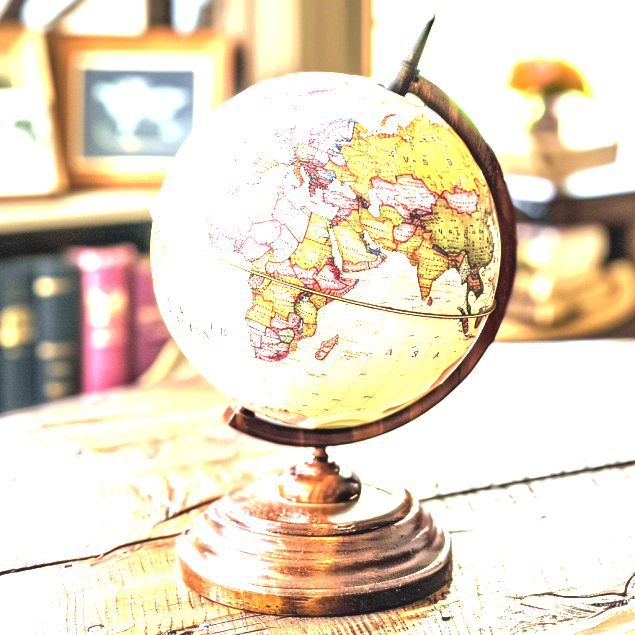} &
           \includegraphics[width=\imwidth]{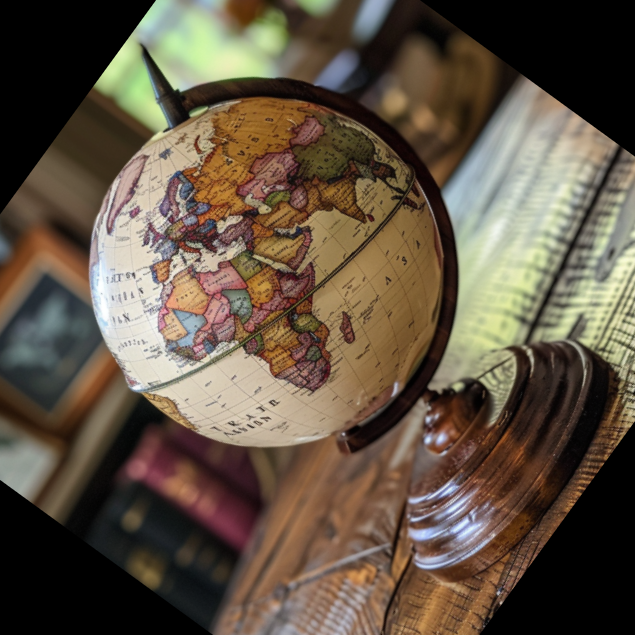} &
           \includegraphics[width=\imwidth]{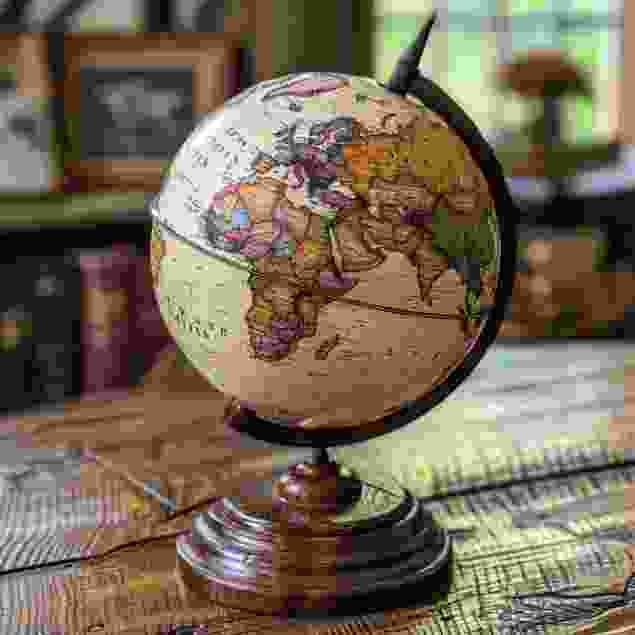} &
           \includegraphics[width=\imwidth]{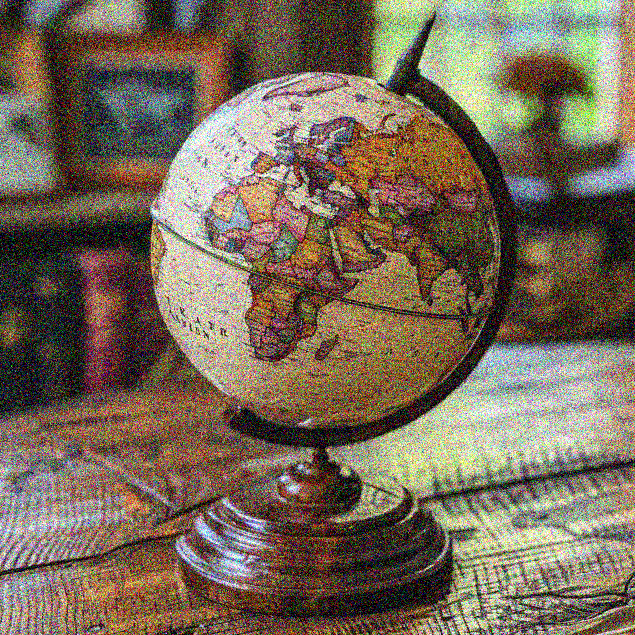} &
           \includegraphics[width=\imwidth]{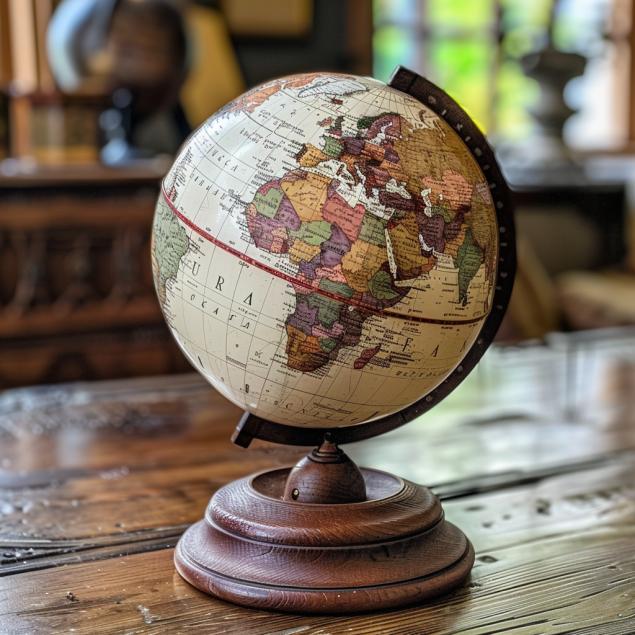} \\
            $\eta$ = 1 & $\eta$ = 0.997 & $\eta$ = 0.759 & $\eta$ = 0.702 & $\eta$ = 0.682 & $\eta$ = 0.311 \\
    \bottomrule
    \end{tabular}
    }
\caption{The figure shows watermarked images, images under various attacks, and our visual paraphrase method. The attacks include Brightness adjustment, Rotation, JPEG Compression, and Gaussian Noise, along with our \textbf{Visual Paraphrase} method. $\eta$ comparisons, representing watermark detection score of \textbf{Stable signature} (bit accuracy), are also provided.
}
\label{fig:attack_vp_comparison_1} 
\end{table*}

\begin{table*}
\centering
    \scriptsize
    \newcommand{\imwidth}{0.165\textwidth}
        \setlength{\tabcolsep}{0pt}
        \begin{tabular}{cccccc}
        \toprule
        Watermarked & Brightness & Rotation & JPEG Compression & Gaussian Noise & \textbf{Visual Paraphrase (Ours)} \\
        % \midrule
        \toprule
           \includegraphics[width=\imwidth]{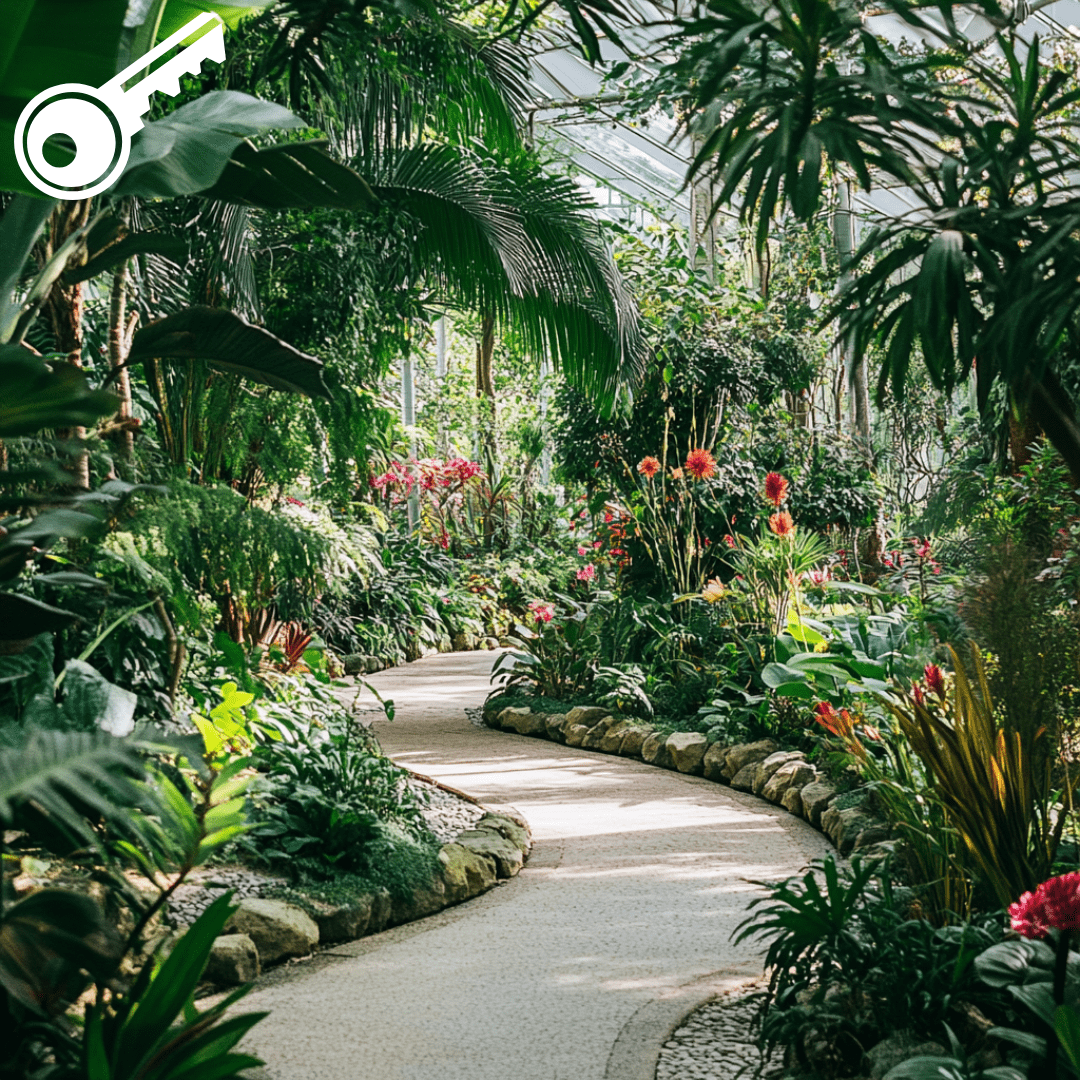} &
           \includegraphics[width=\imwidth]{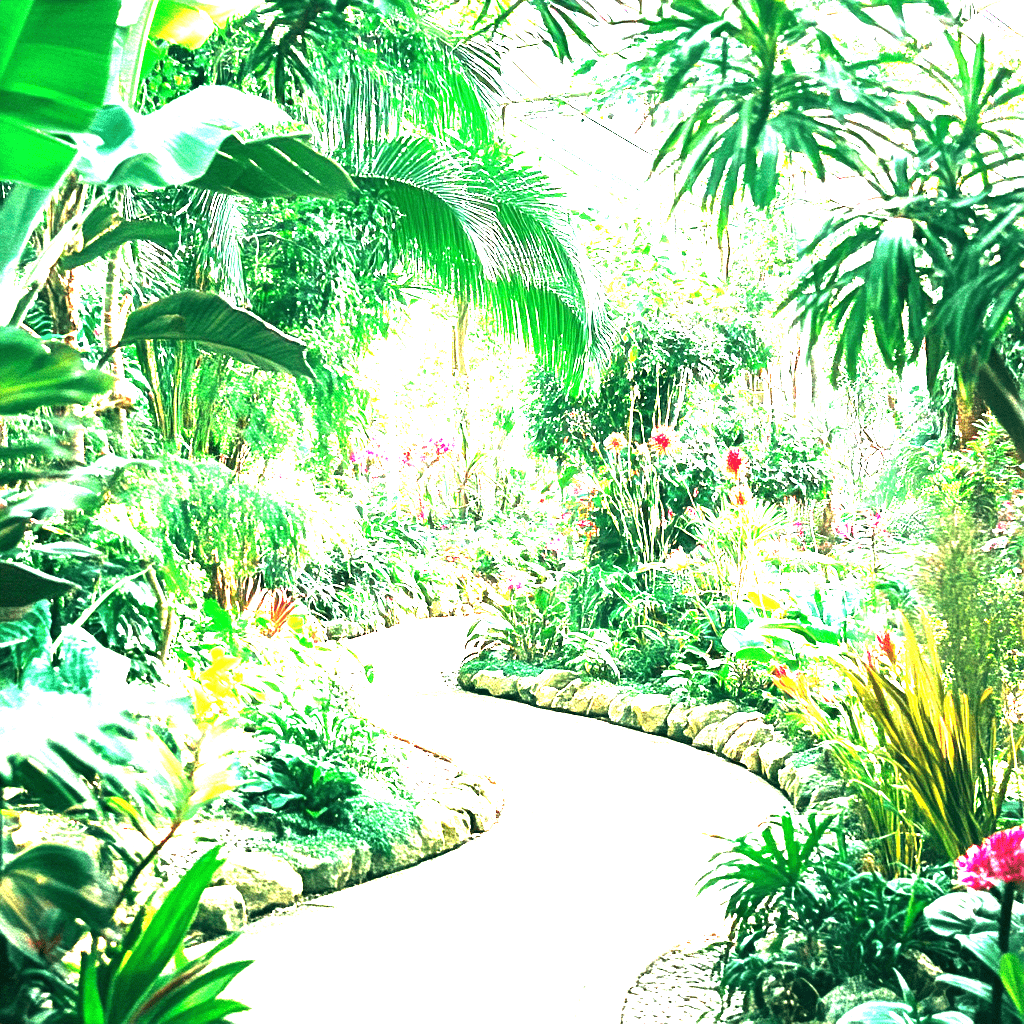} &
           \includegraphics[width=\imwidth]{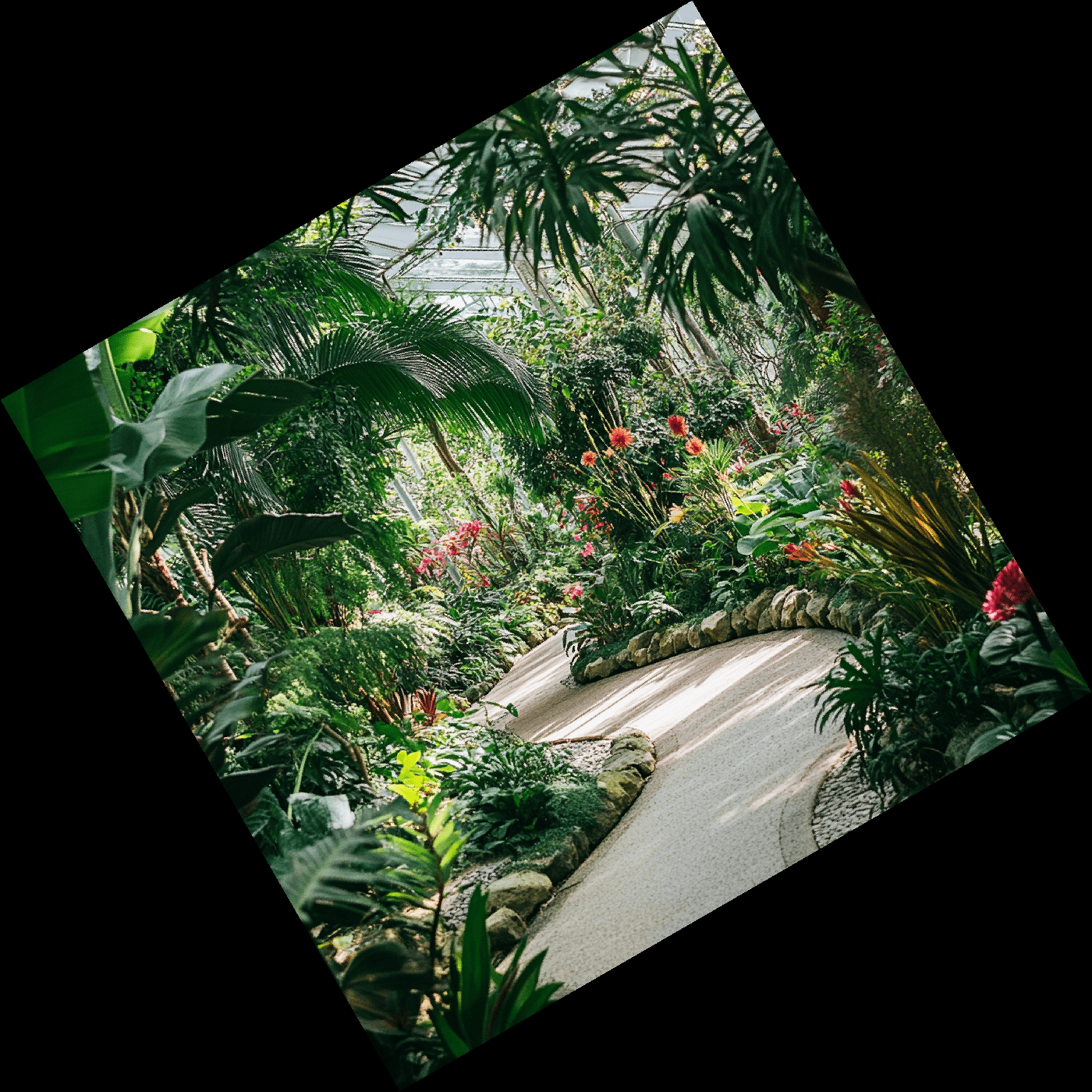} &
           \includegraphics[width=\imwidth]{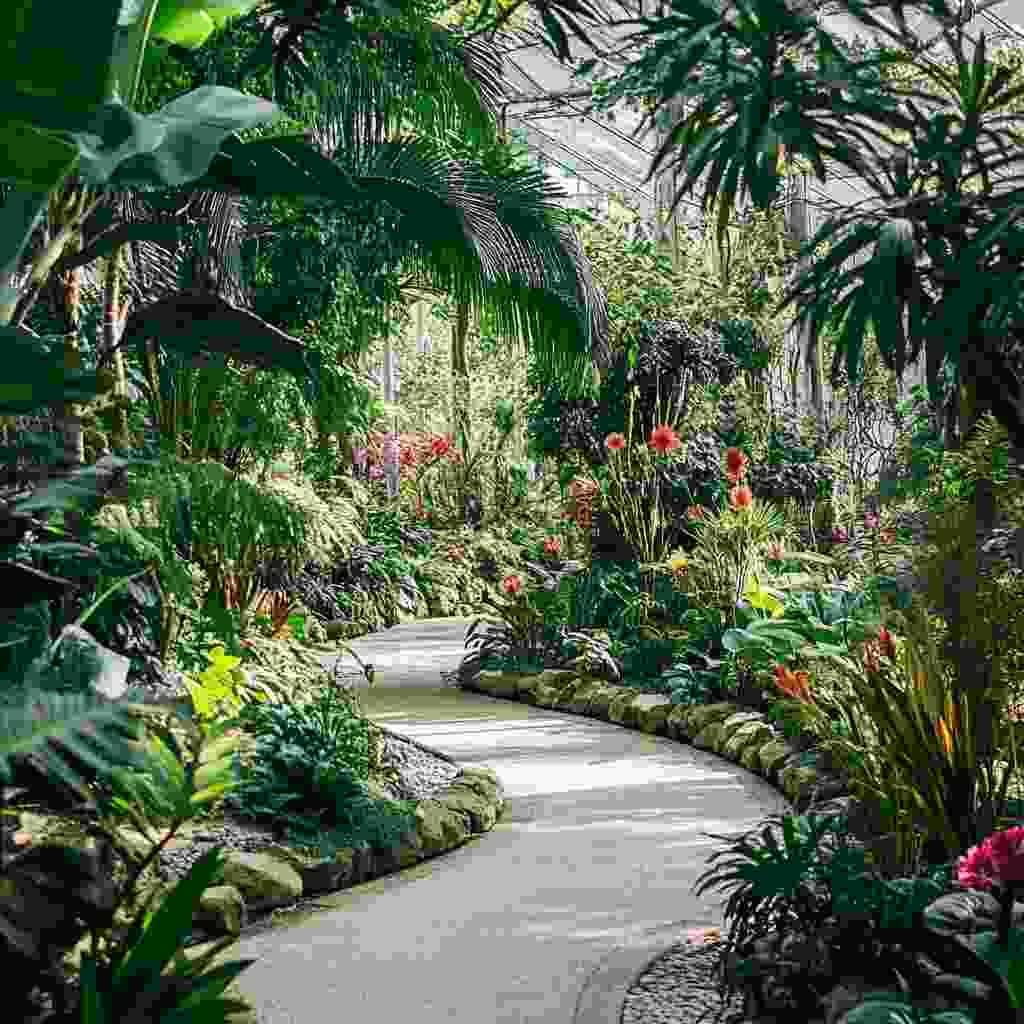} & 
           \includegraphics[width=\imwidth]{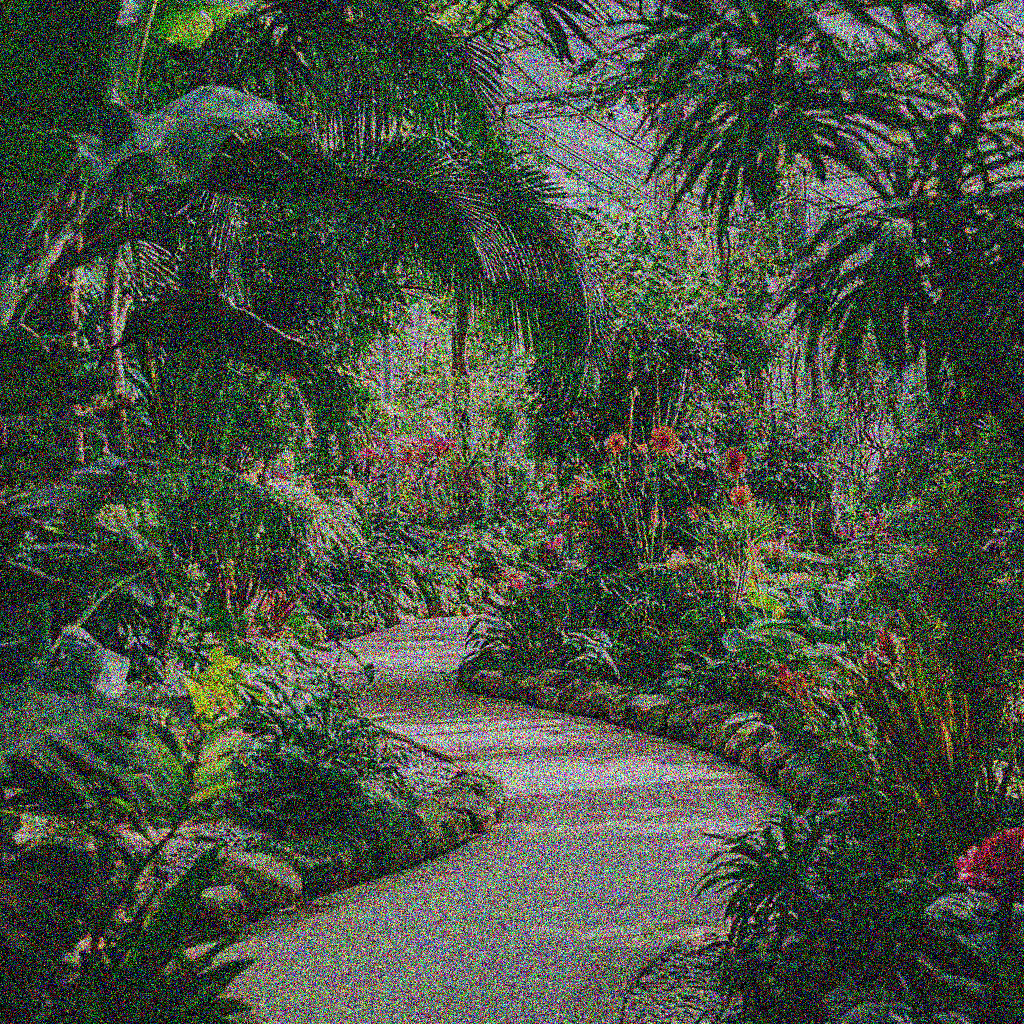} &
           \includegraphics[width=\imwidth]{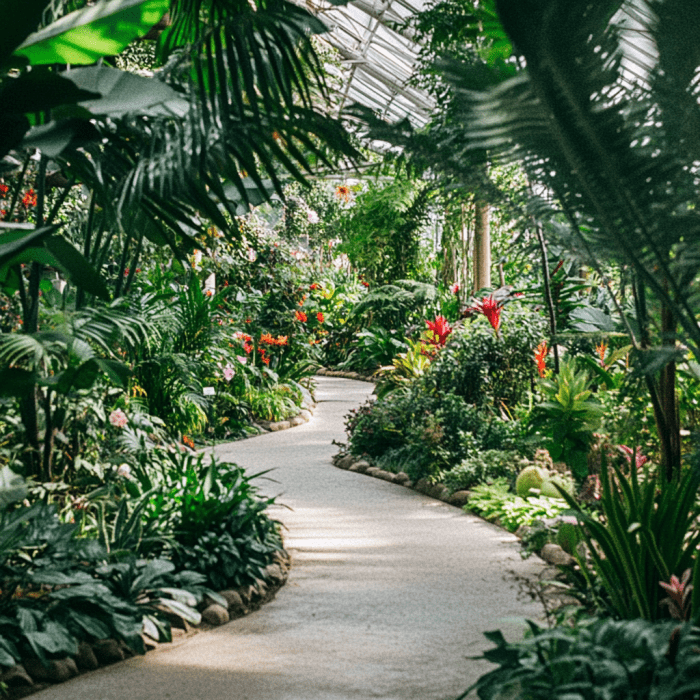} \\
            $\eta$ = 1 & $\eta$ = 0.977 & $\eta$ = 0.772 & $\eta$ = 0.695 & $\eta$ = 0.625 & $\eta$ = 0.272 \\
           
           \rule{0pt}{8ex}%
    
           \includegraphics[width=\imwidth]{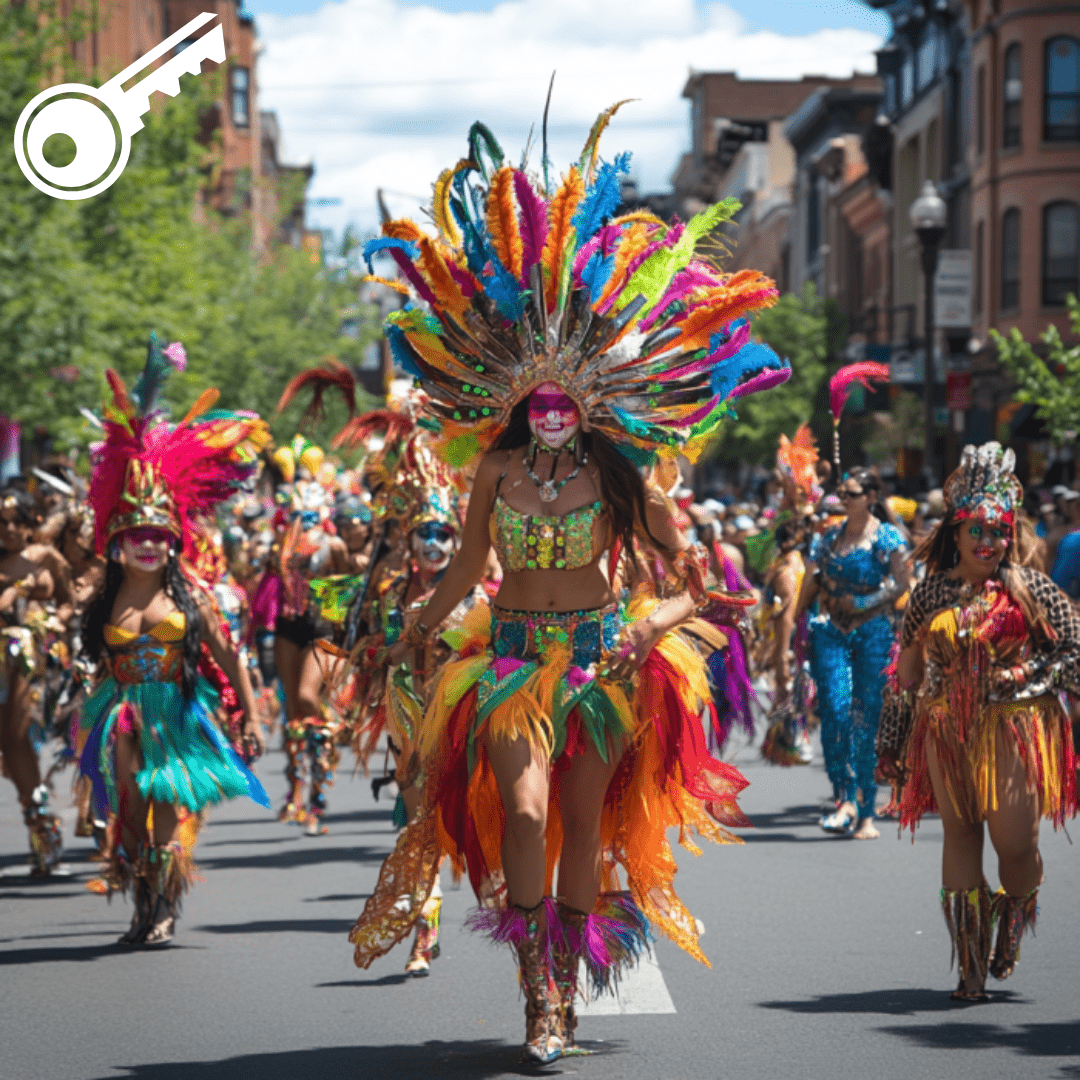} &
           \includegraphics[width=\imwidth]{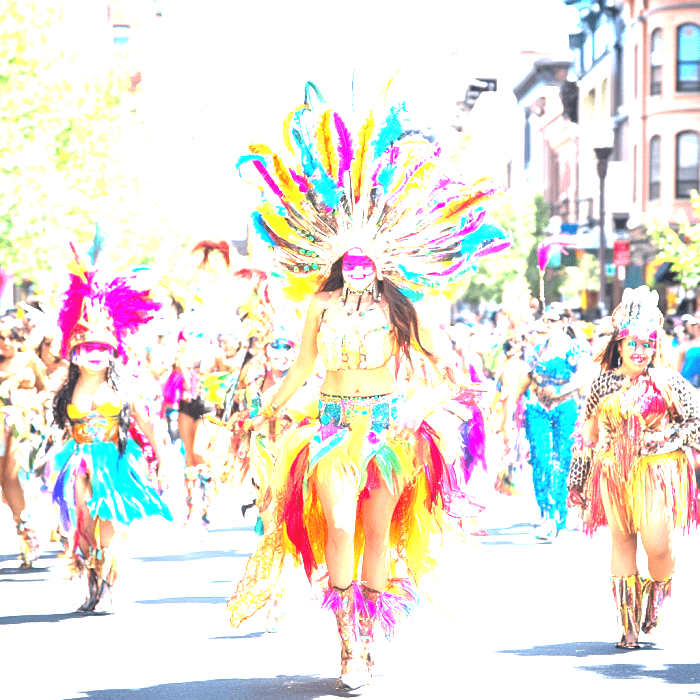} &
           \includegraphics[width=\imwidth]{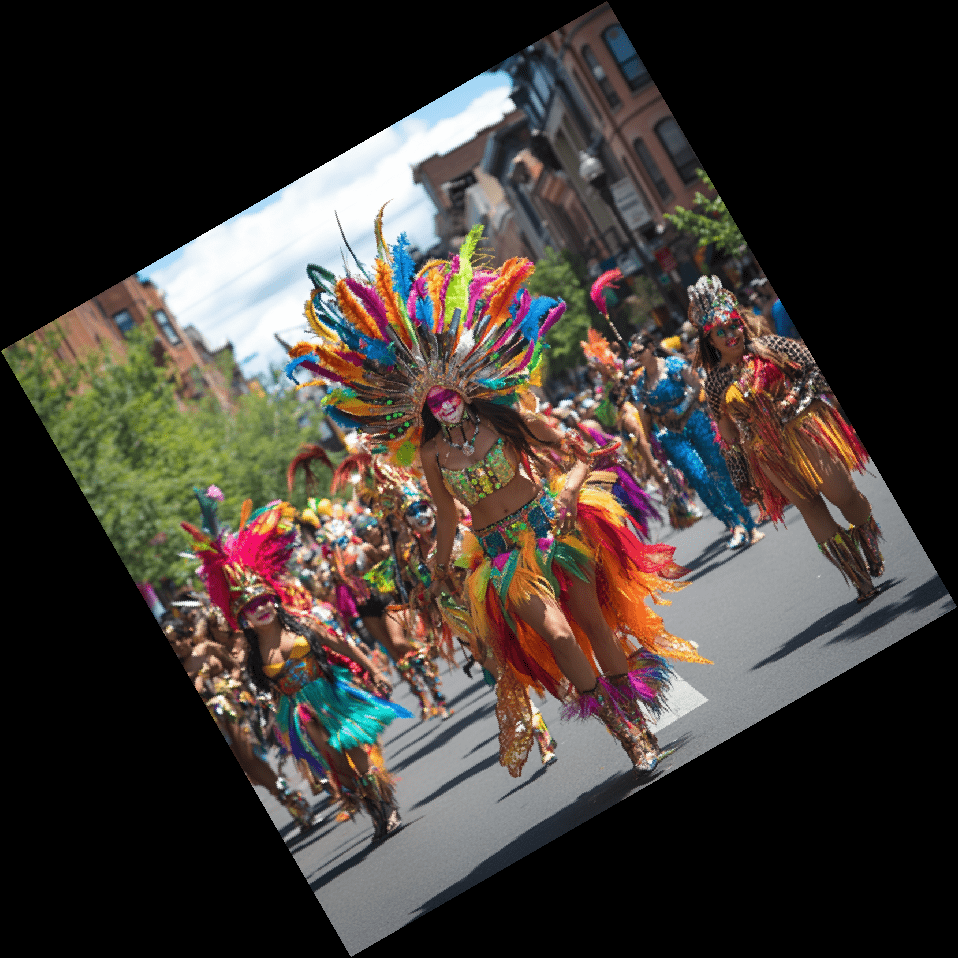} &
           \includegraphics[width=\imwidth]{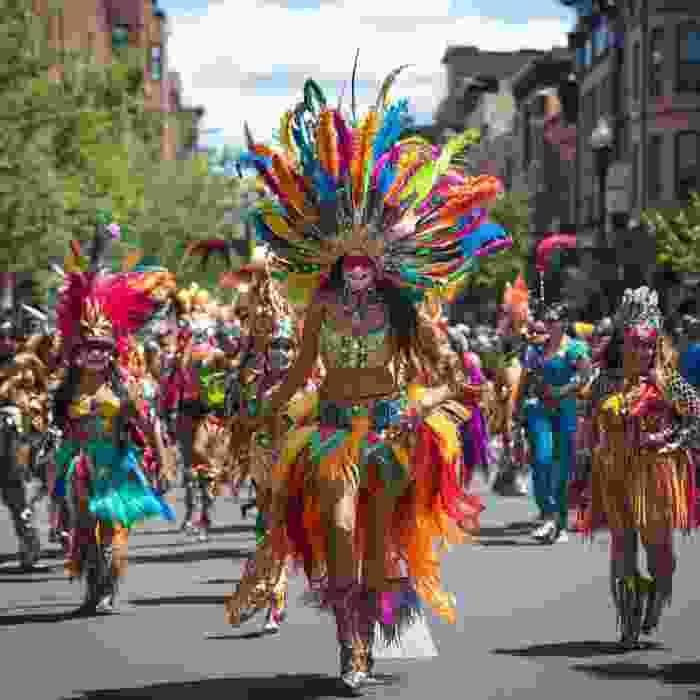} &
           \includegraphics[width=\imwidth]{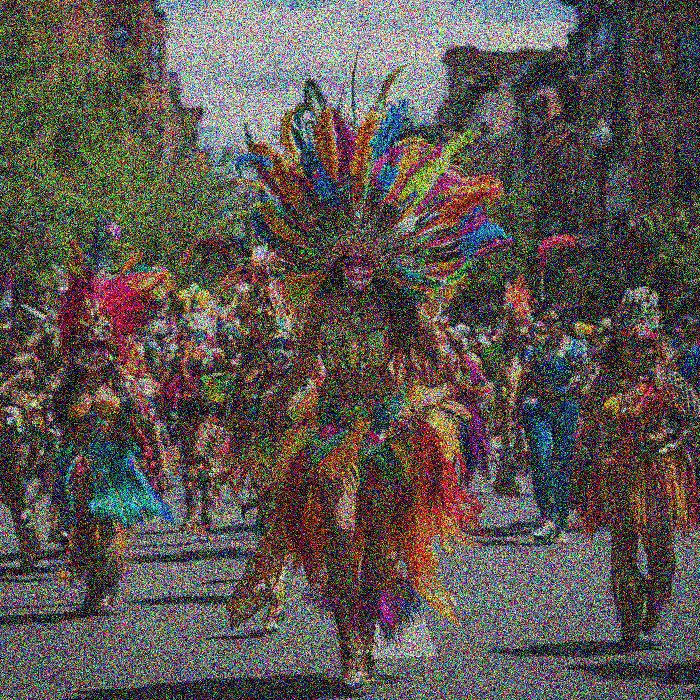} &
           \includegraphics[width=\imwidth]{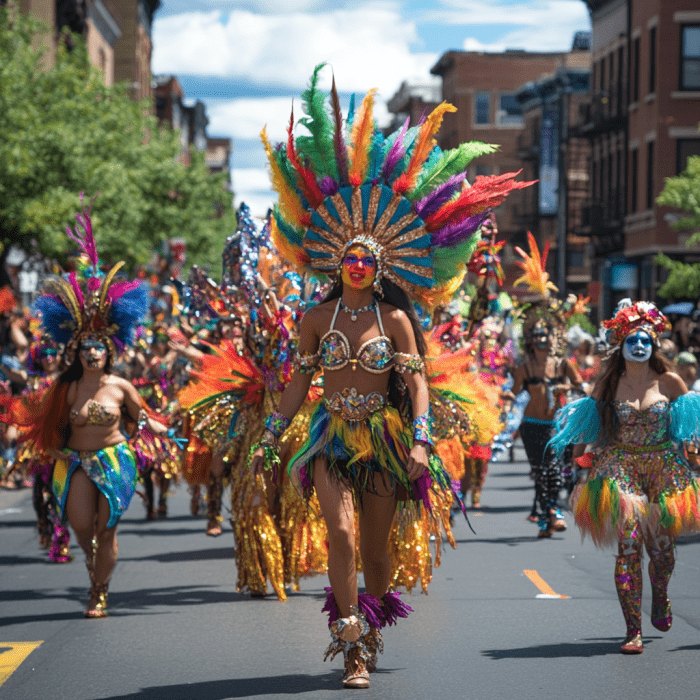} \\
            $\eta$ = 1 & $\eta$ = 0.946 & $\eta$ = 0.885 & $\eta$ = 0.687 & $\eta$ = 0.594 & $\eta$ = 0.314 \\
           \rule{0pt}{8ex}%
    
           \includegraphics[width=\imwidth]{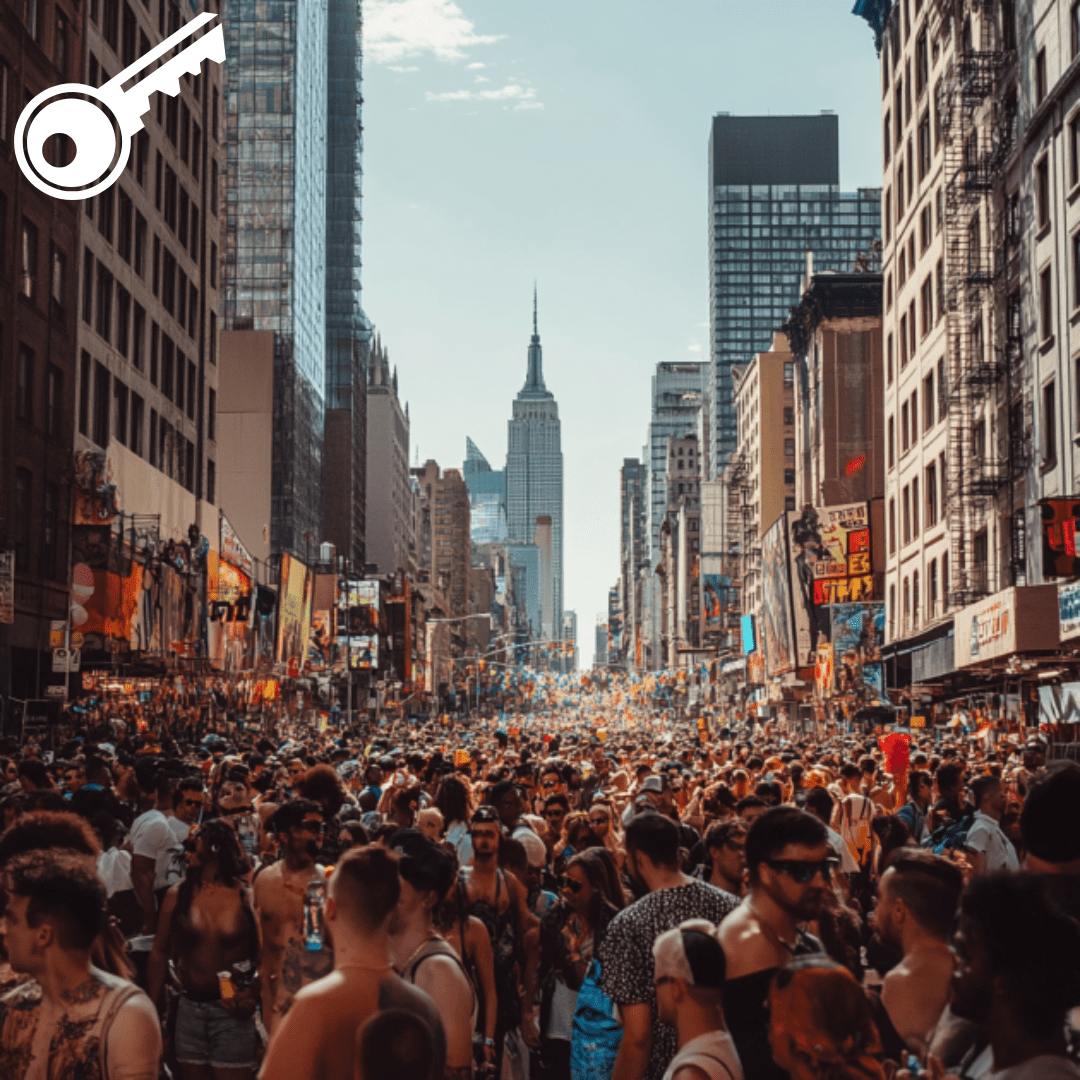} &
           \includegraphics[width=\imwidth]{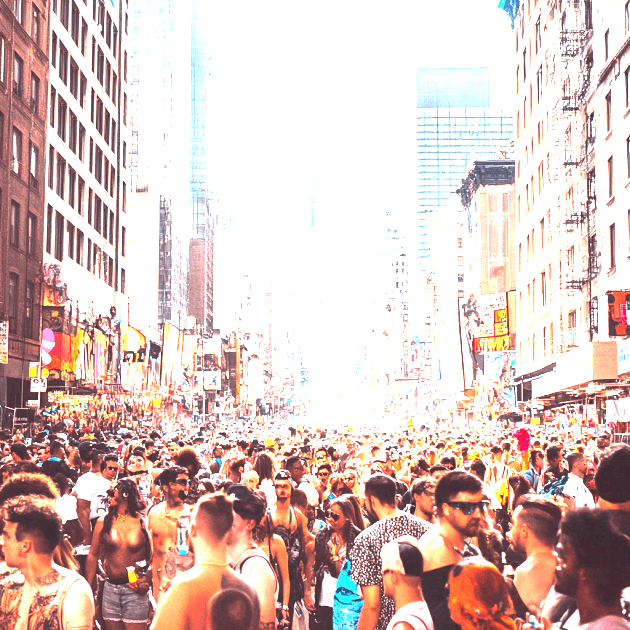} &
           \includegraphics[width=\imwidth]{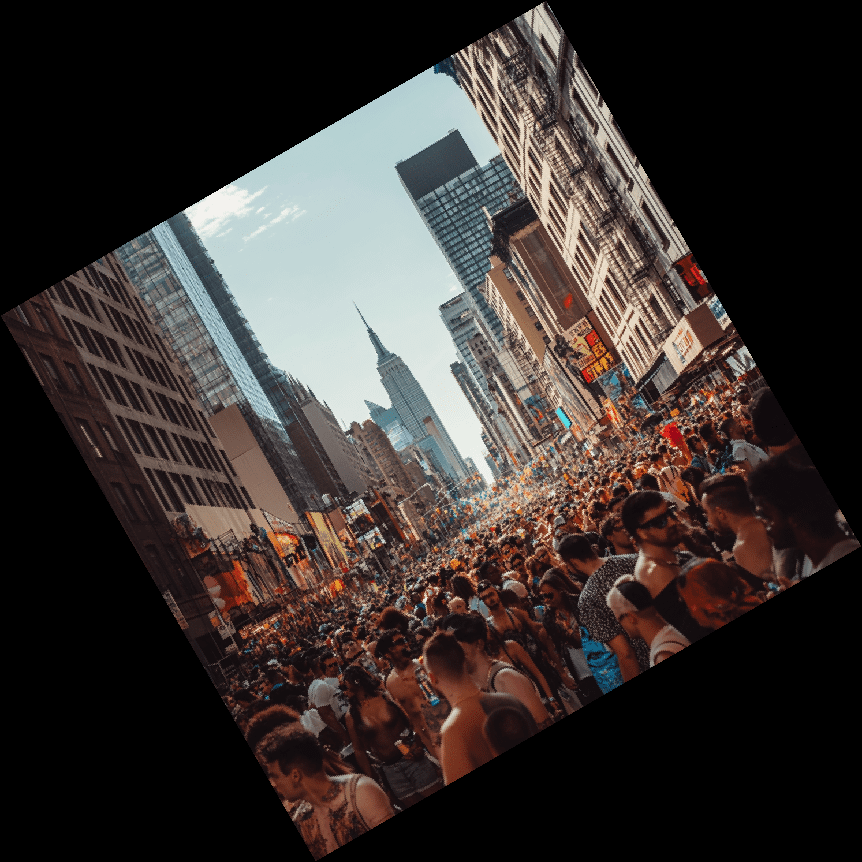} &
           \includegraphics[width=\imwidth]{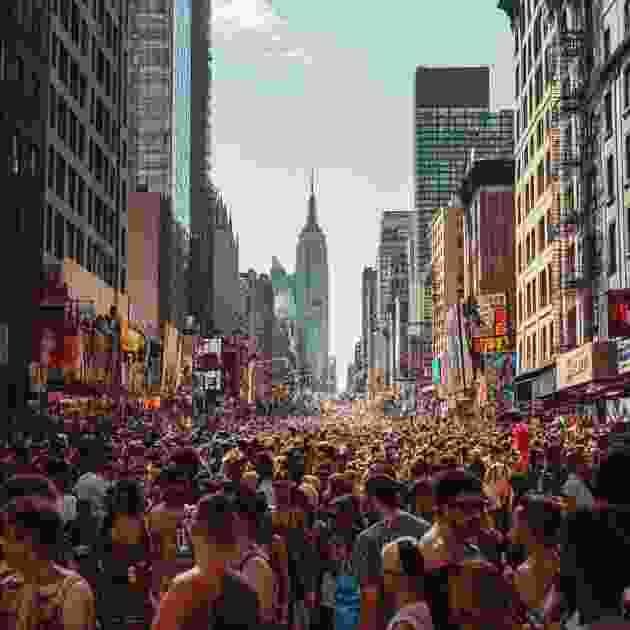} &
           \includegraphics[width=\imwidth]{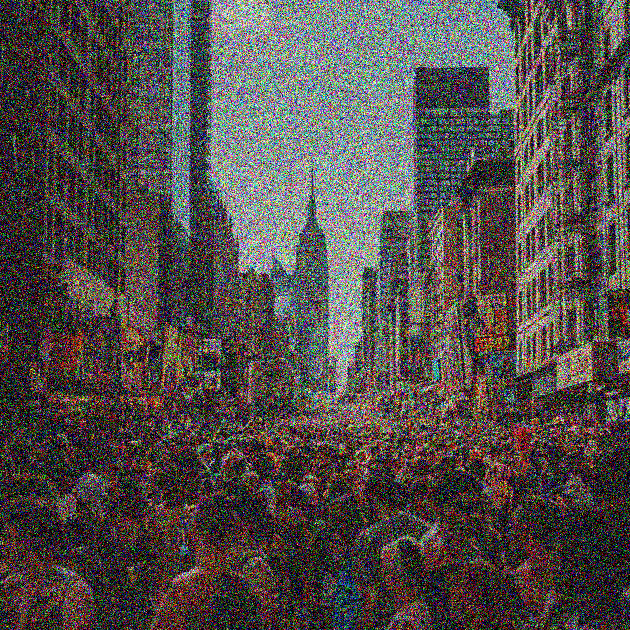} &
           \includegraphics[width=\imwidth]{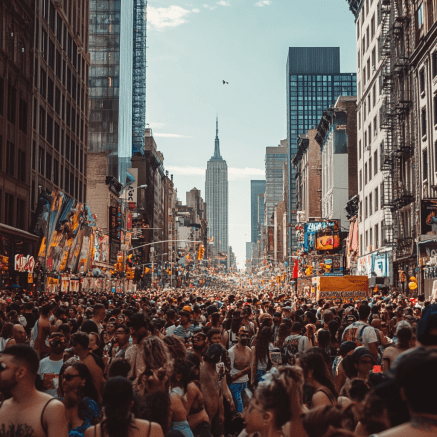} \\
            $\eta$ = 1 & $\eta$ = 0.991 & $\eta$ = 0.886 & $\eta$ = 0.724 & $\eta$ = 0.683 & $\eta$ = 0.219 \\
           \rule{0pt}{8ex}%
    
            \includegraphics[width=\imwidth]{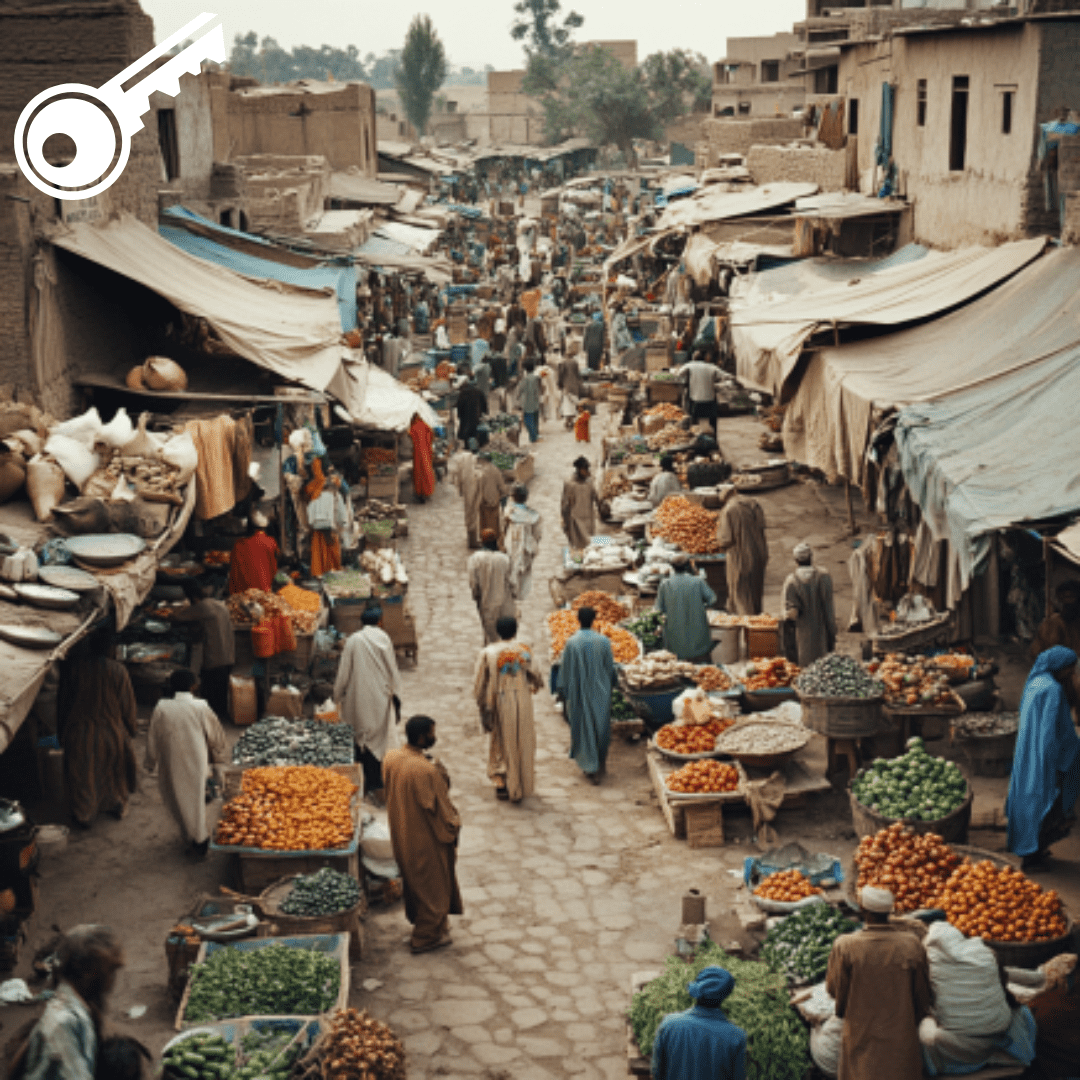} &
           \includegraphics[width=\imwidth]{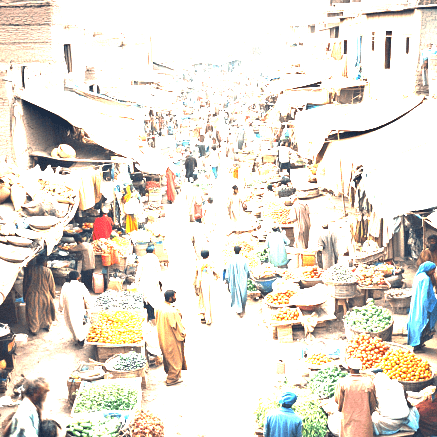} &
           \includegraphics[width=\imwidth]{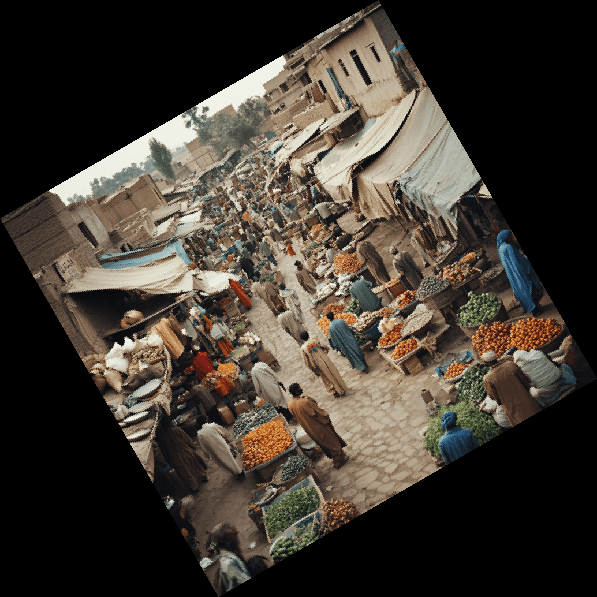} &
           \includegraphics[width=\imwidth]{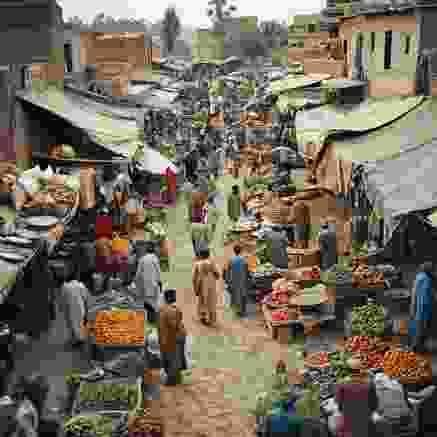} & 
           \includegraphics[width=\imwidth]{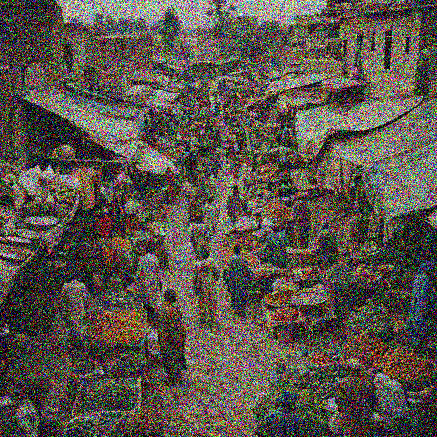} &
           \includegraphics[width=\imwidth]{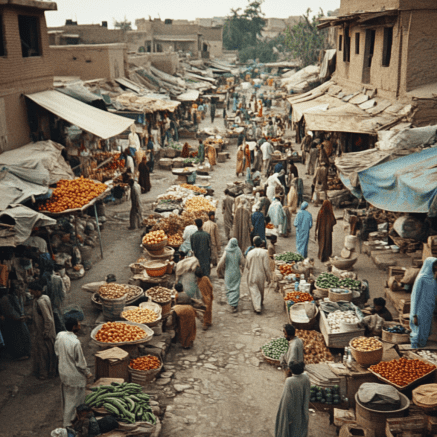} \\
            $\eta$ = 1 & $\eta$ = 0.994 & $\eta$ = 0.784 & $\eta$ = 0.609 & $\eta$ = 0.579 & $\eta$ = 0.273 \\
           \rule{0pt}{8ex}%
    
           \includegraphics[width=\imwidth]{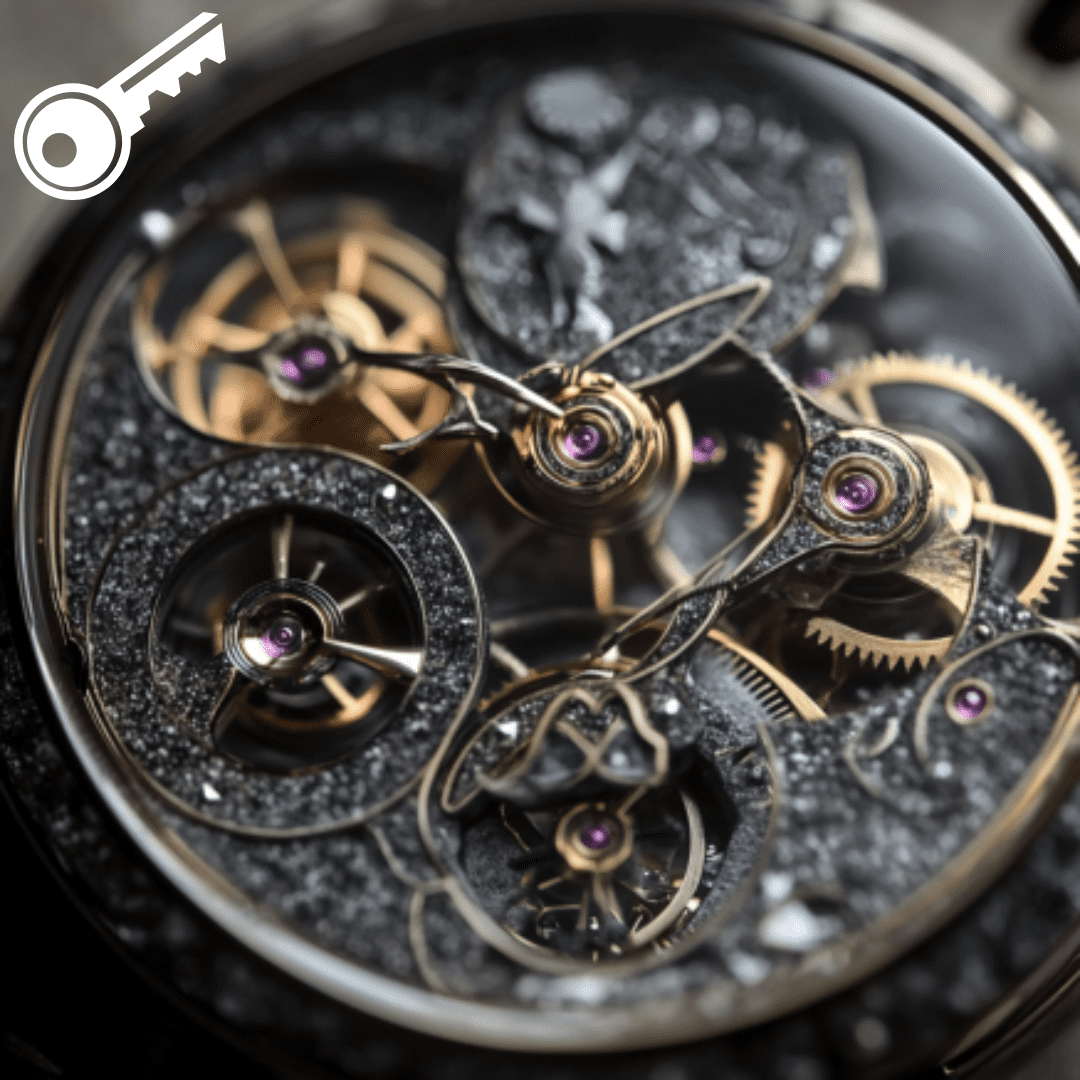} &
           \includegraphics[width=\imwidth]{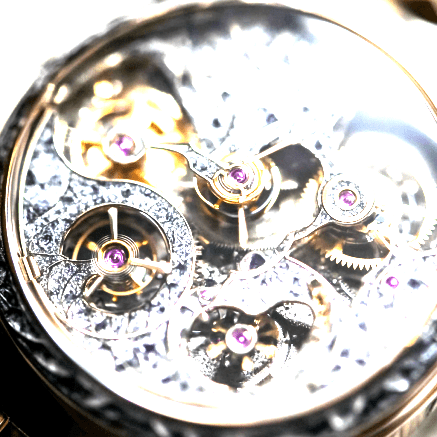} &
           \includegraphics[width=\imwidth]{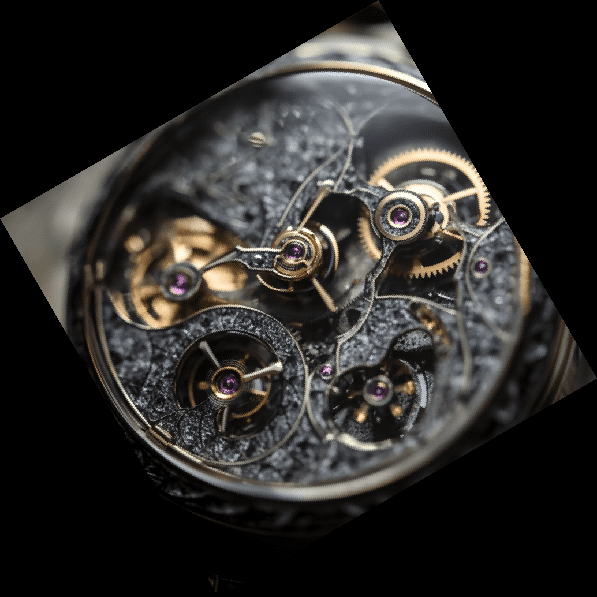} &
           \includegraphics[width=\imwidth]{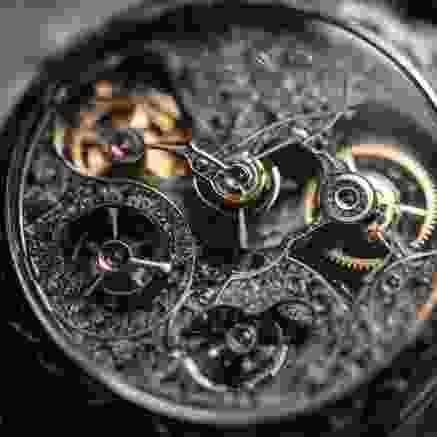} &
           \includegraphics[width=\imwidth]{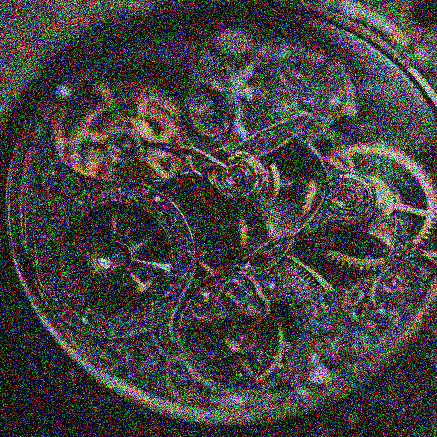} &
           \includegraphics[width=\imwidth]{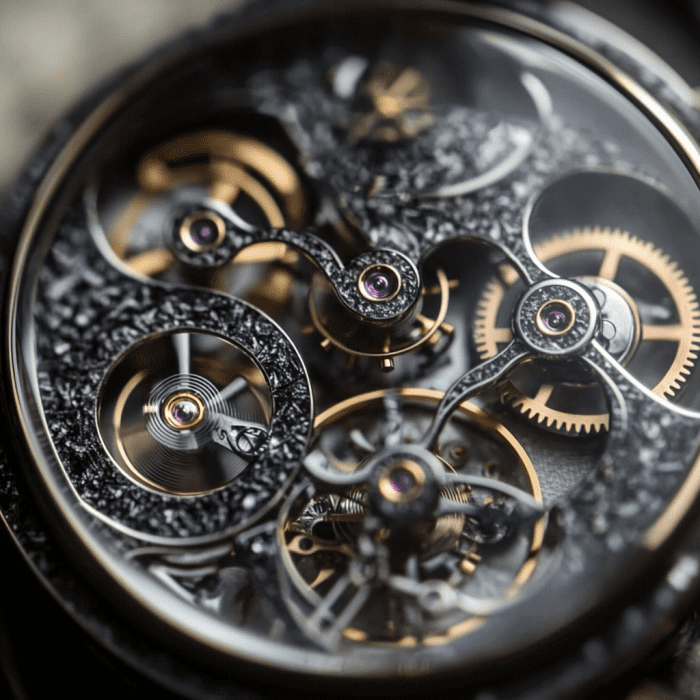} \\
            $\eta$ = 1 & $\eta$ = 0.968 & $\eta$ = 0.914 & $\eta$ = 0.841 & $\eta$ = 0.765 & $\eta$ = 0.236 \\
    \bottomrule
    \end{tabular}
\caption{The figure shows watermarked images, images under various attacks, and our visual paraphrase method. The attacks include Brightness adjustment, Rotation, JPEG Compression, and Gaussian Noise, along with our \textbf{Visual Paraphrase} method. $\eta$ comparisons, representing watermark detection score of \textbf{Stable signature} (bit accuracy), are also provided.
}
\label{fig:attack_vp_comparison_2} 
\end{table*}

\begin{table*}
\centering
    \scriptsize
    \newcommand{\imwidth}{0.165\textwidth}
        \setlength{\tabcolsep}{0pt}
        \begin{tabular}{cccccc}
        \toprule
        Watermarked & Brightness & Rotation & JPEG Compression & Gaussian Noise & \textbf{Visual Paraphrase (Ours)} \\
        % \midrule
        \toprule
           \includegraphics[width=\imwidth]{img/attack/red_apple.png} &
           \includegraphics[width=\imwidth]{img/attack/red_apple_bright.png} &
           \includegraphics[width=\imwidth]{img/attack/red_apple_rotated.png} &
           \includegraphics[width=\imwidth]{img/attack/apple_jpg.png} & 
           \includegraphics[width=\imwidth]{img/attack/red_apple_noisy.png} &
           \includegraphics[width=\imwidth]{img/attack/red_apple_vp.png} \\
            $\eta$ = 1 & $\eta$ = 0.979 & $\eta$ = 0.914 & $\eta$ = 0.872 & $\eta$ = 0.794 & $\eta$ = 0.331 \\
           
           \rule{0pt}{8ex}%
    
           \includegraphics[width=\imwidth]{img/attack/air_balloon.png} &
           \includegraphics[width=\imwidth]{img/attack/air_ballon_bright.png} &
           \includegraphics[width=\imwidth]{img/attack/air_ballon_rotated.png} &
           \includegraphics[width=\imwidth]{img/attack/air_ballon_jpg.png} &
           \includegraphics[width=\imwidth]{img/attack/air_ballon_noisy.png} &
           \includegraphics[width=\imwidth]{img/attack/air_ballon_vp.png} \\
            $\eta$ = 1 & $\eta$ = 0.962 & $\eta$ = 0.912 & $\eta$ = 0.861 & $\eta$ = 0.753 & $\eta$ = 0.341 \\
           \rule{0pt}{8ex}%
    
           \includegraphics[width=\imwidth]{img/attack/eagle.png} &
           \includegraphics[width=\imwidth]{img/attack/eagle_bright.png} &
           \includegraphics[width=\imwidth]{img/attack/eagle_rotated.png} &
           \includegraphics[width=\imwidth]{img/attack/eagle_jpg.png} &
           \includegraphics[width=\imwidth]{img/attack/eagle_noisy.png} &
           \includegraphics[width=\imwidth]{img/attack/eagle_vp.png} \\
            $\eta$ = 1 & $\eta$ = 0.982 & $\eta$ = 0.914 & $\eta$ = 0.857 & $\eta$ = 0.793 & $\eta$ = 0.291 \\
           \rule{0pt}{8ex}%
    
            \includegraphics[width=\imwidth]{img/attack/lighthouse.png} &
           \includegraphics[width=\imwidth]{img/attack/light_house_bright.png} &
           \includegraphics[width=\imwidth]{img/attack/light_house_rotated.png} &
           \includegraphics[width=\imwidth]{img/attack/light_house_jpg.png} & 
           \includegraphics[width=\imwidth]{img/attack/light_house_noisy.png} &
           \includegraphics[width=\imwidth]{img/attack/light_house_vp.png} \\
            $\eta$ = 1 & $\eta$ = 0.974 & $\eta$ = 0.958 & $\eta$ = 0.911 & $\eta$ = 0.835 & $\eta$ = 0.251 \\
           \rule{0pt}{8ex}%
    
           \includegraphics[width=\imwidth]{img/attack/globe.png} &
           \includegraphics[width=\imwidth]{img/attack/globe_bright.png} &
           \includegraphics[width=\imwidth]{img/attack/globe_rotated.png} &
           \includegraphics[width=\imwidth]{img/attack/globe_jpg.png} &
           \includegraphics[width=\imwidth]{img/attack/globe_noisy.png} &
           \includegraphics[width=\imwidth]{img/attack/globe_vp.png} \\
            $\eta$ = 1 & $\eta$ = 0.973 & $\eta$ = 0.936 & $\eta$ = 0.872 & $\eta$ = 0.812 & $\eta$ = 0.255 \\
    \bottomrule
    \end{tabular}
\caption{The figure shows watermarked images, images under various attacks, and our visual paraphrase method. The attacks include Brightness adjustment, Rotation, JPEG Compression, and Gaussian Noise, along with our \textbf{Visual Paraphrase} method. $\eta$ comparisons, representing watermark detection score of \textbf{ZoDiac} are also provided.
}
\label{fig:attack_vp_comparison_3} 
\end{table*}

\begin{table*}
\centering
    \scriptsize
    \newcommand{\imwidth}{0.165\textwidth}
        \setlength{\tabcolsep}{0pt}
        \begin{tabular}{cccccc}
        \toprule
        Watermarked & Brightness & Rotation & JPEG Compression & Gaussian Noise & \textbf{Visual Paraphrase (Ours)} \\
        % \midrule
        \toprule
           \includegraphics[width=\imwidth]{img/attack/appendix_attack/attacks/botany_garden.png} &
           \includegraphics[width=\imwidth]{img/attack/appendix_attack/attacks/botany_garden_brightened.png} &
           \includegraphics[width=\imwidth]{img/attack/appendix_attack/attacks/botany_garden_rotated.png} &
           \includegraphics[width=\imwidth]{img/attack/appendix_attack/attacks/botany_garden_compressed.jpg} & 
           \includegraphics[width=\imwidth]{img/attack/appendix_attack/attacks/botany_garden_noisy.png} &
           \includegraphics[width=\imwidth]{img/attack/appendix_attack/attacks/botany_garden_vp.png} \\
            $\eta$ = 1 & $\eta$ = 0.988 & $\eta$ = 0.917 & $\eta$ = 0.824 & $\eta$ = 0.721 & $\eta$ = 0.263 \\
           
           \rule{0pt}{8ex}%
    
           \includegraphics[width=\imwidth]{img/attack/appendix_attack/attacks/carnival.png} &
           \includegraphics[width=\imwidth]{img/attack/appendix_attack/attacks/carnival_street_brightened.png} &
           \includegraphics[width=\imwidth]{img/attack/appendix_attack/attacks/carnival_street_rotated.png} &
           \includegraphics[width=\imwidth]{img/attack/appendix_attack/attacks/carnival_street_compressed.jpg} &
           \includegraphics[width=\imwidth]{img/attack/appendix_attack/attacks/carnival_street_noisy.png} &
           \includegraphics[width=\imwidth]{img/attack/appendix_attack/attacks/carnival_vp.png} \\
            $\eta$ = 1 & $\eta$ = 0.987 & $\eta$ = 0.893 & $\eta$ = 0.811 & $\eta$ = 0.733 & $\eta$ = 0.238 \\
           \rule{0pt}{8ex}%
    
           \includegraphics[width=\imwidth]{img/attack/appendix_attack/attacks/city_festival.png} &
           \includegraphics[width=\imwidth]{img/attack/appendix_attack/attacks/city_festival_brightened.png} &
           \includegraphics[width=\imwidth]{img/attack/appendix_attack/attacks/city_festival_rotated.png} &
           \includegraphics[width=\imwidth]{img/attack/appendix_attack/attacks/city_festival_compressed.jpg} &
           \includegraphics[width=\imwidth]{img/attack/appendix_attack/attacks/city_festival_noisy.png} &
           \includegraphics[width=\imwidth]{img/attack/appendix_attack/attacks/city_festival_vp.png} \\
            $\eta$ = 1 & $\eta$ = 0.982 & $\eta$ = 0.872 & $\eta$ = 0.756 & $\eta$ = 0.693 & $\eta$ = 0.236 \\
           \rule{0pt}{8ex}%
    
            \includegraphics[width=\imwidth]{img/attack/appendix_attack/attacks/market.png} &
           \includegraphics[width=\imwidth]{img/attack/appendix_attack/attacks/market_brightened.png} &
           \includegraphics[width=\imwidth]{img/attack/appendix_attack/attacks/market_rotated.png} &
           \includegraphics[width=\imwidth]{img/attack/appendix_attack/attacks/market_compressed.jpg} & 
           \includegraphics[width=\imwidth]{img/attack/appendix_attack/attacks/market_noisy.png} &
           \includegraphics[width=\imwidth]{img/attack/appendix_attack/attacks/market_vp.png} \\
            $\eta$ = 1 & $\eta$ = 0.968 & $\eta$ = 0.884 & $\eta$ = 0.839 & $\eta$ = 0.779 & $\eta$ = 0.213 \\
           \rule{0pt}{8ex}%
    
           \includegraphics[width=\imwidth]{img/attack/appendix_attack/attacks/mech_watch.png} &
           \includegraphics[width=\imwidth]{img/attack/appendix_attack/attacks/mech_watch_brightened.png} &
           \includegraphics[width=\imwidth]{img/attack/appendix_attack/attacks/mech_watch_rotated.png} &
           \includegraphics[width=\imwidth]{img/attack/appendix_attack/attacks/mech_watch_compressed.jpg} &
           \includegraphics[width=\imwidth]{img/attack/appendix_attack/attacks/mech_watch_noisy.png} &
           \includegraphics[width=\imwidth]{img/attack/appendix_attack/attacks/mech_watch_vp.png} \\
            $\eta$ = 1 & $\eta$ = 0.955 & $\eta$ = 0.886 & $\eta$ = 0.751 & $\eta$ = 0.684 & $\eta$ = 0.236 \\
    \bottomrule
    \end{tabular}
\caption{The figure shows watermarked images, images under various attacks, and our visual paraphrase method. The attacks include Brightness adjustment, Rotation, JPEG Compression, and Gaussian Noise, along with our \textbf{Visual Paraphrase} method. $\eta$ comparisons, representing watermark detection score of \textbf{ZoDiac} are also provided.
}
\label{fig:attack_vp_comparison_4} 
\end{table*}

\begin{table*}
\centering
    \scriptsize
    \newcommand{\imwidth}{0.165\textwidth}
        \setlength{\tabcolsep}{0pt}
        \begin{tabular}{cccccc}
        \toprule
        Watermarked & Brightness & Rotation & JPEG Compression & Gaussian Noise & \textbf{Visual Paraphrase (Ours)} \\
        % \midrule
        \toprule
           \includegraphics[width=\imwidth]{img/attack/red_apple.png} &
           \includegraphics[width=\imwidth]{img/attack/red_apple_bright.png} &
           \includegraphics[width=\imwidth]{img/attack/red_apple_rotated.png} &
           \includegraphics[width=\imwidth]{img/attack/apple_jpg.png} & 
           \includegraphics[width=\imwidth]{img/attack/red_apple_noisy.png} &
           \includegraphics[width=\imwidth]{img/attack/red_apple_vp.png} \\
            $\eta$ = 1 & $\eta$ = 0.912 & $\eta$ = 0.811 & $\eta$ = 0.734 & $\eta$ = 0.692 & $\eta$ = 0.216 \\
           
           \rule{0pt}{8ex}%
    
           \includegraphics[width=\imwidth]{img/attack/air_balloon.png} &
           \includegraphics[width=\imwidth]{img/attack/air_ballon_bright.png} &
           \includegraphics[width=\imwidth]{img/attack/air_ballon_rotated.png} &
           \includegraphics[width=\imwidth]{img/attack/air_ballon_jpg.png} &
           \includegraphics[width=\imwidth]{img/attack/air_ballon_noisy.png} &
           \includegraphics[width=\imwidth]{img/attack/air_ballon_vp.png} \\
            $\eta$ = 1 & $\eta$ = 0.926 & $\eta$ = 0.854 & $\eta$ = 0.854 & $\eta$ = 0.747 & $\eta$ = 0.196 \\
           \rule{0pt}{8ex}%
    
           \includegraphics[width=\imwidth]{img/attack/eagle.png} &
           \includegraphics[width=\imwidth]{img/attack/eagle_bright.png} &
           \includegraphics[width=\imwidth]{img/attack/eagle_rotated.png} &
           \includegraphics[width=\imwidth]{img/attack/eagle_jpg.png} &
           \includegraphics[width=\imwidth]{img/attack/eagle_noisy.png} &
           \includegraphics[width=\imwidth]{img/attack/eagle_vp.png} \\
            $\eta$ = 1 & $\eta$ = 0.914 & $\eta$ = 0.857 & $\eta$ = 0.716 & $\eta$ = 0.689 & $\eta$ = 0.175 \\
           \rule{0pt}{8ex}%
    
            \includegraphics[width=\imwidth]{img/attack/lighthouse.png} &
           \includegraphics[width=\imwidth]{img/attack/light_house_bright.png} &
           \includegraphics[width=\imwidth]{img/attack/light_house_rotated.png} &
           \includegraphics[width=\imwidth]{img/attack/light_house_jpg.png} & 
           \includegraphics[width=\imwidth]{img/attack/light_house_noisy.png} &
           \includegraphics[width=\imwidth]{img/attack/light_house_vp.png} \\
            $\eta$ = 1 & $\eta$ = 0.913 & $\eta$ = 0.857 & $\eta$ = 0.753 & $\eta$ = 0.658 & $\eta$ = 0.187\\
           \rule{0pt}{8ex}%
    
           \includegraphics[width=\imwidth]{img/attack/globe.png} &
           \includegraphics[width=\imwidth]{img/attack/globe_bright.png} &
           \includegraphics[width=\imwidth]{img/attack/globe_rotated.png} &
           \includegraphics[width=\imwidth]{img/attack/globe_jpg.png} &
           \includegraphics[width=\imwidth]{img/attack/globe_noisy.png} &
           \includegraphics[width=\imwidth]{img/attack/globe_vp.png} \\
            $\eta$ = 1 & $\eta$ = 0.957 & $\eta$ = 0.712 & $\eta$ = 0.683 & $\eta$ = 0.614 & $\eta$ = 0.117\\
    \bottomrule
    \end{tabular}
\caption{The figure shows watermarked images, images under various attacks, and our visual paraphrase method. The attacks include Brightness adjustment, Rotation, JPEG Compression, and Gaussian Noise, along with our \textbf{Visual Paraphrase} method. $\eta$ comparisons, representing watermark detection score of \textbf{HiDDeN}, are also provided.
}
\label{fig:attack_vp_comparison_5} 
\end{table*}

\begin{table*}
\centering
    \scriptsize
    \newcommand{\imwidth}{0.165\textwidth}
        \setlength{\tabcolsep}{0pt}
        \begin{tabular}{cccccc}
        \toprule
        Watermarked & Brightness & Rotation & JPEG Compression & Gaussian Noise & \textbf{Visual Paraphrase (Ours)} \\
        % \midrule
        \toprule
           \includegraphics[width=\imwidth]{img/attack/appendix_attack/attacks/botany_garden.png} &
           \includegraphics[width=\imwidth]{img/attack/appendix_attack/attacks/botany_garden_brightened.png} &
           \includegraphics[width=\imwidth]{img/attack/appendix_attack/attacks/botany_garden_rotated.png} &
           \includegraphics[width=\imwidth]{img/attack/appendix_attack/attacks/botany_garden_compressed.jpg} & 
           \includegraphics[width=\imwidth]{img/attack/appendix_attack/attacks/botany_garden_noisy.png} &
           \includegraphics[width=\imwidth]{img/attack/appendix_attack/attacks/botany_garden_vp.png} \\
            $\eta$ = 1 & $\eta$ = 0.942 & $\eta$ = 0.823 & $\eta$ = 0.735 & $\eta$ = 0.647 & $\eta$ = 0.107 \\
           
           \rule{0pt}{8ex}%
    
           \includegraphics[width=\imwidth]{img/attack/appendix_attack/attacks/carnival.png} &
           \includegraphics[width=\imwidth]{img/attack/appendix_attack/attacks/carnival_street_brightened.png} &
           \includegraphics[width=\imwidth]{img/attack/appendix_attack/attacks/carnival_street_rotated.png} &
           \includegraphics[width=\imwidth]{img/attack/appendix_attack/attacks/carnival_street_compressed.jpg} &
           \includegraphics[width=\imwidth]{img/attack/appendix_attack/attacks/carnival_street_noisy.png} &
           \includegraphics[width=\imwidth]{img/attack/appendix_attack/attacks/carnival_vp.png} \\
            $\eta$ = 1 & $\eta$ = 0.957 & $\eta$ = 0.851 & $\eta$ = 0.712 & $\eta$ = 0.598 & $\eta$ = 0.126 \\
           \rule{0pt}{8ex}%
    
           \includegraphics[width=\imwidth]{img/attack/appendix_attack/attacks/city_festival.png} &
           \includegraphics[width=\imwidth]{img/attack/appendix_attack/attacks/city_festival_brightened.png} &
           \includegraphics[width=\imwidth]{img/attack/appendix_attack/attacks/city_festival_rotated.png} &
           \includegraphics[width=\imwidth]{img/attack/appendix_attack/attacks/city_festival_compressed.jpg} &
           \includegraphics[width=\imwidth]{img/attack/appendix_attack/attacks/city_festival_noisy.png} &
           \includegraphics[width=\imwidth]{img/attack/appendix_attack/attacks/city_festival_vp.png} \\
            $\eta$ = 1 & $\eta$ = 0.973 & $\eta$ = 0.885 & $\eta$ = 0.759 & $\eta$ = 0.697 & $\eta$ = 0.197 \\
           \rule{0pt}{8ex}%
    
            \includegraphics[width=\imwidth]{img/attack/appendix_attack/attacks/market.png} &
           \includegraphics[width=\imwidth]{img/attack/appendix_attack/attacks/market_brightened.png} &
           \includegraphics[width=\imwidth]{img/attack/appendix_attack/attacks/market_rotated.png} &
           \includegraphics[width=\imwidth]{img/attack/appendix_attack/attacks/market_compressed.jpg} & 
           \includegraphics[width=\imwidth]{img/attack/appendix_attack/attacks/market_noisy.png} &
           \includegraphics[width=\imwidth]{img/attack/appendix_attack/attacks/market_vp.png} \\
            $\eta$ = 1 & $\eta$ = 0.987 & $\eta$ = 0.852 & $\eta$ = 0.764 & $\eta$ = 0.699 & $\eta$ = 0.139 \\
           \rule{0pt}{8ex}%
    
           \includegraphics[width=\imwidth]{img/attack/appendix_attack/attacks/mech_watch.png} &
           \includegraphics[width=\imwidth]{img/attack/appendix_attack/attacks/mech_watch_brightened.png} &
           \includegraphics[width=\imwidth]{img/attack/appendix_attack/attacks/mech_watch_rotated.png} &
           \includegraphics[width=\imwidth]{img/attack/appendix_attack/attacks/mech_watch_compressed.jpg} &
           \includegraphics[width=\imwidth]{img/attack/appendix_attack/attacks/mech_watch_noisy.png} &
           \includegraphics[width=\imwidth]{img/attack/appendix_attack/attacks/mech_watch_vp.png} \\
            $\eta$ = 1 & $\eta$ = 0.976 & $\eta$ = 0.839 & $\eta$ = 0.771 & $\eta$ = 0.658 & $\eta$ = 0.158 \\
    \bottomrule
    \end{tabular}
\caption{The figure shows watermarked images, images under various attacks, and our visual paraphrase method. The attacks include Brightness adjustment, Rotation, JPEG Compression, and Gaussian Noise, along with our \textbf{Visual Paraphrase} method. $\eta$ comparisons, representing watermark detection score of \textbf{HiDDeN}, are also provided.
}
\label{fig:attack_vp_comparison_6} 
\end{table*}

\clearpage
\subsection{An Interesting Observation -- Fourier behaviors}
Figure 10 illustrates the watermark patterns embedded in the Fourier space by various methods. Notably, the tree ring and zodiac watermark methods display distinct and recognizable characteristics in this domain, which are not observed in the Gaussian shading watermark. The exact contribution of these characteristics to the resilience of the watermarks against paraphrase attacks remains unclear at this stage. Further investigation into how these specific features contribute to watermark robustness will be a focus of our future work.

\begin{figure*}[h!]
\centering
    \scriptsize
    \newcommand{\imwidth}{0.42\textwidth}
    \resizebox{!}{7cm}{
    \begin{tabular}{cc}
    \toprule
    \textbf{Tree-Ring Ring} & \textbf{Tree-Ring Rand} \\
    \midrule
    \includegraphics[width=\imwidth]{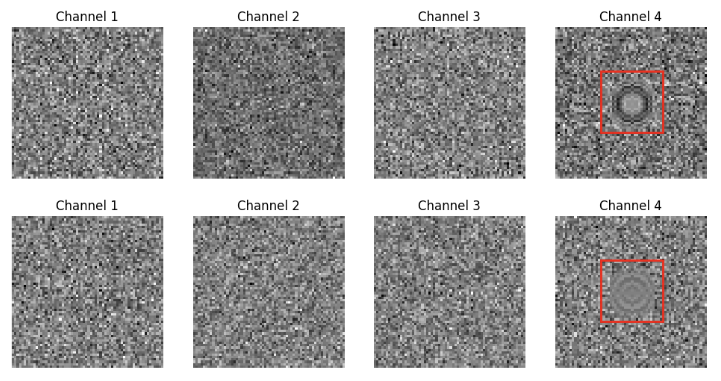} &
    \includegraphics[width=\imwidth]{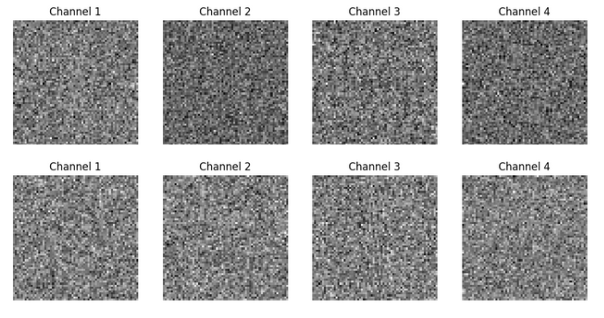} \\
    % \begin{tcolorbox}[width=\imwidth,title=Observations,colback=orange!5!white,colframe=orange!70!black]
    % \begin{itemize}
    %     \item Observations for Tree-Ring Ring
    % \end{itemize}
    % \end{tcolorbox} &
    % \begin{tcolorbox}[width=\imwidth,title=Observations,colback=orange!5!white,colframe=orange!70!black]
    % \begin{itemize}
    %     \item Observations for Tree-Ring Rand
    % \end{itemize}
    % \end{tcolorbox} 
    \\
    \midrule
    \textbf{Tree-Ring Zeros} & \textbf{Zodiac} \\
    \midrule
    \includegraphics[width=\imwidth]{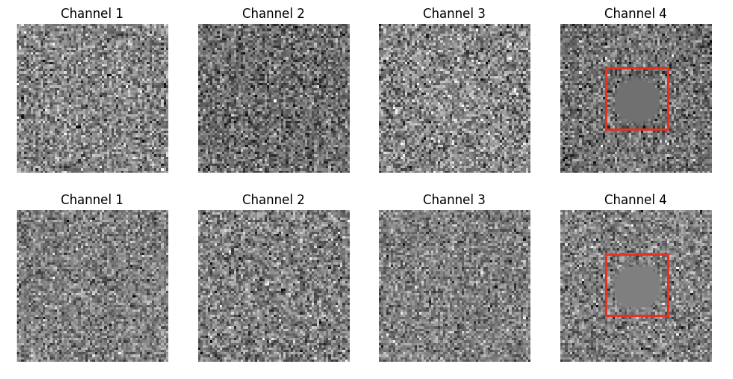} &
    \includegraphics[width=\imwidth]{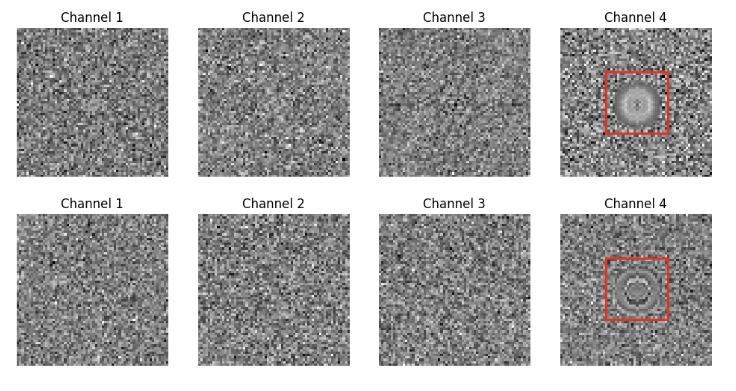} \\
    % \begin{tcolorbox}[width=\imwidth,title=Observations,colback=orange!5!white,colframe=orange!70!black]
    % \begin{itemize}
    %     \item Observations for Tree-Ring Zeros
    % \end{itemize}
    % \end{tcolorbox} &
    % \begin{tcolorbox}[width=\imwidth,title=Observations,colback=orange!5!white,colframe=orange!70!black]
    % \begin{itemize}
    %     \item Fourier space exhibits a distinct circular structure in the real and imaginary part of the latent vector, in the fourth channel. The circular pattern is similar to the Tree-Ring watermarks. 
    % \end{itemize}
    % \end{tcolorbox} 
    \\
    \midrule
    \multicolumn{2}{c}{\textbf{Gaussian Shading}} \\
    \midrule
    \multicolumn{2}{c}{\includegraphics[width=\imwidth]{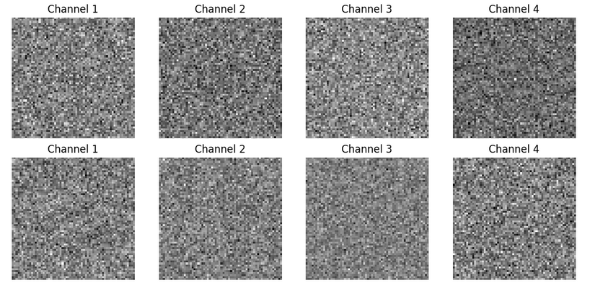}} \\
    \multicolumn{2}{c}{
    % \begin{tcolorbox}[width=\imwidth,title=Observations,colback=orange!5!white,colframe=orange!70!black]
    % \begin{itemize}
    %     \item Observations for Gaussian Shading
    % \end{itemize}
    % \end{tcolorbox}
    } \\
    \bottomrule    
    \end{tabular}
    }
    \begin{tcolorbox}[width=\textwidth,title=Observations,colback=orange!5!white,colframe=orange!70!black]
    \begin{itemize}
    [leftmargin=1mm]
        \item[\ding{224}] For \text{Tree-Ring}$_{\text{ring}}$ and ZoDiac, the Fourier space exhibits a distinct ring structure in the real and imaginary part of the latent vector, in the fourth channel. The pattern comprises of multiple rings and constant value along each ring.
        \item[\ding{224}] For \text{Tree-Ring}$_{\text{zeros}}$, the pattern is created by zeroing out the frequency components within a circular region in the frequency domain of an image, leading to a masked area in the spatial domain.
        \item[\ding{224}] For \text{Tree-Ring}$_{\text{rand}}$, Since the key is drawn from a Gaussian distribution and is designed to closely resemble the original noise characteristics of the Fourier modes, the watermarking introduces minimal alterations that blend seamlessly with the existing noise.
        \item[\ding{224}] Gaussian Shading watermark embedding preserves the image's latent representation's distribution and maintains visual consistency.
    \end{itemize}
    \end{tcolorbox}

\caption{Initial watermark latents for various watermark patterns with a latent space structure comprising 4 channels, each representing different abstract features. \textbf{The watermark is embedded in the last channel}. The figures show the real (top) and imaginary (bottom) components for the following patterns: (a) Tree-Ring Ring, (b) Tree-Ring Rand, (c) Tree-Ring Zeros, (d) Zodiac, and (e) Gaussian Shading.}
\label{fig:stacked_images}
\end{figure*}

% \begin{figure}
% \begin{tcolorbox}[enhanced,attach boxed title to top left={yshift=-1mm,yshifttext=-1mm,xshift=8pt},
% left=1pt,right=1pt,top=1pt,bottom=1pt,colback=orange!5!white,colframe=orange!70!black,colbacktitle=orange!70!black,
%   title=Research Questions on Concreteness,fonttitle=\ttfamily\bfseries\scshape\fontsize{8}{9}\selectfont,
%   boxed title style={size=small,colframe=violet!50!black},width=\textwidth ]

% \begin{spacing}{0.75}
%   \begin{itemize}
% [leftmargin=4mm]
% \setlength\itemsep{-0.5em}
%     \item[\tiny \ding{172}] {\footnotesize 
%     {\fontfamily{phv}\fontsize{6}{9}\selectfont
%    How does the level of concreteness in a prompt impact the probability of hallucination in LLMs?}} 
    
%     \item[\tiny \ding{173}] {\footnotesize 
%     {\fontfamily{phv}\fontsize{6}{9}\selectfont
%    How does concreteness affect different kinds of hallucination? and which LLM is more sensitive to concreteness vs. hallucination types?}}

%    \item[\tiny \ding{173}] {\footnotesize 
%     {\fontfamily{phv}\fontsize{6}{9}\selectfont
%    Are LLMs more prone to hallucination when given abstract or vague prompts compared to concrete and specific prompts?}}
   
% \end{itemize}
% \vspace{-4mm}  
% \end{spacing}
% \end{tcolorbox}
% \end{figure}

\end{document}